\documentclass[sigconf,final,screen,dvipsnames,x11names,table]{acmart}

\usepackage[utf8]{inputenc}
\usepackage[T1]{fontenc}
\usepackage{amsmath,mathtools}
\usepackage{float}
\usepackage{wrapfig}
\usepackage{setspace}
\usepackage{enumitem}
\usepackage{booktabs}
\usepackage{nicefrac}
\usepackage{microtype}
\usepackage{comment}
\usepackage{multirow}
\usepackage[ruled,vlined]{algorithm2e}
\usepackage{subcaption}

\definecolor{purple2}{RGB}{153, 50, 204}
\definecolor{green}{RGB}{0, 128, 0}

\emergencystretch=\maxdimen
\numberwithin{equation}{section}
\allowdisplaybreaks

\AtBeginDocument{
    \hypersetup{
        colorlinks=true,
        linkcolor=red,
        filecolor=DodgerBlue3,      
        urlcolor=red,
        citecolor=red
    }
}

\renewcommand\footnotetextcopyrightpermission[1]{}
\makeatletter
\newcommand{\vast}{\bBigg@{3}}
\newcommand{\Vast}{\bBigg@{4}}
\makeatother

\newcommand{\EX}{\mathbb{E}}

\renewcommand{\text}[1]{\textnormal{#1}}

\begin{document}

\title{Pseudo-differential-enhanced physics-informed neural networks}

\author{Andrew Gracyk}
\affiliation{%
  \institution{Department of Mathematics, Purdue University}
  \city{West Lafayette}
  \state{IN}
  \postcode{47907}
  \country{United States}
  \\
  \texttt{agracyk@purdue.edu}
}

\renewcommand{\shortauthors}{A. Gracyk}

\begin{abstract}
We present pseudo-differential enhanced physics-informed neural networks (PINNs), an extension of gradient enhancement but in Fourier space. Gradient enhancement of PINNs dictates that the PDE residual is taken to a higher differential order than prescribed by the PDE, added to the objective as an augmented term in order to improve training and overall learning fidelity. We propose the same procedure after application via Fourier transforms, since differentiating in Fourier space is multiplication with the Fourier wavenumber under suitable decay. Our methods are fast and efficient. Our methods oftentimes achieve superior PINN versus numerical error in fewer training iterations, potentially pair well with few/fixed samples in collocation, and can achieve breakthrough rather than gradual effects on the error. Moreover, our methods are suitable for fractional derivatives. We establish that our methods, due to the dynamical effects, improve spectral eigenvalue decay of the neural tangent kernel (NTK), and so our methods contribute towards the learning of high frequencies in early training, mitigating the effects of frequency bias up to the polynomial order and possibly greater with smooth activations. In particular, our primary contribution can be characterized as gradient enhancement affects spectral bias, and we specialize to Fourier space, although our theory does not prove the ambient case exactly due to a result with the Plancherel theorem on the descent paths. Our methods accommodate advanced techniques in PINNs, such as Fourier feature embeddings. A pitfall of discrete Fourier transforms via the Fast Fourier Transform (FFT) is mesh subjugation, and so we demonstrate compatibility of our methods for greater mesh flexibility and invariance on alternative Euclidean and non-Euclidean domains via Monte Carlo methods and otherwise, although possessing trade-offs of their own.

\vspace{2mm}
\end{abstract}


\maketitle

\pagestyle{fancy}
\fancyhead{} 
\renewcommand{\headrulewidth}{0pt} 

\fancyhead[LO,LE]{\sffamily\footnotesize Pseudo-differential-enhanced physics-informed neural networks}
\fancyhead[RO,RE]{\sffamily\footnotesize A. Gracyk}
\fancyfoot[C]{\sffamily\footnotesize \thepage}

{
\hypersetup{linkcolor=BurntOrange}
\tableofcontents
}

\section{Introduction}

We contribute to the Fourier modality for the physics-informed neural network (PINN) \cite{raissi2017physicsinformeddeeplearning} via the well-established gradient-enhancement augmentation but in Fourier space. Fourier options to PINNs are seasoned in literature as well \cite{Yu_2026} \cite{arni2025physicsinformedneuralnetworksfourier} \cite{cooley2025fourierpinnsstrongboundary} \cite{wang2023expertsguidetrainingphysicsinformed} \cite{huang2025fourierheuristicpinnssolve} due to the favorably disposed properties of differentiation done in a Fourier setting, since it is fundamental that differentiation before the Fourier transform results in differentiation of the Fourier basis function. The wavenumber, admitted as a multiplier, becomes the derivative for suitable decaying function done via integration by parts, meaning
\begin{gather}
\mathcal{F}[f'](\xi) = 2\pi i \xi \int_{-\infty}^{\infty} f(x) e^{-2\pi i \xi x} dx = 2\pi i  \xi \widehat{f}(\xi) 
\\
\text{or equivalently in higher dimensions}
\\
\mathcal{F}[\partial_j f](\xi) = 2\pi i  \xi_j \widehat{f}(\xi)   .
\end{gather}
The above identity will serve as the foundation for our work here, but not through the PDE residual standalone but rather differentiation of the PDE residual. In the above, we have used the decay condition
\begin{align}
\lim_{R \to \infty} \oint_{|x|=R} f(x) e^{-2\pi i \langle \xi , x \rangle} \nu_j  d\sigma = 0 .
\end{align}
\noindent Our approaches will be done with relatively simple linear differential operators (not pertaining to the PDE operator solution but rather the gradient enhanced operators) that have polynomial representations $P(\xi)$ in Fourier space called the \textit{symbol}. Thus, linear applications of linear differential operators to PDE residuals are products of PDE residuals with wavenumber representations and mere polynomials, and taking products with polynomials is fast and efficient. In particular, there exists an equivalence
\begin{align}
P(D) u = f \iff P(\xi) \widehat{u}(\xi) = \widehat{f}(\xi) ,
\end{align}
and so $P(D)$ as a nontrivial differentiation operator is transmuted through use of the Fourier transform to its polynomial characterization $P(\xi)$.

\vspace{2mm}

\noindent Gradient enhancement \cite{Yu_2022} \cite{iyer2025gradientenhancedselftrainingphysicsinformed} is admittedly a strong technique to reinforce the fidelity of PINNs, especially in low-collocation settings, and overall is relatively welcomed in most training settings for PINNs aside from the downfall of cost. Automatic differentiation in large scale scenarios is computationally nontrivial. We have observed this ourselves via our Navier-Stokes experiments, which undoubtedly require much larger scale differentiation of neural networks than univariate domain-type PDEs, especially of low order. Thus, especially in the multivariate domain scenario, the very same cost is incurred, and the benefits of gradient enhancement are mitigated via this tradeoff. This is where our methods provide new light: differentiation, expensive in physical space, has minute cost in Fourier space, since this differentiation becomes multiplication. We maintain the physics loss in physical space, which typically has low cost and is certainly lower than its gradient-enhanced counterpart. The typical autograd procedure in order to establish gradient enhancement is superannuated via the Fourier transform. Our overarching method is gradient enhancement in Fourier space, but we will attempt to develop a perspective towards these techniques via a looking glass of spectral bias.

\vspace{2mm}

\noindent \textbf{Frequency bias of neural networks.} It is a well known phenomenon that neural network capture low frequencies in data before high frequencies \cite{molina2024understandingdynamicsfrequencybias}, known as the (spectral) frequency bias \cite{basri2020frequencybiasneuralnetworks} \cite{xu2024overviewfrequencyprinciplespectralbias} \cite{rahaman2019spectralbiasneuralnetworks}. This manifests in one sense via early training by capturing essential features (low frequency) of data over finer details (high frequency). Our methods reconcile this phenomenon with the learning process by penalizing high frequencies with greater urgency, since these correspond to the Fourier wavenumber $\xi$. In particular, after taking a pseudo-differential of the PDE, additional terms involving polynomial orders of $\xi$ appear, and thus these higher frequencies have greater penalty in the loss of the form
\begin{align}
\label{eqn:spectral_loss}
\mathcal{L} = \int_{\mathcal{T}} \int_{\Xi} \underbrace{ \sum_i ( a_i |\xi|^i ) }_{\text{pseudo-differential appearance}} \underbrace{ ( \widehat{R}(\xi,t) ) }_{\text{physics-informed risk in Fourier space}} d \xi dt  ,
\end{align}
due to the higher scaling with large $|\xi|$. We remark, in the above in \ref{eqn:spectral_loss}, the pseudo-differential corresponds to multiplication with $\xi$ since differentiation is multiplication in Fourier space (which we will elaborate upon in detail with rigor later via the pseudo-differential).

\vspace{2mm}

\noindent \textbf{Gradient enhancement.} \noindent Using multi-index notation for nontrivial orders, let us consider a PDE residual of the form
\begin{align}
\dot{u}(x,t) + \Gamma \Big[ u(x,t), \dot{u}(x,t), \{\partial_x^{\beta} \partial_t^{\gamma} u(x,t)\}_{\beta,\gamma} \Big] ,
\end{align}
where $\Gamma : \{\partial_x^{\beta} \partial_t^{\gamma} u(x,t)\}_{\beta,\gamma} \in \prod_i C^1(\mathcal{X} \times [0,T]; \mathbb{R}) \cap W^{1,p}(\mathcal{X} \times [0,T]; \mathbb{R}) \rightarrow C^1(\mathcal{X} \times [0,T]; \mathbb{R}) \cap W^{1,p}(\mathcal{X} \times [0,T]; \mathbb{R})$ is an operator between function spaces, which is traditionally solved with the (made discrete) loss
\begin{align}
\mathcal{L}_{\text{physics}} = \EX_{u \sim \delta_{\nu}}  \Big| \Big| \dot{u}(x,t) + \Gamma \Big[ u(x,t), \dot{u}(x,t), \{\partial_x^{\beta} \partial_t^{\gamma} u(x,t)\}_{\beta,\gamma} \Big] \Big| \Big|_{L^2(\mathcal{X}\times [0,T])}   .
\end{align}
We will typically allow Clairaut's theorem (endow regularity in the above). Gradient enhancement differentiates the above, and so
\begin{align}
& \EX_{u \sim \delta_{\nu}}   \Big| \Big| \ \partial^{\alpha} \Big[  \dot{u}(x,t) + \Gamma \Big[ u(x,t), \dot{u}(x,t), \{\partial_x^{\beta} \partial_t^{\gamma} u(x,t)\}_{\beta,\gamma} \Big]  \Big] \  \Big| \Big|_{L^2(\mathcal{X} \times [0,T])}  
\\
& = \EX_{u \sim \delta_{\nu}}   \Big| \Big| \   \partial_t^{\alpha_0} \partial_{x_1}^{\alpha_1} \hdots \partial_{x_n}^{\alpha_n} \Big[  \dot{u}(x,t) 
\\
& \ \ \ \ \ \ \ \ \ \ \ \ \ + \Gamma \Big[ u(x,t), \dot{u}(x,t), \{\partial_x^{\beta} \partial_t^{\gamma} u(x,t)\}_{\beta,\gamma} \Big]  \Big] \  \Big| \Big|_{L^2(\mathcal{X} \times [0,T])}  
\end{align}
is added to the loss. It is notable to distguish partial differentiation from multi-index notation.

\vspace{2mm}

\noindent \textbf{Spectral PINNs.} Spectral PINNs (SINNs) \cite{Yu_2026} are a method to train a PINN in the classical sense but the physics loss is instead solved in Fourier space. The neural network directly outputs $\widehat{u}$ in Fourier space, thus Monte Carlo losses still apply, maintaining overall mesh invariance since there is no use of the FFT in the training stage. Our methods are similar but differ: our PINNs output $u$ as in traditional Euclidean space, but we transform differentiated $u$ into Fourier space and subsequently minimize this as a subsidiary residual. Thus, our methods maintain mesh invariance via the physics loss, and the enhanced Fourier loss can be done with or without meshes. To summarize, SINNs primarily examine losses of the form
\begin{align}
\text{physics loss} := \EX_{u \sim \delta_{\nu}} \Big| \Big| \widehat{\dot{u}}(\xi,t) + \widehat{\Gamma[u ]}(\xi,t) \Big| \Big|_{L^2(\Omega \times [0,T])} .
\end{align}

\vspace{2mm}

\noindent \textbf{Our methods on non-Euclidean geometries and without meshes.} PINNs most typically take domains of boxes $[a_1, b_1] \hdots [a_n, b_n] \times [0,T] \subseteq \mathbb{R}^n \times \mathbb{R}^+$, but not always \cite{zhou2025physicsinformedneuralnetworksirregular}. Our methods are adaptable to more nonstandard domains via Monte-Carlo methods and non-uniform FFTs. We will investigate the performance of Monte-Carlo methods as well. We refer to Appendix \ref{sec:grid_vs_montecarlo} for more details.

\vspace{2mm}

\noindent \textbf{Supplemental theoretical contributions.} Aside from proving an NTK result in Appendix \ref{app:spectral_decay}, in Appendix \ref{app:plancherel_sec}, we show $L^2$ loss equivalences with the Plancherel Theorem. We furthermore show that the gradients on the loss in the physical and Fourier spaces cases do not match, thus the parameter gradient flows upon the descent follow different trajectories. In Appendix \ref{app:spectral_preconditioning}, we provide a theoretical framework by moving the pseudo-differential operator inside the PDE instead of gradient-enhancing the outside. This is a preconditioning method for spectral filtration. With this method, we can scale the eigenvalues of the NTK not only up to the polynomial order, but by up to the scaling functions of the Fourier symbols. Moreover, by applying a pseudo-differential preconditioner, we can reduce the number of convolutions from a quadratic order, when inside, to a linear order, while achieving the same effects. We remark these sections are mostly theoretical and indeed support the work but not primarily.

\begin{figure*}[htbp]
  \vspace{0mm}
  \centering
  \includegraphics[scale=0.9]{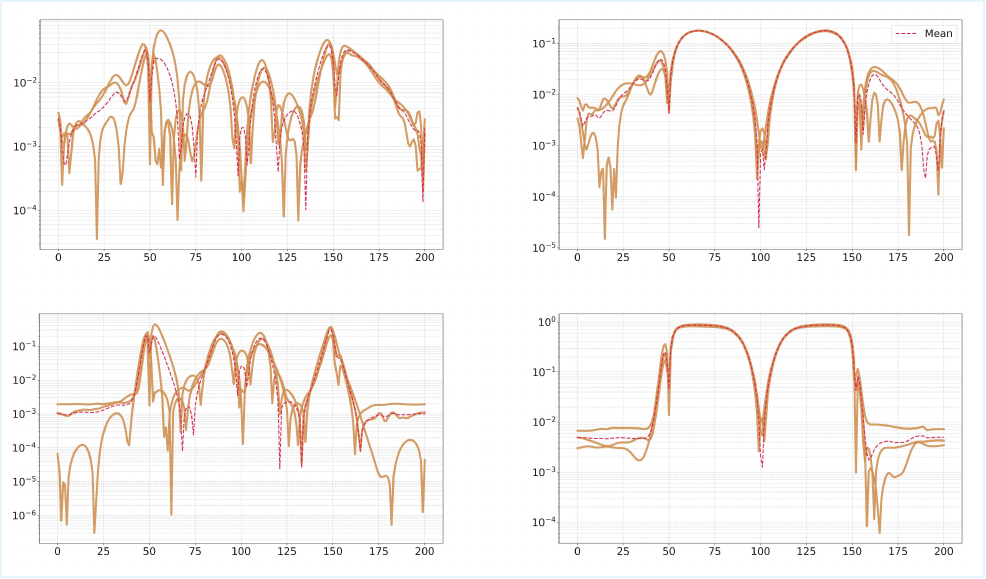}
  \caption{We plot (left) our Fourier enhanced PINN solution pointwise error on discretizations on three instances of retraining versus (right) a vanilla PINN on a log scale on two indices corresponding to $t=0.25, 0.75$ on the Allen-Cahn equation. Lower is better. In particular, we demonstrate Fourier enhanced PINNS (with FFTs) can achieve breakthrough loss effects rather than gradual loss descent effects.}
\label{fig:allen_cahn_samples_sidebyside}
\end{figure*}

\section{Notations and conventions}

We will denote 
\begin{gather}
\text{in the case of the PINN } \EX_{u \sim \delta_{\nu}} [ \mathcal{E}^{\dagger}[u] ]
\\
\text{and in the case of the (physics-informed) neural operator }
\\
\EX_{u \sim \mathcal{P}(\mathcal{A})} [ \mathcal{E}^{\dagger}[u] ]
\end{gather}
a loss term in the loss function, typically taking $\mu = \delta_{\nu}$ to be a Dirac probability mass at a particular function, i.e. $\int_{\mathcal{A}} \Psi(f) d\delta_{\nu} = \Psi(\nu)$, and $\mathcal{P}(\mathcal{A})$ denotes the collection of probability measures on Banach space $\mathcal{A}$, such as $L^1(\Omega)$ or $W^{k,p}(\Omega)$. $\mathcal{E}^{\dagger}$ is some (exact) cost function. We use this notation since it is most generalized and applicable for various machine learning archetypes. We mention neural operators \cite{quackenbush2024geometricneuraloperatorsgnps} \cite{fanaskov2024spectralneuraloperators} \cite{li2021fourierneuraloperatorparametric} \cite{Lu_2021} because they complement PDE learning in physics-informed approaches \cite{li2023physicsinformedneuraloperatorlearning} \cite{goswami2022physicsinformeddeepneuraloperator} \cite{karumuri2025physicsinformedlatentneuraloperator}, but we will not study this paradigm in our work here, but our work is easily extended for these approaches as well. We will denote $\mathcal{X}$ a compact domain in Euclidean space for the PINN and $\Omega$ a compact domain in Fourier space for the enhanced loss. We will always use the conventional approximation
\begin{align}
\| f \|_{L^2(\mathcal{X} \times [0,T])} \approx \text{const} \cdot  \sqrt{ \mathop{\sum \dots \sum \sum}\limits_{[(x,t)]_i \sim \text{Unif}(\mathcal{X}) \otimes \text{Unif}([0,T])} f(x_{1,i}, \dots, x_{n,i}, t_i)^2 } .
\end{align}
The above is taken with respect to the uniform Lebesgue measure (non-weighted). Moreover, for simplicity of argument, in our primary contribution section of \ref{sec:contribution}, we will assume our PDEs take the form
\begin{align}
& \Gamma \Big[ u(x,t), \dot{u}(x,t), \{\partial_x^{\beta} \partial_t^{\gamma} u(x,t)\}_{\beta,\gamma} \Big]
\\ & = \sum_{|\alpha|} \partial^{\alpha} u(x,t) = \sum_{|\alpha|}  \nabla_{x_1}^{\alpha_1} \hdots \nabla_{x_n}^{\alpha_n} u(x,t)  ,
\end{align}
although this assumption is more for convenience of argument and our methods are entirely applicable to more generalized PDEs. We refer to Appendices \ref{sec:generalized_linear_PDEs}, \ref{app:burger's}, \ref{app:allen-cahn}, \ref{app:kdv}, \ref{app:navier-stokes} for details on more sophisticated and nonlinear PDEs. Lastly, we will assume there exists a diffeomorphism
\begin{align}
\Psi^{\dagger}: \mathcal{X} \rightarrow \Omega ,
\end{align}
between our compact physical and Fourier domains (generally boxes).

\vspace{2mm}

\noindent All integrals in this work will be taken with respect to Lebesgue measure. We will assume regularity such as order of differentiation and integration can be exchanged, etc. Our regularity conditions will primarily be justified by (Sobolev) smoothness and compactness (our summations will be finite, and Lebesgue dominated convergence conditions will be met under sufficient smoothness and via the mean value theorem; if necessary, assume the domain is convex). For our experiments, our hyperparameters will vary heavily. We list hyperparameters in each figure respectively, as opposed to listing global hyperparameters for each experiment. We will slightly abuse notation by sometimes working with 1D notations, but our methods extend to multi-dimensions as well.

\vspace{2mm}

\noindent We will use the $2\pi i $ in the exponent Fourier transform convention, which means the factor of $1/(2\pi)^n$ is omitted outside our integrals. Sometimes the convention in harmonic analysis/PDEs $D_x = (1/2\pi i ) \partial_x$ is used: we will not use this convention. We will use the convolution theorem variously. We will most predominately use
\begin{align}
\mathcal{F}[uv] =  \mathcal{F}[u] * \mathcal{F} [ v ] ,
\end{align}
which derives from the more common formulation $uv = \mathcal{F}^{-1} [ \mathcal{F}[u] * \mathcal{F}[v] ]$ \cite{weisstein_convolution}. Occasionally in our experiments, we will refer to vanilla PINNs. This refers to a PINN with the exact same setup as without the Fourier enhanced loss; this does not mean a vanilla PINN with zero advanced techniques, such as Fourier feature embeddings.

\section{Pseudo-differential operator background}
\label{sec:pseudo_background}

Let us consider a (linear) differential operator $
P(D) = \sum_{\alpha} a_{\alpha}(x,t) D^{\alpha} $.
This operator can be rewritten as a polynomial $
P(\xi) = \sum_{\alpha} a_{\alpha}(x,t) \xi^{\alpha}$.
with respect to Fourier space in the sense that \cite{chai2025frozengaussianapproximationfractional}
\begin{align}
P(D)u(x) =  \int_{\mathbb{R}^n} \int_{\mathbb{R}^n} e^{2 \pi i \langle x-y , \xi \rangle} P(\xi) u(y)  dy d\xi .
\end{align}
Here, $P(\xi)$ is called a \textit{symbol} and $P(D)$ is called the \textit{pseudo-differential}. Thus, the differential operators acts as multiplication in Fourier space.

\vspace{2mm}

\noindent A pseudo-differential operator $P(x,D)$ on $\mathbb{R}^n$ is an operator acting on $u$ such that
\begin{align}
\label{eqn:pseudo_diff_operator}
P(x,D) u (x) =  \int_{\mathbb{R}^n} e^{2 \pi i \langle x, \xi \rangle} P(x,\xi) \widehat{u}(\xi) d\xi .
\end{align}
It is equivalent to the above but acts in nontrivial, nonclassical cases, for example fractional derivatives and other operators impossible to write in the prototypical Leibniz notation. Using the definition of the Fourier transform, we also have
\begin{align}
P(x,D) u (x) =  \int_{\mathbb{R}^n} \int_{\mathbb{R}^n} e^{2 \pi i \langle x - y, \xi \rangle} P(x,\xi) u(y) dy d\xi ,
\end{align}
where we have rewritten the above using linearity of the integral (notice this is different than Fubini's/Tonelli's) and since $P \in S_{\rho,\delta}^m$. In particular, it can be noted $P(\xi)$ belongs in the class
\begin{align}
& p(x,\sigma) \in S_{\rho,\delta}^m 
\\
& := \Bigg\{ p \in C^{\infty}(\mathcal{X} \times \dot{\mathbb{R}}^n ) , \Bigg| D_x^{\rho} D_{\sigma}^{\delta} p(x,\sigma) \Bigg| \leq C_{\rho, \delta} ( 1 + | \sigma |)^{m - |\delta|} \Bigg\}
\end{align}
under $m$, and $\dot{\mathbb{R}}^n = \mathbb{R}^n \setminus \{ 0 \}$, which ensures suitable decay and overall regularity. As a last remark, for empirical reasons, we will almost always consider
\begin{align}
P(x,D) u (x) \approx  \int_{\mathbb{R}^n} \int_{\mathbb{R}^n} e^{2 \pi i \langle x - y, \xi \rangle} P(x,\xi) u(y) \cdot \chi_{K}(y) \cdot dy \cdot \chi_{\Omega}(\xi) \cdot d\xi ,
\end{align}
where $\chi$ is the indicator, and $K,\Omega$ are suitable compact domains to ensure the above is approximated over a sufficiently large finite domain, since it is not reasonable for us to search all of $\mathbb{R}^n$ to evaluate the integrals.

\vspace{2mm}

\noindent The pseudo-differential operator is mathematically archetypal because it is well-defined under fractional derivatives. We will consider
\begin{align}
P(\xi) = \sum_{|\alpha|\leq m, \alpha \notin \mathbb{N}} a_{\alpha} \xi^{\alpha} ,
\end{align}
for example in a risk minimization framework
\begin{align}
\tilde{R}_{\theta}(x,t) & \in \Bigg\{ R(x,t) \Bigg| \int_{\mathbb{R}^n} \int_{\mathbb{R}^n} \Big|  (-\Delta)^{s/2} R(x,t) \Big| 
\\
& \ \ \ \ \ \ \ \ \times \chi_{\mathcal{X} \times [0,T]}(x,t) dx dt < \delta, s \in \mathbb{N}, \delta \in \mathbb{R}^+ \Bigg\} .
\end{align}
We will study this pseudo-differential in an empirical setting and its effects on PDE learning via Fourier gradient enhancement.

\vspace{2mm}

\noindent Moreover, note the following Fourier transform of equation \ref{eqn:pseudo_diff_operator}
\begin{align}
\widehat{(P(x,D) u (x))}(\eta) & =  \int_{\mathbb{R}^n} e^{-2\pi i \langle x, \eta \rangle} \int_{\mathbb{R}^n} e^{2 \pi i \langle x, \xi \rangle} P(x,\xi) \widehat{u}(\xi) d\xi dx
\\
& = \int_{\mathbb{R}^n}  \int_{\mathbb{R}^n}  e^{2 \pi i \langle x, \xi - \eta \rangle} P(x,\xi) \widehat{u}(\xi) d\xi dx .
\end{align}
Denoting 
\begin{align}
\widehat{P}(\zeta, \xi) = \int_{\mathbb{R}^n} e^{-2 \pi i \langle x, \zeta \rangle} P(x,\xi) dx  ,
\end{align}
observe
\begin{align}
\widehat{(P(x,D) u)}(\eta) & =  \int_{\mathbb{R}^n} \widehat{P}(\eta - \xi, \xi) \widehat{u}(\xi) d\xi = (\widehat{P} * \widehat{u})(\eta) , 
\end{align}
thus if $a$ depends on $x$, the above is expressed as a convolution.

\section{Our contribution}
\label{sec:contribution}

\begin{figure}[htbp]
  \vspace{0mm}
  \centering
  \includegraphics[scale=0.5]{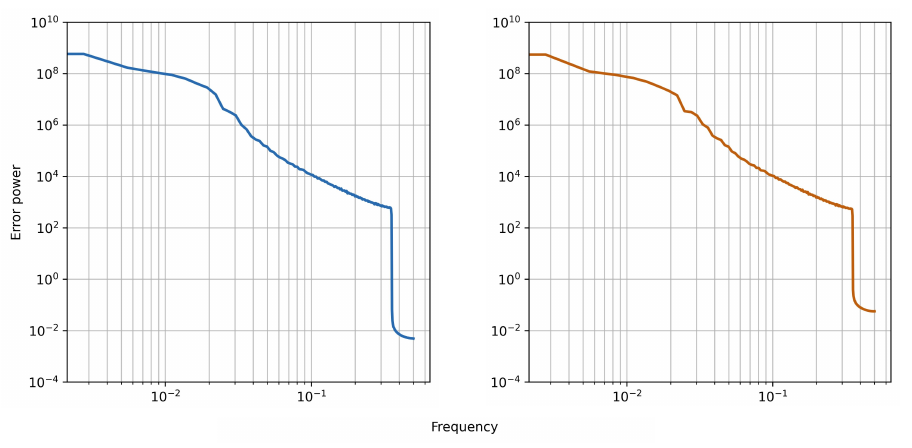}
  \caption{We demonstrate on our Navier-Stokes experiment with severe Fourier enhanced loss tuned coefficient that high frequencies are learned faster. We train a Navier-Stokes PINN with (left) and without (right) Fourier enhancement with coefficient $0.5 \times \lambda_{\text{physics}}$ on a vanilla MLP with Fourier feature embedding and the SOAP optimizer for $\sim 500$ epochs. High error power at high frequency corresponds to inability to learn high frequency. Thus, our methods mitigate spectral bias since lower is better. We remark that this figure also demonstrates lower frequencies are less affected (although this is not true in general; see Figure \ref{fig:burgers_nonsquare_error_power}). We also remark this example is just a slice at a specific iteration, and is not characteristic of spectral bias across the training process (we refer to Figures \ref{fig:kdv_spectral_training_combined}, \ref{fig:kdv_spectral_training_veryhighbatch},  \ref{fig:kdv_spectral_training_fftsmall},
  \ref{fig:kdv_spectral_training_montecarlo}, \ref{fig:navier_stokes_spectral_training}).}
  \label{fig:error_power_navier_stokes}
\end{figure}

Consider the pseudo-differential gradient-enhanced PDE residual
\begin{align}
P(x,t,D)  \Bigg[ \partial_t u + \Gamma \Big[ u(x,t), \{\partial_x^{\beta} \partial_t^{\gamma}  u(x,t)\}_{\beta,\gamma} \Big] \Bigg] = 0 .
\end{align}
We have taken the pseudo-differential outside the PDE residual. In general, Clairaut's theorem is not applicable to the above and we do not have commutativity (see Appendix \ref{app:spectral_preconditioning}). Moreover, let us restrict $P$ constant in space and time.

\vspace{2mm}

\noindent \textbf{Theorem 1.} Let $u \in C^1(\mathcal{X} \times [0,T]; \mathbb{R}) \cap W^{p,1}(\mathcal{X} \times [0,T]; \mathbb{R})$ be a solution to a PDE of the form $\dot{u} + \Gamma[u, \hdots]=0$, where $\Gamma$ is a linear operator with identity coefficients. Let $P(x,t,D) = \sum_{\alpha} a_{\alpha}(x,t) D^{\alpha}$ be a linear differential operator. In the case that $a_{\alpha}$ are constants, i.e. $P(x,t,D) = P(D)$,
\begin{align}
P(x,t,D) \partial_t u & = \Bigg(  \int_{\mathbb{R}^n} \int_{\mathbb{R}^n} e^{2 \pi i \langle x - y, \xi \rangle} P(\xi) \dot{u}(y) dy d\xi \Bigg) 
\\
& = \sum_{|\alpha|\leq m} a_\alpha D_x^\alpha \dot{u} = \mathcal{F}^{-1} \Big( P(\xi) \cdot \mathcal{F}(\dot{u}) \Big) ,
\\[1em]
P(x,t,D) D_x^{\beta} u  & = \sum_{|\alpha| \leq m} a_{\alpha} \Bigg(  \int_{\mathbb{R}^n} \int_{\mathbb{R}^n} e^{2 \pi i \langle x - y, \xi \rangle} (2 \pi i )^{\beta} \xi^{\alpha + \beta} u(y) dy d\xi \Bigg)  
\\
& = \mathcal{F}^{-1} \Big( P(\xi) \cdot (2 \pi i \xi)^\beta \cdot \mathcal{F}(u) \Big) .
\end{align}

\vspace{2mm}

\noindent \textit{Proof of Theorem 1.} The proof is routine. Observe, using $P(x,t,\xi) = \sum_{|\alpha|\leq m} a_\alpha \xi^\alpha$
\begin{align}
P(D)u & =  \int_{\mathbb{R}^n} e^{2 \pi i \langle x, \xi \rangle} P(x,t,\xi) \Big( \int_{\mathbb{R}^n} e^{-2 \pi i \langle y, \xi \rangle} u(y) dy \Big) d\xi
\\
&  =  \int_{\mathbb{R}^n} e^{2 \pi i \langle x, \xi \rangle} P(\xi) \widehat{u}(\xi) d\xi ,
\end{align}
which implies by simply exchanging $u \rightarrow \dot{u}$ and under regularity (the sum is finite)
\begin{gather}
P(D)\dot{u} = \sum_{|\alpha|\leq m} a_{\alpha} \Big(  \int_{\mathbb{R}^n} e^{2 \pi i \langle x, \xi \rangle} \xi^\alpha \mathcal{F}(\dot{u})(\xi) d\xi \Big)
\\
 =  \int_{\mathbb{R}^n} e^{2 \pi i \langle x, \xi \rangle} \Big( P(\xi) \cdot \mathcal{F}(\dot{u})(\xi) \Big) d\xi = \mathcal{F}^{-1} \Big( P(\xi) \cdot \mathcal{F}(\dot{u}) \Big) .
\end{gather}
Now, in the spatial derivative case, since $a$ is constant,
\begin{gather} P(D) D_x^\beta u  = \sum_{|\alpha|\leq m} a_{\alpha} \Big(  \int_{\mathbb{R}^n} e^{2 \pi i \langle x, \xi \rangle} \xi^\alpha \cdot (2 \pi i \xi)^\beta \widehat{u}(\xi) d\xi \Big) 
\\=  \int_{\mathbb{R}^n} e^{2 \pi i \langle x, \xi \rangle} \Big( P(\xi) \cdot (2 \pi i \xi)^\beta\cdot \mathcal{F}(u)(\xi) \Big) d\xi
\\
= \mathcal{F}^{-1} \Big( P(\xi) \cdot (2 \pi i \xi)^\beta \cdot \mathcal{F}(u) \Big) . \end{gather}

\noindent $\square $

\subsection{Objectives}

Taking the Fourier transform of both terms in the previous section, our enhanced physics loss becomes
\begin{align}
\label{eqn:fourier_physics_loss}
  \mathcal{L}_{\text{Fourier enhanced}} = \mathbb{E}_{u \sim \delta_{\nu}} \left[ \Bigg| \Bigg| P(\xi) \mathcal{F}( \dot{u} ) + \sum_{\beta} P(\xi) (2 \pi i \xi)^{\beta} \mathcal{F}(u) \Bigg| \Bigg|_{L^2(\Omega \times [0,T])}^2 \right]
\end{align}
We crucially remark the above ignores scaling functions of the operator of the PDE, but the above can easily be extended for those cases as well. For clarity, note that
\begin{align}
& \mathbb{E}_{u \sim \delta_{\nu}} \Bigg[ \Bigg| \Bigg| \underbrace{ P(\xi) }_{\substack{\text{weight contribution from} \\ \text{pseudo-differential}}}  \mathcal{F}( \dot{u} ) 
\\
& \ \ \ \ \ \ \ + \sum_{\beta}  P(\xi)\underbrace{ (2 \pi i \xi)^{\beta} }_{\substack{\text{contribution from the} \\ \text{derivatives of the PDE}}} \mathcal{F}(u) \Bigg| \Bigg|_{L^2(\Omega \times [0,T])}^2 \Bigg] .
\end{align}
The above is exactly the PDE transformed into Fourier space added as a residual, but it is weighed by $P(\xi)$, which is nonconstant, thus
\begin{gather}
P(\xi) := \text{spectral weight} 
\\
P(\xi) \text{ typically } = (2\pi i \xi) + (2\pi i \xi)^{2} + \hdots 
\\
\text{or in higher dimensions},
\\
P(\xi) = \sum_{|\alpha| \geq 1, \max_{\alpha} |\alpha| = m < \infty}(2 \pi i)^{|\alpha|} \xi_{1}^{\alpha_1} \hdots \xi_n^{\alpha_n} ,
\end{gather}
where the above can terminate to any order corresponding to a derivative order. Thus, we also have the loss argument as
\begin{gather}
 \EX_{u \sim \delta_{\nu}} \Bigg| \Bigg| \Big[ (2\pi i \xi) + (2\pi i \xi)^{2} + \hdots  \Big] \cdot \Big[  \mathcal{F}( \dot{u} ) + \sum_{\beta}  (2 \pi i \xi)^{\beta} \mathcal{F}(u)  \Big]  \Bigg| \Bigg|_{L^2(\Omega \times [0,T])}^2 
\\
\text{or equivalently}
\\
\EX_{u \sim \delta_{\nu}} \Bigg| \Bigg|  \Big[ \text{spectral weights from gradient enhancement in Fourier space} \Big] 
\\
\cdot \Big[\text{PDE residual in Fourier space} \Big]  \Bigg| \Bigg| .
\end{gather}We reiterate: $P(\xi)$ also corresponds to derivatives, since the above is gradient enhancement in Fourier space. Note that constant coefficients are generally easier to have because nonconstant coefficients require convolution, i.e.
\begin{align}
& \EX_{u \sim \delta_{\nu}} \Bigg| \Bigg|  \sum_{\alpha}  \Bigg[ (\widehat{a}_{\alpha}(\xi,t)  * \Big[  (2 \pi i \xi)^k \Big( \mathcal{F}( \dot{u} ) 
\\
& \ \ \ \ \ \ \  \ \ \ \ \ \ \   \ \ \ \ \ \ \ \ \ \ \ \ \ \ \ \ \ \ \ \ \  + \sum_{\beta}  (2 \pi i \xi)^{\beta} \mathcal{F}(u)  \Big) \Big]  \Bigg]  \Bigg| \Bigg|_{L^2(\Omega \times [0,T])}^2  .
\end{align} 
This matches the derivation we saw at the end of section \ref{sec:pseudo_background}.

\vspace{2mm}

\noindent In fact, $P(\xi)$ is a function, not a constant, so it cannot be "divided out", which would therefore leave the PDE transformed into Fourier space alone in the loss. Omitting $P(\xi)$ fundamentally changes the loss landscape, which would overall bypass the pseudo-differential aspect of the loss, since solving the transformed PDE in a physics loss without $P(\xi)$ is a task independent of a pseudo-differential operator.

\vspace{2mm}

\noindent If we denote $u : \mathcal{X} \times [0,T] \times \Theta \rightarrow L^1(\mathcal{X} \times [0,T]; \mathbb{R})$ be the neural network physics solution, the total loss in the case of the PINN is
\begin{gather}
\inf_{\theta \in \Theta} \  \EX_{u \sim \delta_{\nu}} \lambda_{\text{physics}} \Big|\Big| \ \partial_t u_{\theta} + \Gamma[u(x,t), \dot{u}(x,t), \{\partial_x^{\beta} \partial_t^{\gamma} u(x,t)\}_{\beta,\gamma}]  \ \Big|\Big|_{L^2(\mathcal{X} \times [0,T])}^2  
\\
+ \lambda_{\text{boundary}} \Big|\Big| \ \partial_t u_{\theta}(\cdot, 0) + \Gamma[u(x,0), \dot{u}_{\theta}(x,0), \hdots ] \ \Big|\Big|_{L^2(\mathcal{X})}^2
\\
+ \lambda_{\text{boundary}} \Big|\Big| \  \partial_t u_{\theta}(x, t) \Big|_{x \in \partial \Omega} + \Gamma[u_{\theta}(x,t), \dot{u}_{\theta}(x,t), \hdots ]  \Big|_{x \in \partial \Omega} \ \Big|\Big|_{L^2([0,T])}^2 
\\
+ \lambda_{\text{Fourier}} \Big| \Big| \ P(\xi) \mathcal{F}( \dot{u}_{\theta} ) + \sum_{\beta} P(\xi) (2 \pi i \xi)^{\beta} \mathcal{F}(u_{\theta}) \ \Big| \Big|_{L^2(\Omega \times [0,T])}^2 .
\end{gather}

\vspace{2mm}

\noindent It may be the case $P(\xi)$ blows up (this is supposed to be happen by definition of the derivative extended to Fourier space), in which the enhanced loss will dominate. We rectify this by normalizing by a maximum, i.e.
\begin{align}
& \mathcal{L}_{\text{enhanced}} = \EX_{u \sim \delta{\nu}}  \Big| \Big| \frac{ P(\xi) }{|| P(\xi) ||_{L^{\infty} ( \Omega \times [0,T])}} \mathcal{F}( \dot{u} ) 
\\
& \ \ \ \ \ \ \ \ \ + \sum_{\beta} \frac{ P(\xi) }{|| P(\xi) ||_{L^{\infty} ( \Omega \times [0,T])}} (2 \pi i \xi)^{\beta} \mathcal{F}(u) \Big| \Big|_{L^2(\Omega \times [0,T])}^2  .
\end{align}
Dividing by a norm $|P(\xi)|$ dependent on $\xi$ can distort the weighing effect. One could add a smoothed-in update rule of the form $\lambda \leftarrow \alpha \lambda_{\text{new}} + (1-\alpha) \lambda_{\text{new}}$ if this is found suitable. Computationally, we remark we detach the norm. Since $P(\xi)$ is complex-valued, we use the norm
\begin{align}
& ||P(\xi)||_{L^{\infty}( \Omega \times [0,T]) } 
\\
& = \inf_{\tau} \Bigg\{  \sqrt{\text{Re}(P(\xi))^2 + \text{Im}(P(\xi))^2}  \leq \tau \Big| \xi \in \Omega, t \in [0,T] \Bigg\} .
\end{align}

\vspace{2mm}

\noindent In Appendix \ref{app:plancherel_sec}, we discuss loss equivalences with the Plancherel Theorem. Moreover, we show using the complex conjugate product version of the Plancherel Theorem, along with the chain rule on quadratic loss, the gradients of the $L^p$ losses are not equivalent and lead to different training processes.

\subsection{Quantile loss}

\noindent We find the Fourier loss sensitive to small perturbations and outliers that come with the frequency-based distortions that are brought in by the FFT. Therefore, we find a performance increase in the loss
\begin{align}
& \mathcal{L}_{\text{enhanced}} = \EX_{u \sim \delta_{\nu}}  \text{quantile} \Bigg\{ \frac{ P(\xi) }{|| P(\xi) ||_{L^{\infty}(\Omega \times [0,T])}}  
\\
& \ \ \ \ \ \ \ \ \ \ \times \Bigg( \mathcal{F}( \dot{u} ) + \sum_{\beta}  (2 \pi i \xi)^\beta \mathcal{F}(u) \Bigg) \Bigg\} 
\\
& = \EX_{u \sim \delta_{\nu}} \inf_{z}  \Bigg\{ z \in \mathbb{R}^+ : \text{Pr}_{\xi \sim \mu} \Bigg( \Bigg|  \frac{ P(\xi) }{|| P(\xi) ||_{L^{\infty}(\Omega \times [0,T])}}  
\\
& \ \ \ \ \ \ \ \ \ \ \times  \Bigg( \mathcal{F}( \dot{u} ) + \sum_{\beta}  (2 \pi i \xi)^\beta \mathcal{F}(u) \Bigg) \Bigg| \leq z \Bigg) \geq \tau \Bigg\} ,
\end{align}
yielding a more stable, well-behaved objective. We crucially remark the quantile parameter $\tau$ should remain large to mitigate cutting off the more severe scenarios of spectral bias ($0.9-0.95$ works well or even $0.99$). Recall the quantile function is not differentiable, but it has existing subgradients.

\subsection{Randomized grid training}

We found empirical success in randomly sampling the mesh sizes each epoch when FFTs were used (see Appendix \ref{alg:diverse_posterior} for scenarios when FFTs are not used and non-square domains are of interest). Recall FFTs are (typically) mappings from meshes to meshes, although more flexible domains for FFTs exist \cite{armstrong2024directsolutioninterpolativeinverse}. Thus, we consider FFTs of the form
\begin{align}
\widehat{f}(\xi) = \Big( \prod_{j=1}^n N_j \Big) \EX_{m \sim \mathcal{U}(\mathcal{G})} \Bigg[ f(m_1,\hdots,m_n) e^{-2 \pi i \sum_{j=1}^n \xi_j m_j / N_j} \Bigg] 
\end{align}
for suitable uniform distribution $\mathcal{U}(\mathcal{G})$.

\subsection{Eigenvalue decay}

In this section, we discuss how our methods improve eigenvalue decay of the neural tangent kernel (NTK).

\vspace{2mm}

\noindent Let us discuss preliminary background work that we will use. \cite{bietti2021deepequalsshallowrelu} shows that for ReLU networks, the eigenvalues of the neural tangent kernel follow
\begin{align}
\lambda_{\xi} \sim \text{constant} \cdot \xi^{-d - 2\nu + 1} 
\end{align}
where $C_1, C_2, d,\nu$ are constants, where $\xi$ is the wavenumber (frequency). ReLU-type activation is generally adverse for PINNs due to possibly nonexistent orders of continuous differentiability and this result is not primarily applicable for us.

\vspace{2mm}

\noindent \cite{murray2023characterizingspectrumntkpower} is a crucial work for us and discusses NTK eigenvalue decay under smooth activations. Specifically,
\begin{align}
\lambda_{\xi} \sim \mathcal{O}(\xi^{-d+1}a^{-\xi} ), \ \ \ \ \ \lambda_{\xi} \sim \Omega( \xi^{-d/2+1} 2^{-\xi} a^{-\xi} ) .
\end{align}
This work emphasizes spherical domains under normalization, but suitable for PINNs under diffeomorphisms mapping compact, bounded domains as in PINNs to the spherical manifold. \cite{Li2023OnTE} is a work that discusses NTK eigenvalue decay on more generalized domains via Theorem 8. We refer to \cite{holzmüller2025reluactivationsaffectneural} for other relevant literature on NTKs with smooth activations.

\vspace{2mm}

\noindent \textit{\textbf{Theorem 2.} Let $u \in C^1(\mathcal{X} \times [0,T]; \mathbb{R}) \cap W^{p,1}(\mathcal{X} \times [0,T]; \mathbb{R})$ be a solution to a PDE of the form $\dot{u} + \Gamma[u, \hdots]=0$. Let $\omega(\xi)$ be a polynomial weight dependent on wavenumber in Fourier space. Suppose the domain is periodic. Assume sufficient regularity conditions are met. Suppose $\text{tanh}(\cdot)$ is used. Then, with the Fourier enhanced loss $\mathcal{L}_{\text{Fourier enhanced}}$ and under gradient descent on parameter $\theta$, the infinite-width limit NTK eigenvalue decay becomes $
\tilde{\lambda}(\xi) \sim |\xi|^{s} F(\xi)$ for some $F(\xi)$, where $s > 0$. }

\vspace{2mm}

\noindent\textit{Remark.} It is known that the neural tangent kernel (NTK) eigenvalue decay with certain smooth activation is exponential such that $\lambda(\xi) \sim |\xi|^{-q} Q^{-|\xi|}$ for some $q,Q$ dependent on wavenumber and the order of the PDE. This product of functions is less established in literature in the PINN setting (see proof in Appendix \ref{app:spectral_decay} for more details of the NTK for PINNs). NTK eigenvalue decay is discussed in \cite{gan2025neuraltangentkernelneural} and theoretically proves that eigenvalue decay is largely unaffected by differential operators. For simplicity, we denote spectral decay with $F(\xi)$.

\vspace{2mm}

\noindent \textit{Sketch of proof.} Forward discrete gradient descent corresponds to a continuum-limit ODE. After differentiating Fourier risk and the Fourier loss, we can insert the gradient flow, yielding the NTK. The pseudo-weight term collateralizes with the polynomial term degree.

\vspace{2mm}

\noindent \textit{Remark.} We will also develop, primarily in section \ref{app:spectral_preconditioning}, that we can scale the eigenvalue decay with new functions so the eigenvalue decay is of the form
\begin{align}
\tilde{\lambda}(\xi) \sim \sum_{\alpha} \phi_{\alpha}(\xi) |\xi|^{\alpha} F(\xi) ,
\end{align}
therefore there is function contribution from possible non-polynomials in the gradients.

\subsection{Additional results}

\textit{\textbf{Theorem 3.} Let $u_{\theta} \in C^1(\mathcal{X} \times [0,T]; \mathbb{R}) \cap W^{p,1}(\mathcal{X} \times [0,T]; \mathbb{R})$ be a parameterized function optimized to solve a partial differential equation. Assume the training objectives are evaluated over compact, truncated spatial and frequency domains. While the physical-space enhanced loss and the Fourier enhanced loss are equivalent over unbounded domains via Plancherel's Theorem, gradient descent on these two truncated objectives follows different optimization trajectories. }

\subsection{Additional discussion}

\textbf{Fractional calculus.} Spectral methods in PINNs are suitable for fractional calculus-type PDEs. For example, on losses of the form
\begin{align}
& \inf_{\theta \in \Theta} \ \EX_{u \sim \delta_{\nu}} \Big| \Big| \widehat{\dot{u}} + \widehat{\Gamma \Big[ u, \dot{u}, \{ (-\Delta)^s u\}_s \Big] } \Big| \Big|_{L^2(\Omega \times [0,T])}^2 
\\
& \ \ \ \ \ \ \ \ + \Big| \Big| P(\xi) \Big[ \widehat{\dot{u}} + \widehat{\Gamma \Big[ u, \dot{u}, \{ (-\Delta)^s u\}_s \Big] } \Big] \Big| \Big|_{L^2(\Omega \times [0,T])}^2 ,
\end{align}
plus other terms. Note that the physics loss is spectral here.

\section{Experiments}

\vspace{2mm}

\begin{figure}[htbp]
  \centering
  \begin{subfigure}[b]{\linewidth}
    \centering
    \includegraphics[scale=0.29]{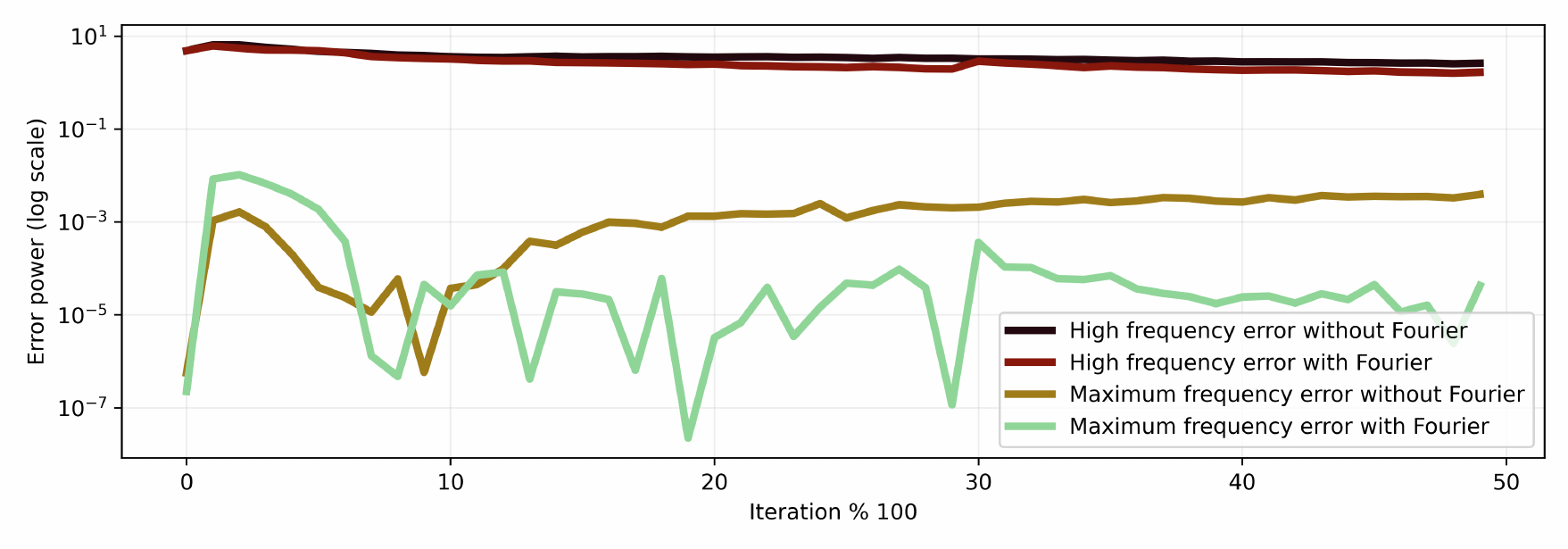}
    \label{fig:kdv_spectral_training_highbatch}
  \end{subfigure}
  
  \vspace{-0.5cm} 
  
  \begin{subfigure}[b]{\linewidth}
    \centering
    \includegraphics[scale=0.29]{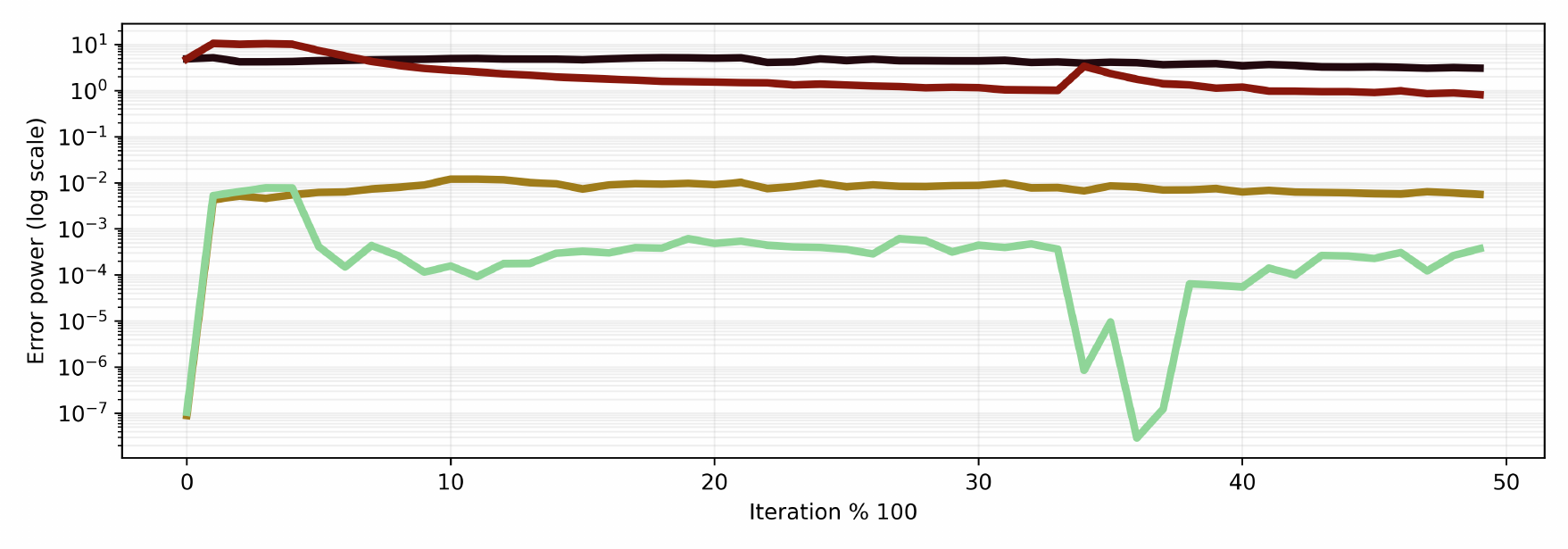}
    \label{fig:kdv_spectral_training_veryhighbatch}
  \end{subfigure}

  \vspace{-0.5cm}
  \caption{We plot the spectral bias maximum error power according to Algorithm \ref{alg:radial_power_alg} and high error power for the KdV equation across a Fourier enhanced PINN with FFTs and a baseline PINN across 5,000 training iterations. We use learning rate $1\text{e}{-3}$; (top) batch size of 600 and (bottom) a very high batch size of 1,500; and use $\lambda_{\text{Fourier}} = 0.02$. We use an enhanced MLP with $\sigma=3.0$ and initial condition $u(x,0) = -0.65 \cos( \pi x)$. We use a larger batch size in both, which is notable because with high probability the collocation points will encompass the domain (especially in 1D, which is what we have here) with high coverage, i.e. the largest approximate open intervals possible that do not contain collocation points will have small Lebesgue measure. Moreover, the FFTs will not seriously cover points "that are missed" due to such an existence of maximally-taken small-measured open intervals. This conceptually is important to several of our figures, although less applicable to our 2D Navier-Stokes.}
  \label{fig:kdv_spectral_training_combined}
\end{figure}

\begin{figure}[htbp]
  \vspace{0mm}
  \centering
  \includegraphics[scale=0.29]{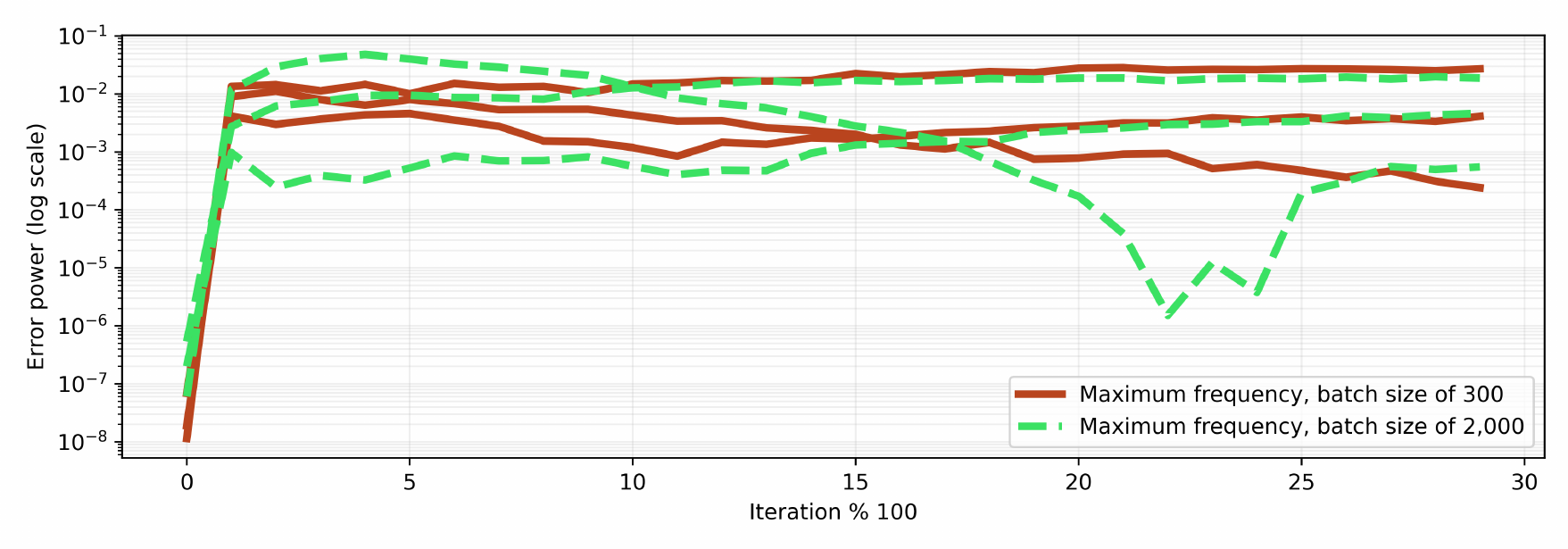}
  \caption{We plot maximum frequency spectral bias learning with a vanilla (non-Fourier enhanced) PINN across 3 instances of training with: (1) a moderate batch size of 300; (2) a large batch size of 2,000. Since the spectral error power is mostly equal for both, the spectral bias is largely invariant of batch size assuming it is sufficient large. Therefore, we can conclude our spectral bias results are mostly unaffected by batch size (this is notable since FFTs can possibly encompass a large batch since it is grid-based). We also refer to Figure \ref{fig:kdv_fft_smallvslargegrid} for a similar effect with FFTs.}
\label{fig:kdv_spectral_batch}
\end{figure}

\subsection{Error analysis}

We will primarily assess our experiments with the discrete $L^2$ error analysis using
\begin{align}
& \frac{ | | u(x,t)  - u_{\theta}(x,t) | |_{L^2(\mathcal{X} \times [0,T]) } }{ | | u(x,t) | |_{L^2(\mathcal{X} \times [0,T])}} 
\\
& \approx
\frac{ \sqrt{ \sum_{(x,t) \in  {\overline{\mathcal{X}} \times \{ j \Delta t : j \in \mathbb{N}, (\max j) \Delta t = T \}} } (u(x,t) - u_{\theta}(x,t))^2  } }{ \sqrt{ \sum_{(x,t) \in  {\overline{\mathcal{X}} \times \{ j \Delta t : j \in \mathbb{N}, (\max j) \Delta t = T \}} } (u(x,t) )^2 } } ,
\end{align}
where $u$ is a numerical solution (computed with FFTs typically) $\overline{\mathcal{X}} = \{ a + \langle j, (\Delta x, \hdots, \Delta x) \rangle : a \in \mathbb{R}^n, j \in \mathbb{N}^n \cup \{ 0 \} \}$ is a suitable mesh in space. Thus, we highlight we use a metric of the actual error, not the loss, which is a different metric. It is notable that we do indeed keep the square roots in our experiments.

\subsection{Fourier modes}

A downside of our methods is that since differentiation corresponds to multiplication by polynomials in Fourier space, these polynomials tend to blow up along the boundary of compact $\Omega$, thus the weighing due to the differentiation will favor the boundary. We resolve this issue by truncating the Fourier modes. This is a standard technique, for example used in \cite{li2021fourierneuraloperatorparametric}. Thus we focus on low-dimensionality techniques, and we emphasize low $|\xi|$. Mathematically, this is
\begin{align}
& \chi_{\mathcal{M}}(\xi) = \begin{cases} 1 & |\xi| \leq \Xi^* \\ 0 & |\xi| > \Xi^* \end{cases},  
\\
& \mathcal{L} = \int_{-\infty}^{\infty} \int_{-\infty}^{\infty}  \hdots \int_{-\infty}^{\infty} \Big| W(\xi) \cdot \widehat{R}(\xi) \cdot \chi_{\mathcal{M}}(\xi) \Big|^2 \prod_j d\xi_j
\end{align}
is our residual for value $\mathbb{R} \ni \Xi^* < \infty$. Discretely, this loss is
\begin{gather}
\mathcal{L} = \sum_{\xi_1=0}^{K_1} \sum_{\xi_2=0}^{K_2} \dots \sum_{\xi_n=0}^{K_d} \Big| W_{\xi_1,\hdots,\xi_n}  \cdot \widehat{R}_{\xi_1,\hdots,\xi_n}  \cdot \chi_{\mathcal{M},\xi_1,\hdots,\xi_n} \Big|^2 \in \mathcal{F} \subseteq \mathbb{R}
\\
\mathcal{F} = \Bigg\{ \sum_{\xi \in \mathcal{I}_{\tilde{\Xi}}} | W_{\xi_1,\hdots,\xi_n} \cdot \widehat{R}_{\xi_1,\hdots,\xi_n}  \Bigg|^2 \ \Bigg| \   \mathcal{I}_{\tilde{\Xi}} = \big\{ \xi \in \mathbb{Z}^n | |\xi_i| \leq \tilde{\Xi}_i \big\} \Bigg\} ,
\end{gather}
for suitable maximum modes $K_i$ and truncated modes $\tilde{K}_i$. We crucially remark we do not want to truncate too severely to mitigate effects of spectral bias learning.

\subsection{Grid versus Monte-Carlo training}
\label{sec:grid_vs_montecarlo}

\noindent Our primary means of our enhanced Fourier loss is through the use of FFTs that cohere to a mesh of the form
\begin{align}
P(D) u(x) \approx \sum_{\xi_1=0}^{N_1 - 1} \hdots \sum_{\xi_j =0}^{N_j} \widehat{u}(\xi_1,\hdots,\xi_j) P(\xi_1,\hdots,\xi_j) e^{2 \pi i \sum_k x_k \xi_k / s_k} ,
\end{align}
and to evaluate the enhanced loss over the entire mesh at once. It can be noted this is traditionally how the Fourier neural operator was constructed as in \cite{li2021fourierneuraloperatorparametric} via page 5. This is our preferred our method but it does have drawbacks For example, the traditional PINN possesses the mesh invariant quality due to the random, uniform collocation sampling, and cohering to the grid loses this quality. Moreover, a strength of the gradient-enhanced PINN is that the enhanced residual is evaluated at the same locations. This allows well-behaved backpropagation, whereas we empirically found evaluating at Fourier locations along a mesh and random physics locations to not harmonize but rather conflict in training, and overall this enhanced approach did not yield much improvement.

\vspace{2mm}

\noindent Another approach is to use
\begin{align}
& [P(\xi) u(\xi)](t)  =  \int_{\mathbb{R}^n} e^{-2 \pi i \langle \xi, x \rangle} P(x) u(x,t) \rho(x) dx
\\
& \propto \EX_{x \sim \rho} \Big[ e^{-2 \pi i \langle \xi, x \rangle} P(x) u(x,t) \Big] \propto \sum_{ x_j \sim \mathcal{P}(\Omega), j \in [J]} e^{-2 \pi i \langle \xi, x_j \rangle} P(x_j) u(x_j,t) ,
\end{align}
where $\mathcal{P}(\Omega)$ is a suitable measure on finite domain $\Omega$. In general, this option is less computationally trivial since it requires an integral approximation. This method also has pitfalls. For example, this method requires an integral approximation, which is the best developed in the case when collocation batch size $N_f$ is large; however, this is problematic because the gradient-enhanced PINN \cite{Yu_2022} has historically been proven to function best in the low-data regime. Generally, it is desirable to keep the Fourier modes reasonable in value (<12). High-frequency modes are negligible and overall not well-captured according to \ref{alg:mc_enhanced_loss}; higher modes fall below a floor level where they are indistinguishable from existing noise. Moreover, the weights $P(\xi)$ will be large in the high-frequency cases, so these will effectively multiply noise rather than meaningful values. Essentially, the physics component of the Fourier loss is small noise that is amplified by blowing-up values at high modes with Monte Carlo methods. Therefore, it is better to keep the number of Fourier modes in this setting reasonable.

\subsection{Complements with advanced techniques of PINNs}

In this subsection, we describe archetypal PINN augmentations to strengthen capability that are well-estblished in literature.

\vspace{2mm}

\noindent The PINN enhancement we consider most predominantly is the Fourier feature embedding \cite{tancik2020fourierfeaturesletnetworks} \cite{wang2023expertsguidetrainingphysicsinformed}, where neural network $\psi : \mathbb{R}^{2 m } \times \Theta \rightarrow L^1(\Omega \times [0,T]; \mathbb{R}), \psi_{\theta}$ inputs are mapped $
(x,t) \rightarrow \Big( \sin(  2 \pi B (x,t)^T ), \cos(2 \pi B (x,t)^T ) \Big) \ \ \rightarrow  \ \ \psi_{\theta}  \Big( \sin( 2 \pi B (x,t)^T ), \cos( 2 \pi B (x,t)^T ) \Big) $, where $B \in \mathbb{R}^{m \times (n+1)}$, $B_{ij} \sim \mathcal{N}(0,\sigma^2)$ before being mapped through the layers according to $\psi_{\theta}$. Empirically, we found this technique most crucial for PINN success.

\vspace{2mm}

\noindent We will also consider: (1) the modified MLP architecture of \cite{wang2023expertsguidetrainingphysicsinformed}, which proceeds iteratively as
\begin{align}
h^{i+1} & = (1 - \sigma( W^i h^{i} + b^i ) ) \odot \widehat{h} + \sigma( W^i h^{i} + b^i ) \odot \tilde{h},  
\\
\widehat{h} & = \sigma(\widehat{W} X + \widehat{b}), \tilde{h} = \sigma(\tilde{W} X + \tilde{b}) 
\end{align}

for hidden node $h$; (2) the SOAP optimizer (Shampoo with Adam) \cite{vyas2025soapimprovingstabilizingshampoo}, which is second order, and second order optimizers have demonstrated competitive performance in PINN settings \cite{kiyani2025optimizingoptimizerphysicsinformedneural}; (3) the grad norm coefficient tuning procedure of \cite{wang2023expertsguidetrainingphysicsinformed}. We generally will use $\text{sin}(\cdot)$ activation. Recall $\text{relu}(\cdot)$ activations are often adverse for PINNs since they are not smooth.

\subsection{Experiment details according to PDE}

\begin{table*}[!htbp]
\caption{We list various statistics identifying the learning power of frequency bias corresponding to the data as in Algorithm \ref{alg:radial_power_alg} pertaining to radial power spectral density. All of our results are done with severe early stopping and high $\mathcal{L}_{\text{Fourier enhanced}}$ coefficient to overemphasize the spectral bias learning effects. The frequency \% refers to a percentile (high is better since high frequencies learned is desirable). The ratio quantifies the amount of frequency that exists in that quantile range (high ratio in high frequency is desirable).}
\label{tab:method_errors}
\centering
\renewcommand{\arraystretch}{1.2}%
\scriptsize
\begin{tabular}{
  >{\raggedright\arraybackslash}p{2.5cm}
  >{\centering\arraybackslash}p{1.5cm}
  >{\centering\arraybackslash}p{1.5cm}
  >{\centering\arraybackslash}p{1.5cm}
  >{\centering\arraybackslash}p{1.5cm}
  >{\centering\arraybackslash}p{1.5cm}
  >{\centering\arraybackslash}p{1.5cm}
  >{\centering\arraybackslash}p{1.5cm}
}
\toprule
\multirow{2}{*}{Scenario} & \multicolumn{7}{c}{Method [$\downarrow$]} \\
\cmidrule(lr){2-8}
& Total log error power [$\downarrow$] & Average log error power [$\downarrow$] & Frequency 50 [$\uparrow$] & Frequency 90 [$\uparrow$] & Ratio 0.0-0.1 [$\downarrow$] & Ratio 0.1-0.25 [$\uparrow$] & Ratio 0.25-0.5 [$\uparrow$] \\
\midrule
\rowcolor{Apricot!25} Allen-Cahn FE & ${1.812}\mathrm{e}{+1}$ & ${1.201}\mathrm{e}{0}$ & ${0.000}\mathrm{e}{0}$ & ${7.092}\mathrm{e}{-3}$ & ${9.999}\mathrm{e}{-1}$ & ${8.789}\mathrm{e}{-5}$ & ${1.094}\mathrm{e}{-5}$ \\
\addlinespace
\rowcolor{CornflowerBlue!25} Allen-Cahn vanilla & ${1.903}\mathrm{e}{+1}$ & ${1.384}\mathrm{e}{0}$ & ${0.000}\mathrm{e}{0}$ & ${3.546}\mathrm{e}{-3}$ & ${9.999}\mathrm{e}{-1}$ & ${6.346}\mathrm{e}{-5}$ & ${8.788}\mathrm{e}{-6}$ \\
\addlinespace
\rowcolor{Apricot!25} Burger's FE & ${1.557}\mathrm{e}{+1}$ & ${5.843}\mathrm{e}{-1}$ & ${7.092}\mathrm{e}{-3}$ & ${1.064}\mathrm{e}{-2}$ & ${9.999}\mathrm{e}{-1}$ & ${1.051}\mathrm{e}{-4}$ & ${1.696}\mathrm{e}{-5}$\\
\addlinespace
\rowcolor{CornflowerBlue!25} Burger's vanilla & ${1.596}\mathrm{e}{+1}$& ${6.753}\mathrm{e}{-1}$ & ${0.000}\mathrm{e}{0}$ & ${1.064}\mathrm{e}{-2}$ & ${9.999}\mathrm{e}{-1}$ & ${8.644}\mathrm{e}{-5}$ & ${1.404}\mathrm{e}{-5}$ \\
\addlinespace
\rowcolor{Apricot!25} KdV FE & ${1.489}\mathrm{e}{+1}$& ${8.737}\mathrm{e}{-1}$ & ${2.837}\mathrm{e}{-2}$ & ${3.901}\mathrm{e}{-2}$ & ${9.989}\mathrm{e}{-1}$ & ${9.842}\mathrm{e}{-4}$ & ${1.423}\mathrm{e}{-4}$  \\
\addlinespace
\rowcolor{CornflowerBlue!25} KdV vanilla & ${1.530}\mathrm{e}{+1}$& ${7.861}\mathrm{e}{-1}$ & ${1.064}\mathrm{e}{-2}$ & ${3.546}\mathrm{e}{-2}$ & ${9.991}\mathrm{e}{-1}$ & ${7.631}\mathrm{e}{-4}$ & ${1.116}\mathrm{e}{-4}$  \\
\addlinespace
\rowcolor{Apricot!25} Navier-Stokes FE & ${1.962}\mathrm{e}{+1}$& ${2.986}\mathrm{e}{0}$ & ${5.525}\mathrm{e}{-3}$ & ${1.657}\mathrm{e}{-2}$ & ${9.991}\mathrm{e}{-1}$ & ${7.881}\mathrm{e}{-4}$ & ${1.217}\mathrm{e}{-4}$ \\
\addlinespace
\rowcolor{CornflowerBlue!25} Navier-Stokes vanilla & ${2.006}\mathrm{e}{+1}$& ${3.039}\mathrm{e}{0}$ & ${5.525}\mathrm{e}{-3}$ & ${1.381}\mathrm{e}{-2}$ & ${9.994}\mathrm{e}{-1}$ & ${5.080}\mathrm{e}{-4}$ & ${7.880}\mathrm{e}{-5}$ \\
\addlinespace
\bottomrule
\end{tabular}
\end{table*}

\noindent In this subsection, we discuss the experimental details we find most relevant.

\begin{figure}[htbp]
  \vspace{0mm}
  \centering
  \includegraphics[scale=0.255]{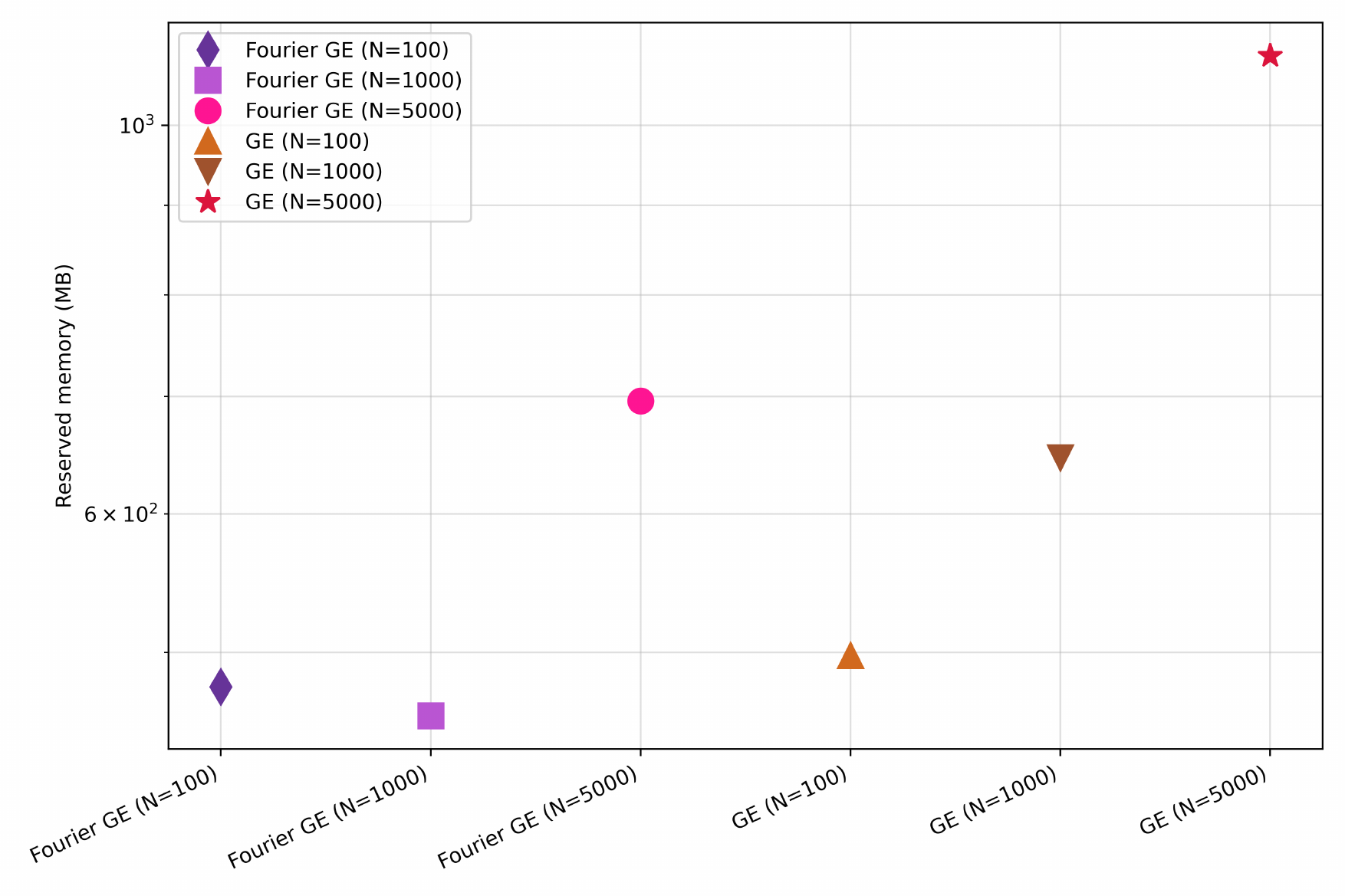}
  \caption{We plot reserved memory usage on a Fourier gradient enhanced versus a traditional gradient enhanced PINN on Burger's equation with a triangular domain at a single iteration (it is about constant per epoch).}
  \label{fig:reserved_memory_figure_burges}
\end{figure}

\vspace{2mm}

\noindent \textbf{Burger's equation.} In these experiments, we consider Burger's equation on both a square and triangular domain. The triangular domain case is interesting because it nullifies our typical grid based FFT application and instead uses a Monte Carlo technique (see Subsection \ref{sec:grid_vs_montecarlo}). Based on Figure \ref{fig:burgers_nonsquare_images}, we can see our methods provide a superior error quality more uniformly, but can fail on the classic rift of the Burger's equation potentially more than using a physics loss only. Moreover, we demonstrate the frequency bias effects of our methods in Figure \ref{fig:kdv_error_power}.

\vspace{2mm}

\noindent \textbf{Allen-Cahn equation.} In these experiments, we use the FFT we intensify the capabilities of our methods by diminishing the collocation point number to values. We demonstrate rigorously in Figures \ref{fig:allen_cahn_samples_sidebyside}, \ref{fig:allencahn_batched} that our methods can achieve loss plateau drop-off due to the Fourier enhanced loss, which has minimal contribution to the loss gradients (thus the breakthroughs are not results of greater physics-type gradient contributions). A potential reason for this is because the FFT obtains greater levels of contribution across many points. Our results are still notable for reasons such as a drop-off effect, and the effect is not gradual, which is anticipated due to a higher collocation quantity.

\vspace{2mm}

\noindent \textbf{Korteweg-De Vries (KdV) equation.} Here, we attempt to demonstrate the frequency bias learning effects of our method more rigorously, since solutions to the KdV equation exhibit high oscillation and can more suitably characterize spectral bias effects of neural network lerarning processes. In Figure \ref{fig:kdv_images_4samples}, we demonstrate our methods over four training samples after severe scenarios of early stopping. We remark it is unclear if this exactly demonstrates frequency bias learning effects, so we remark it is potential. We refer to Figure \ref{fig:kdv_error_power} for more clear evidence of spectral bias in these experiments.

\vspace{2mm}

\noindent \textbf{Navier-Stokes.} The Navier-Stokes experiment is interesting because it represents a more severe case of spectral bias along with the KdV equation. We found visually our results did not affect learning so significantly; however, the effects of our methods are observable in a frequency analysis (see Figure \ref{fig:error_power_navier_stokes}, Table \ref{tab:method_errors}). We found loss converged rather quickly in this experiment. We refer to Figure \ref{fig:navier_stokes_pinn_example} for the example used in experiments.

\vspace{2mm}

\textbf{Effects of batch size versus our methods.} In order to properly determine our methods affect spectral bias rather than relying on a higher number of collocation points to affect this, we implement general strategies to mitigate spectral bias due to high Fourier enhanced collocation: (1) for Monte Carlo techniques, we collocate on the same locations as the physics loss, for example in Figures \ref{fig:burgers_nonsquare_images} \ref{fig:burgers_nonsquare_3samples}, \ref{fig:burgers_nonsquare_error_power}; (2) for experiments with use of the FFT, we take a large batch size, generally 250-400 for the Allen-Cahn and KdV equations and 2,000-3,000 for Navier-Stokes, also corresponding to Table \ref{tab:method_errors}. We often take FFT grid sizes between 151-301, but we also consider smaller grids of size 31-61 in Figures \ref{fig:kdv_spectral_training_fftsmall}, \ref{fig:kdv_fft_smallvslargegrid}. These remarks are also notable because gradient enhanced PINNs are often for low batch size (<50).

\section{Conclusions and limitations}
\label{sec:conclusions}

We developed gradient enhancement in Fourier space. Our approaches were highly theoretical but also supported with experiments. Our primary results are likely our neural tangent kernel (NTK) results, thereby showing differentiation affects spectral bias. Our proofs were done in Fourier space, therefore a natural question arises whether standard gradient enhancement affects spectral bias. Since spectral bias is affected by dynamical effects \cite{khodakarami2026spectralbiasphysicsinformedoperator}, we conjecture it is so. The primary pitfall of our methods is that gradient enhancement is best classically for the low-data regime \cite{Yu_2022}, and while the Monte Carlo technique can help accommodate this, Monte Carlo methods are plagued by high variance, especially in low-collocation settings. We also discussed Monte Carlo limitations in \ref{sec:grid_vs_montecarlo}.

\section{Acknowledgements}

I gratefully acknowledge financial support from Purdue University Department of Mathematics. I would also like to thank Monica Torres and Antônio Sá Barreto at Purdue University for their valuable lectures relevant to this work.

\bibliographystyle{plainnat}
\bibliography{bibliography}

\appendix

\onecolumn

\newpage

\section{On more generalized linear PDE differential operators}
\label{sec:generalized_linear_PDEs}

We previously established the loss
\begin{align}
\mathbb{E}_{u \sim \delta_{\mu}} \Bigg[ \Bigg| \Bigg|  P(\xi)   \mathcal{F}( \dot{u} ) + \sum_{\beta}  P(\xi) (2 \pi i \xi)^{\beta}  \mathcal{F}(u) \Bigg| \Bigg|_{L^2(\Omega \times [0,T])}^2 \Bigg] .
\end{align}
The contribution from $(2 \pi i \xi)^{\beta}$ assumes a naïve PDE differential operator. Let us elaborate more for more sophisticated \textit{linear} PDE differential operators. We remark nonlinear PDEs require more of a case-by-case analysis, which we discuss in Appendices \ref{app:burger's}, \ref{app:allen-cahn}, \ref{app:kdv}, \ref{app:navier-stokes}.

\vspace{2mm}

\noindent Let us consider a linear PDE of the form
\begin{align}
\dot{u}(x,t) + \sum_{\alpha} \partial^{\alpha_0} \Big( a_0(x,t) \partial^{\alpha_1} \Big( a_1(x,t) \hdots \partial^{\alpha_d} \Big( a_d(x,t) u(x,t) \Big) \hdots \Big) \Big) = 0.
\end{align}
Now, the generalized frequency-domain representation is
\begin{align}
\partial_t \widehat{u}(\xi, t) + \sum_{\alpha} (2\pi i \xi)^{\alpha_0} \Big( \widehat{a}_0 * \Big( (2\pi i \xi)^{\alpha_1} \Big( \widehat{a}_1 * \dots * \Big( (2\pi i \xi)^{\alpha_d} (\widehat{a}_d * \widehat{u}) \Big) \dots \Big) \Big) \Big) = 0
\end{align}
by the convolution theorem. Therefore, for more generalized linear PDEs, the loss is
\begin{align}
\mathbb{E}_{u \sim \delta_{\mu}} \Bigg[ \Bigg| \Bigg|  P(\xi) \cdot \Bigg[   \mathcal{F}( \dot{u} ) + \sum_{\alpha} (2\pi i \xi)^{\alpha_0} \Big( \widehat{a}_0 * \Big( (2\pi i \xi)^{\alpha_1} \Big( \widehat{a}_1 * \dots * \Big( (2\pi i \xi)^{\alpha_d} (\widehat{a}_d * \widehat{u}) \Big) \dots \Big) \Big) \Big)  \Bigg] \Bigg| \Bigg|_{L^2(\Omega \times [0,T])}^2 \Bigg] .    
\end{align}

\section{Fourier integral equivalence with the Plancherel Theorem}
\label{app:plancherel_sec}

Observe our Fourier loss of equation \ref{eqn:fourier_physics_loss} has further equivalence to a traditional physics loss via the Plancherel theorem \cite{raban2020fourier}. Recall the Plancherel Theorem states
\begin{align}
\int_{\mathbb{R}^n} |f(x)|^2 dx = \int_{\mathbb{R}^n } |\widehat{f(\xi)}|^2 d\xi .
\end{align}
Observe
\begin{align}
\label{eqn:plancherel}
& \EX_{u \sim \delta_{\nu}} \Big[ \ \Big| \Big| P(\xi) \mathcal{F}( \dot{u} ) + \sum_{\beta} P(\xi) (2 \pi i \xi)^{\beta} \mathcal{F}(u) \Big| \Big|_{L^2(\Omega \times [0,T])}^2 \ \Big]
\\
& \stackrel{P(\xi)\mathcal{F}(u) = \mathcal{F}(P(D)u)}{\cong} \EX_{u \sim \delta_{\nu}} \Big[ \ \Big| \Big| 
\ \mathcal{F} \Big[ P(D)  \dot{u}  + \sum_{\beta} P(D) D_x^{\beta} u \Big] \ \Big| \Big|_{L^2(\mathcal{X} \times [0,T])}^2 \ \Big]
\\
& \stackrel{\text{Plancherel Theorem}}{\cong} \EX_{u \sim \delta_{\nu}} \Big[ \ \Big| \Big| 
\  P(D)  \dot{u}  + \sum_{\beta} P(D) D_x^{\beta} u  \ \Big| \Big|_{L^2(\mathcal{X} \times [0,T])}^2 \ \Big] .
\end{align}
Note that the Plancherel Theorem is typically over unbounded Euclidean space, and our PINNs are developed over compact truncations, so we note the proportionality and non-exactness. We will develop upon this momentarily using characteristic functions.

\vspace{2mm}

\textit{\textbf{Theorem 3.} Let $u_{\theta} \in C^1(\mathcal{X} \times [0,T]; \mathbb{R}) \cap W^{p,1}(\mathcal{X} \times [0,T]; \mathbb{R})$ be a parameterized function optimized to solve a partial differential equation. Assume the training objectives are evaluated over compact, truncated spatial and frequency domains. While the physical-space enhanced loss and the Fourier enhanced loss are equivalent over unbounded domains via Plancherel's Theorem, gradient descent on these two truncated objectives follows different optimization trajectories. }

\vspace{2mm}

\noindent \textit{Proof.} Observe the norms via \ref{eqn:plancherel} are the same but they do not train the same objective. In order to apply Lebesgue dominated convergence on the difference quotient, we notice there exists a $\mathcal{G}$ such that
\begin{align}
& \left| \nabla_{\theta} \left| P(\xi) \mathcal{F}( \dot{u}_{\theta} ) + \sum_{\beta} P(\xi) (2 \pi i \xi)^{\beta} \mathcal{F}(u_{\theta}) \right|^2 \right| 
\\
& = \lim_{h \rightarrow 0} \left|  \frac{ \left| P(\xi) \mathcal{F}( \dot{u}_{\theta+h \theta'} ) + \sum_{\beta} P(\xi) (2 \pi i \xi)^{\beta} \mathcal{F}(u_{\theta+h \theta'}) \right|^2 - \left| P(\xi) \mathcal{F}( \dot{u}_{\theta} ) + \sum_{\beta} P(\xi) (2 \pi i \xi)^{\beta} \mathcal{F}(u_{\theta}) \right|^2 }{h} \right|
\\
& \ \ \ \ \ \ \ \ \ \ \ \ \ \ \ \ \ \ \ \ \leq \mathcal{G}(\xi, t),  \ \ \ \iint_{[0,T] \times \Omega} \mathcal{G}(\xi, t) dt d\xi < \infty .
\end{align}
We remark $\mathcal{G}$ is real-valued since the gradient on the norm is taken (but the interior of the norm is complex). In particular, we will endow smoothness, and our domains are compact (and finite measure), so $\mathcal{G}$ exists under the supremum over the spaces (the parameter space $\Theta$ is also open; we can generally allow $\Theta$ to be enclosed in a compact set but note this assumption has various levels of reasonableness, discussed in levels in works such as \cite{nagarajan2021uniformconvergenceunableexplain} \cite{poggio2019theoreticalissuesdeepnetworks}, and we will go ahead and assume sufficient conditions for regularity; also as a remark, the exchange is allowed by the Leibniz integral rule, which is proven with dominated convergence). Note that a similar argument follows by the mean value theorem. Also, a similar argument holds for regularity under the Fourier transform integral, as we will assume $u_{\theta} \in C_0^{\infty}(\mathcal{X} \times [0,T]; \mathbb{R})$ due to the activation, and $\text{supp}(\partial^{\alpha} u) \subseteq \text{supp}(u)$. Recall the following identity \cite{Lee_2023} \cite{Li2008}
\begin{align}
\nabla_{\theta} |f(\theta)|^2 = 2 \text{Re} \Big( \overline{f(\theta)} \nabla_{\theta } f \Big) .
\end{align}
We will use the following notation:
\begin{align}
\int \nabla_{\theta} \Psi = \Big( \int \nabla_{\theta_1} \Psi , \hdots, \int \nabla_{\theta_D} \Psi )^T ,
\end{align}
so the integration is element-wise upon the gradients. Under regularity for the following, for gradients on the domain-bounded training objective:
\begin{align}
& \nabla_{\theta} \int_{[0,T]} \int_{\Omega}  | P(\xi) \mathcal{F}( \dot{u}_{\theta} ) + \sum_{\beta} P(\xi) (2 \pi i \xi)^{\beta} \mathcal{F}(u_{\theta})  |^2 d\xi dt
\\
& =  \int_{[0,T]} \int_{\Omega}  \nabla_{\theta} | P(\xi) \mathcal{F}( \dot{u}_{\theta} ) + \sum_{\beta} P(\xi) (2 \pi i \xi)^{\beta} \mathcal{F}(u_{\theta})  |^2 d\xi dt
\\
& =  \int_{[0,T]} \int_{\Omega}  2 \text{Re} \Bigg( \overline{ ( P(\xi) \mathcal{F}( \dot{u}_{\theta} ) + \sum_{\beta} P(\xi) (2 \pi i \xi)^{\beta} \mathcal{F}(u_{\theta})  )  }
\\
& \ \ \ \ \ \ \ \ \ \ \ \times \Bigg\{ P(\xi) \nabla_{\theta}  \int_{\mathbb{R}^n} e^{-2 \pi i \langle x, \xi \rangle} \dot{u}_{\theta}(x) dx + \sum_{\beta} P(\xi) (2 \pi i \xi)^{\beta} \nabla_{\theta} \int_{\mathbb{R}^n} e^{-2 \pi i \langle x, \xi \rangle} u_{\theta}(x) dx  \Bigg\} \Bigg) d\xi dt
\\
& = \int_{[0,T]} \int_{\Omega}  2 \text{Re} \Bigg( \overline{ ( P(\xi) \mathcal{F}( \dot{u}_{\theta} ) + \sum_{\beta} P(\xi) (2 \pi i \xi)^{\beta} \mathcal{F}(u_{\theta})  )  }
\\
& \ \ \ \ \ \ \ \ \ \ \ \times \Bigg\{ P(\xi) \int_{\mathbb{R}^n} e^{-2 \pi i \langle x, \xi \rangle} \nabla_{\theta} \dot{u}_{\theta}(x) dx + \sum_{\beta} P(\xi) (2 \pi i \xi)^{\beta} \int_{\mathbb{R}^n} e^{-2 \pi i \langle x, \xi \rangle} \nabla_{\theta}  u_{\theta}(x) dx  \Bigg\} \Bigg) d\xi dt .
\end{align}
The last line simply follows by compact support and smoothness. We remark it of interest to apply the Riemann-Lebesgue lemma in the last line, but that does not guarantee integrability. In physical space,
\begin{align}
&\nabla_{\theta} \int_{[0,T]} \int_{\mathcal{X}}  ( P(D) \dot{u}_{\theta} + \sum_{\beta} P(D) D_x^{\beta} u_{\theta} )^2 dt dx
\\
\label{eqn:loss_physical}
& = \int_{[0,T]} \int_{\mathcal{X}} 2 ( P(D) \dot{u}_{\theta} + \sum_{\beta} P(D) D_x^{\beta} u_{\theta} ) \Big\{ P(D) \nabla_{\theta} \dot{u}_{\theta} + \sum_{\beta} P(D) D_x^{\beta} \nabla_{\theta} u_{\theta} \Big\}  dx dt .
\end{align}
Note that Plancherel's Theorem also states
\begin{align}
\int_{-\infty}^{\infty} \hdots \int_{-\infty}^{\infty} f(x) \overline{g(x)} dx = \int_{-\infty}^{\infty} \hdots \int_{-\infty}^{\infty} \widehat{f}(\xi) \overline{\widehat{g}(\xi)} d\xi .
\end{align}
Thus, we have
\begin{align}
& \int_{[0,T]} \int_{\mathbb{R}^n} 2 ( P(\xi) \mathcal{F}( \dot{u}_{\theta} ) + \sum_{\beta} P(\xi) (2 \pi i \xi)^{\beta} \mathcal{F}(u_{\theta})  ) 
\\
& \ \ \ \ \ \ \ \ \ \ \ \times \overline{ \Bigg\{ P(\xi) \int_{\mathbb{R}^n} e^{-2 \pi i \langle x, \xi \rangle} \nabla_{\theta} \dot{u}_{\theta}(x) dx + \sum_{\beta} P(\xi) (2 \pi i \xi)^{\beta} \int_{\mathbb{R}^n} e^{-2 \pi i \langle x, \xi \rangle} \nabla_{\theta}  u_{\theta}(x) dx  \Bigg\} } d\xi dt
\\
& = \int_{[0,T]} \int_{\mathbb{R}^n} 2 ( P(D) \dot{u}_{\theta} + \sum_{\beta} P(D) D_x^{\beta} u_{\theta} ) \overline{ \Big\{ P(D) \nabla_{\theta} \dot{u}_{\theta} + \sum_{\beta} P(D) D_x^{\beta} \nabla_{\theta} u_{\theta} \Big\}  }  dx dt
\\
& = \int_{[0,T]} \int_{\mathbb{R}^n} 2 ( P(D) \dot{u}_{\theta} + \sum_{\beta} P(D) D_x^{\beta} u_{\theta} ) \Big\{ P(D) \nabla_{\theta} \dot{u}_{\theta} + \sum_{\beta} P(D) D_x^{\beta} \nabla_{\theta} u_{\theta} \Big\}    dx dt .
\end{align}
Here, both $f,g$ are real-valued. Thus, the gradients are not the same, and solve a different objective. Here, we have noted
\begin{align}
& \mathcal{F}(P(D)\dot{u}) = P(\xi)\mathcal{F}(\dot{u})
\\
& \mathcal{F}(P(D)D_x^\beta u) = P(\xi)\mathcal{F}(D_x^\beta u) = P(\xi)(2 \pi i \xi)^\beta \mathcal{F}(u) .
\end{align}
Notice we have truncated $\mathbb{R}^n$ to compact sets. Let us attempt to apply the Plancherel Theorem more rigorously restricted to our compact sets. For example, returning to the loss on physical space, we can relate this to Fourier space with
\begin{align}
& \int_{[0,T]} \int_{\mathcal{X}} 2 ( P(D) \dot{u}_{\theta} + \sum_{\beta} P(D) D_x^{\beta} u_{\theta} ) \Big\{ P(D) \nabla_{\theta} \dot{u}_{\theta} + \sum_{\beta} P(D) D_x^{\beta} \nabla_{\theta} u_{\theta} \Big\}    dx dt
\\
& = \int_{[0,T]} \int_{\mathbb{R}^n}  2 ( P(D) \dot{u}_{\theta} + \sum_{\beta} P(D) D_x^{\beta} u_{\theta} ) \overline{ \Big\{ P(D) \nabla_{\theta} \dot{u}_{\theta} + \sum_{\beta} P(D) D_x^{\beta} \nabla_{\theta} u_{\theta} \Big\}     \chi_{\mathcal{X}}(x) }  dx dt
\\
& \stackrel{\text{Plancherel Theorem}}{=} \int_{[0,T]} \int_{\mathbb{R}^n} 2 
\mathcal{F} \Big[ ( P(D) \dot{u}_{\theta} + \sum_{\beta} P(D) D_x^{\beta} u_{\theta} ) \Big] 
\overline{ \mathcal{F} \Big[ \Big\{ P(D) \nabla_{\theta} \dot{u}_{\theta} + \sum_{\beta} P(D) D_x^{\beta} \nabla_{\theta} u_{\theta} \Big\}    \chi_{\mathcal{X}}(x) \Big] }   dx dt
\\
\label{eqn:physical_to_fourier}
& \stackrel{\text{Convolution Theorem}}{=} \int_{[0,T]} \int_{\mathbb{R}^n} 2 ( P(\xi) \mathcal{F}( \dot{u}_{\theta} ) + \sum_{\beta} P(\xi) (2 \pi i \xi)^{\beta} \mathcal{F}(u_{\theta})  ) 
\\
& \ \ \ \ \ \ \ \ \ \ \ \times \overline{ \Bigg\{ P(\xi) \int_{\mathbb{R}^n} e^{-2 \pi i \langle x, \xi \rangle} \nabla_{\theta} \dot{u}_{\theta}(x) dx + \sum_{\beta} P(\xi) (2 \pi i \xi)^{\beta} \int_{\mathbb{R}^n} e^{-2 \pi i \langle x, \xi \rangle} \nabla_{\theta}  u_{\theta}(x) dx  \Bigg\} * \widehat{\chi_{\mathcal{X}} } } d\xi dt ,
\end{align}
and the convolution truncates the domain in a decaying fashion but not compactly. It can be noted \cite{Martin2012}
\begin{align}
\widehat{\chi_{\mathcal{X}}}(\xi) = \prod_{j=1}^n 2R \frac{\sin(2 \pi R \xi_j)}{2 \pi R \xi_j} = \prod_{j=1}^n 2R \text{sinc}(2 R \xi_j) 
\end{align}
over a box $\mathcal{X} = [-R, R]^n$. Also note that the Plancherel Theorem defines an isometry. Note that dominated convergence also holds for $\mathbb{C}$-valued measurable functions. For the Fourier loss, we can relate it to physical space with
\begin{align}
\label{eqn:fourier_grads}
& \int_{[0,T]} \int_{\Omega}  2 \text{Re} \Bigg( \overline{ ( P(\xi) \mathcal{F}( \dot{u}_{\theta} ) + \sum_{\beta} P(\xi) (2 \pi i \xi)^{\beta} \mathcal{F}(u_{\theta})  )  }
\\
& \ \ \ \ \ \times \Bigg\{ P(\xi) \int_{\mathbb{R}^n} e^{-2 \pi i \langle x, \xi \rangle} \nabla_{\theta} \dot{u}_{\theta}(x) dx + \sum_{\beta} P(\xi) (2 \pi i \xi)^{\beta} \int_{\mathbb{R}^n} e^{-2 \pi i \langle x, \xi \rangle} \nabla_{\theta}  u_{\theta}(x) dx  \Bigg\} \Bigg) d\xi dt 
\\[1em]
& = 2 \text{Re} \int_{[0,T]} \int_{\mathbb{R}^n}  \overline{ ( P(\xi) \mathcal{F}( \dot{u}_{\theta} ) + \sum_{\beta} P(\xi) (2 \pi i \xi)^{\beta} \mathcal{F}(u_{\theta})  )   \times \chi_{\Omega}(\xi) }
\\
& \ \ \ \ \  \times \Bigg\{ P(\xi) \int_{\mathbb{R}^n} e^{-2 \pi i \langle x, \xi \rangle} \nabla_{\theta} \dot{u}_{\theta}(x) dx + \sum_{\beta} P(\xi) (2 \pi i \xi)^{\beta} \int_{\mathbb{R}^n} e^{-2 \pi i \langle x, \xi \rangle} \nabla_{\theta}  u_{\theta}(x) dx  \Bigg\}   d\xi dt
\\[1em]
& \stackrel{\text{Plancherel Theorem}}{=} 2 \text{Re} \int_{[0,T]} \int_{\mathbb{R}^n}  \overline{ \mathcal{F}^{-1} \Big[ ( P(\xi) \mathcal{F}( \dot{u}_{\theta} ) + \sum_{\beta} P(\xi) (2 \pi i \xi)^{\beta} \mathcal{F}(u_{\theta})  )    \chi_{\Omega}(\xi)  \Big] } 
\\
& \ \ \ \ \ \ \ \ \ \ \ \ \ \ \  \ \ \ \ \ \ \ \ \ \ \ \ \ \ \ \ \ \ \ \ \ \ \ \ \ \times  \Big\{ P(D) \nabla_{\theta} \dot{u}_{\theta} + \sum_{\beta} P(D) D_x^{\beta} \nabla_{\theta} u_{\theta} \Big\}     dx dt
\\[1em]
& \stackrel{\text{Convolution Theorem}}{=} 2 \text{Re} \int_{[0,T]} \int_{\mathbb{R}^n}   \mathcal{F}^{-1} \Big[ ( P(\xi) \mathcal{F}( \dot{u}_{\theta} ) + \sum_{\beta} P(\xi) (2 \pi i \xi)^{\beta} \mathcal{F}(u_{\theta})  ) \Big] *     
\mathcal{F}^{-1} \Big[ \chi_{\Omega}(\xi)   \Big] 
\\
& \ \ \ \ \ \ \ \ \ \ \ \ \ \ \ \ \ \ \ \ \ \ \ \ \ \ \ \ \ \ \ \ \ \ \ \ \ \ \ \ \times  \Big\{ P(D) \nabla_{\theta} \dot{u}_{\theta} + \sum_{\beta} P(D) D_x^{\beta} \nabla_{\theta} u_{\theta} \Big\}     dx dt
\\[1em]
\label{eqn:fourier_plancherel}
& = 2  \int_{[0,T]} \int_{\mathbb{R}^n}  \Bigg\{ ( P(D) \dot{u}_{\theta} + \sum_{\beta} P(D) D_x^{\beta} u_{\theta} ) *   
\mathcal{F}^{-1} \Big[ \chi_{\Omega}(\xi)   \Big] \Bigg\}    \Bigg\{ P(D) \nabla_{\theta} \dot{u}_{\theta} + \sum_{\beta} P(D) D_x^{\beta} \nabla_{\theta} u_{\theta} \Bigg\}     dx dt ,
\end{align}
where the last line also follows under the assumption the PDE is real-valued (it is possible $\mathcal{F}^{-1}[\chi_{\Omega}]$ is complex-valued). Note that the complex conjugate can be dropped because of this. As before,
\begin{align}
\mathcal{F}^{-1}[\chi_\Omega](x) = \prod_{j=1}^n \frac{\sin(2\pi R x_j)}{\pi x_j} = (2R)^n \prod_{j=1}^n \text{sinc}(2 R x_j) 
\end{align}
on $\Omega = [-R, R]^n$. Observe this function has decay due to the linear function in the denominator.

\vspace{2mm}

\noindent What we just showed was that the gradients on the PDE losses in both physical and Fourier space have truncated representations to compact sets, and we can successfully apply Plancherel's Theorem after applying characteristic functions $\chi$. We have grouped $\chi$ into existing functions in the PDE residuals using a basic product between functions. Even moreso, we can acquire representations of these losses in the alternative space between physical-Fourier space. The reason this connection is meaningful is because as we showed in the beginning of the section, we can relate physical to Fourier losses and their enhanced representations using the Plancherel Theorem, having exactness when our losses are restricted to $\mathbb{R}^n$. Thus, the two gradient paths follow different trajectories because the correspondence is not exact. Notably \ref{eqn:fourier_plancherel} does not match \ref{eqn:loss_physical} and \ref{eqn:fourier_grads} does not match \ref{eqn:physical_to_fourier}. In all of our arguments, we have assumed Lebesgue dominated convergence applies, which is mostly a reasonable assumption, especially restricted to compact sets (we are also assuming smoothness).

$ \square $

\section{Convergence rates of high frequencies with Fourier gradient-enhancement}
\label{app:spectral_decay}

In this section, we discuss spectral bias of neural networks via the neural tangent kernel (NTK). We refer to \cite{gan2025neuraltangentkernelneural} \cite{li2024eigenvaluedecayratesclass} for relevant literature.

\vspace{2mm}

\noindent \textbf{Theorem 2.} Let $u \in C^1(\mathcal{X} \times [0,T]; \mathbb{R}) \cap W^{p,1}(\mathcal{X} \times [0,T]; \mathbb{R})$ be a solution to a PDE of the form $\dot{u} + \Gamma[u, \hdots]=0$. Let $\omega(\xi)$ be a polynomial weight dependent on wavenumber in Fourier space. Suppose the domain is periodic. Assume sufficient regularity conditions are met. Suppose $\text{tanh}(\cdot)$ is used. Then, with the Fourier enhanced loss $\mathcal{L}_{\text{Fourier enhanced}}$ and under gradient descent on parameter $\theta$, the infinite-width limit NTK eigenvalue decay becomes $
\tilde{\lambda}(\xi) \sim |\xi|^{s} F(\xi)$ for some $F(\xi)$, where $s > 0$.

\vspace{2mm}

\noindent \textit{Proof.} Let us consider the loss we have established
\begin{align}
\mathcal{L}(\theta) = \sum_{\xi} \omega(\xi) | \widehat{R}_{\theta}(\xi)|^2 ,
\end{align}
where $\omega(\xi)$ is the Fourier symbol weight and $\widehat{R}$ is the PDE residual risk in Fourier space. Assuming the parameter follows a continuous version of gradient descent,
\begin{align}
\dot{\theta} = - \nabla_{\theta} \mathcal{L}(\theta).
\end{align}
Differentiating the Fourier PDE risk w.r.t time,
\begin{align}
\label{eqn:grad_descent_risk}
\frac{d \widehat{R}_{\theta}(\xi)}{dt} = \langle \nabla_{\theta} \widehat{R}_{\theta}(\xi), \dot{\theta} \rangle .
\end{align}
By the chain rule on the loss, observe
\begin{align}
\nabla_{\theta} \mathcal{L}(\theta) & \propto \sum_{\xi} \omega(\xi) \text{Re} \Big( \overline{ \widehat{R}_\theta(\xi)} \nabla_{\theta} \widehat{R}_\theta(\xi) \Big) 
\\
& \propto \sum_{\xi} \omega(\xi) \left[ \overline{\widehat{R}_\theta(\xi)} \nabla_{\theta} \widehat{R}_\theta(\xi) + \widehat{R}_\theta(\xi) \overline{\nabla_{\theta} \widehat{R}_\theta(\xi)} \right]
\end{align}
(the first is the same identity as discussed in \ref{app:plancherel_sec}, \cite{Li2008}). Substituting this into \ref{eqn:fourier_physics_loss},
\begin{align}
\label{eqn:fourier_risk_gradient_flow}
\frac{d \widehat{R}_{\theta}(\xi)}{dt} &= \sum_k \frac{\partial \widehat{R}_\theta(\xi)}{\partial \theta_k} \dot{\theta}_k = - \sum_k \frac{\partial \widehat{R}_\theta(\xi)}{\partial \theta_k} \frac{\partial \mathcal{L}(\theta)}{\partial \theta_k} \nonumber \\
&= - \sum_k \frac{\partial \widehat{R}_\theta(\xi)}{\partial \theta_k} \sum_{\xi'} \omega(\xi') \left[ \widehat{R}_\theta(\xi') \overline{\frac{\partial \widehat{R}_\theta(\xi')}{\partial \theta_k}} + \overline{\widehat{R}_\theta(\xi')} \frac{\partial \widehat{R}_\theta(\xi')}{\partial \theta_k} \right] \nonumber \\
&= - \sum_{\xi'} \omega(\xi') \left[\widehat{R}_\theta(\xi') \sum_k \frac{\partial \widehat{R}_\theta(\xi)}{\partial \theta_k} \overline{\frac{\partial \widehat{R}_\theta(\xi')}{\partial \theta_k}} + \overline{\widehat{R}_\theta(\xi')} \sum_k \frac{\partial \widehat{R}_\theta(\xi)}{\partial \theta_k} \frac{\partial \widehat{R}_\theta(\xi')}{\partial \theta_k} \right] .
\end{align}
Now, let 
\begin{align}
K_{FE}(\xi,\xi') = \langle \nabla_{\theta} \widehat{R}_{\theta}(\xi), \nabla_{\theta } \widehat{R}_{\theta}(\xi') \rangle 
\end{align}
(we have denoted FE as Fourier-enhanced). The NTK for PINNs \cite{zhou2024neuraltangentkernelpinns}  \cite{wang2020pinnsfailtrainneural} can be defined as
\begin{align}
&\begin{bmatrix} 
\dot{R}_{\theta}[u](x_f) \\ \dot{u}_{\theta}(x_b) \end{bmatrix} = - \begin{bmatrix} K_{RR} & K_{Rb} \\ K_{bR} & K_{bb} \end{bmatrix} \begin{bmatrix} R[u]_{\theta}(x^f) \\ u_{\theta}(x^b) - g(x^b) \end{bmatrix}
\\
& K_{RR, ij} = \Bigg\langle \nabla_{\theta} R_{\theta}[u](x_i^f), \nabla_{\theta} R_{\theta}[u](x_j^f) \Bigg\rangle
\\
& K_{Rb, ij} = \Bigg\langle \nabla_{\theta} R_{\theta}[u](x_i^f), \nabla_{\theta} u_{\theta}(x_j^b) \Bigg\rangle
\\
& K_{bb, ij} = \Bigg\langle \nabla_{\theta} u_{\theta}(x_i^b), \nabla_{\theta} u_{\theta}(x_j^b) \Bigg\rangle .
\end{align}
Thus, the FE NTK is the enhanced PDE kernel. Now, converting the NTK to Fourier space on a periodic domain as an operator \cite{lin2026decoupleddiffusionsamplinginverse} (this reference uses it as a convolution kernel operator but it holds more generalized with pseudo representations)
\begin{align}
\label{eqn:fourier_kernel}
\widehat{K}_{RR} = \mathcal{F} \circ K_{RR} \circ \mathcal{F}^{-1} .
\end{align}
Hence, notice the neural tangent kernel relation 
\begin{align}
\widehat{K}_{RR}(\xi, \xi') = \iint e^{-2\pi i \langle \xi, x \rangle} K_{RR}(x, y) e^{2\pi i \langle \xi', y \rangle} dx dy .
\end{align}
We omit the domain of integration notation for simplicity to allow generalized domains since we assume a periodic domain. Moreover under regularity,
\begin{align}
K_{FE}(\xi, \xi') & = \langle \nabla_\theta \widehat{R}_\theta(\xi), \nabla_\theta \widehat{R}_\theta(\xi') \rangle = \sum_k \frac{\partial \widehat{R}_\theta(\xi)}{\partial \theta_k} \overline{\frac{\partial \widehat{R}_\theta(\xi')}{\partial \theta_k}}
\\
& = \sum_k \big( \int e^{-2\pi i \langle \xi, x \rangle} \frac{\partial R_\theta(x)}{\partial \theta_k} dx \big) \overline{ \big( \int e^{-2\pi i \langle \xi', y \rangle} \frac{\partial R_\theta(y)}{\partial \theta_k} dy \big)  }
\\
& = \sum_k \big( \int e^{-2\pi i \langle \xi, x \rangle} \frac{\partial R_\theta(x)}{\partial \theta_k} dx \big)  \big( \int e^{2\pi i \langle \xi', y \rangle} \frac{\partial R_\theta(y)}{\partial \theta_k} dy \big)  
\\
&  = \iint e^{-2\pi i \langle \xi, x \rangle} \left( \sum_k \frac{\partial R_\theta(x)}{\partial \theta_k} \frac{\partial R_\theta(y)}{\partial \theta_k} \right) e^{2\pi i \langle \xi', y \rangle} dx dy.
\end{align}
It can be noted the complex conjugate of the risk in physical space is itself since the risk is real-valued. Equivalently, 
\begin{align}
\widetilde{K}_{FE}(\xi, \xi') & = \sum_k \frac{\partial \widehat{R}_\theta(\xi)}{\partial \theta_k} \frac{\partial \widehat{R}_\theta(\xi')}{\partial \theta_k}
\\
& = \sum_k \big( \int e^{-2\pi i \langle \xi, x \rangle} \frac{\partial R_\theta(x)}{\partial \theta_k} dx \big)  \big( \int e^{- 2\pi i \langle \xi', y \rangle} \frac{\partial R_\theta(y)}{\partial \theta_k} dy \big)  
\\
&  = \iint e^{-2\pi i \langle \xi, x \rangle} \left( \sum_k \frac{\partial R_\theta(x)}{\partial \theta_k} \frac{\partial R_\theta(y)}{\partial \theta_k} \right) e^{- 2\pi i \langle \xi', y \rangle} dx dy.
\end{align}
It is clear that 
\begin{align}
K_{FE}(\xi,\xi') = \widetilde{K}_{FE}(\xi,-\xi') = \widehat{K}_{RR}(\xi,\xi') .
\end{align}
Substituting back into \ref{eqn:fourier_risk_gradient_flow},
\begin{align}
\frac{d \widehat{R}_{\theta(t)}(\xi)}{dt} \propto - \sum_{\xi'} \omega(\xi') \left[ \widehat{R}_{\theta}(\xi') \widehat{K}_{RR}(\xi,\xi') + \overline{\widehat{R}_{\theta}(\xi')} \widehat{K}_{RR} (\xi,-\xi') \right] .
\end{align}
In particular, in the infinite width limit \cite{jacot2020neuraltangentkernelconvergence}, we have cross-frequency interactions vanish because the NTK converges to a translation-invariant kernel $K_{RR}(x-y)$, which is reasonable on periodic domains with Fourier feature embeddings \cite{tancik2020fourierfeaturesletnetworks}, thus $\widehat{K}_{RR}(\xi, \eta) = 0$ for all $\eta \neq \xi$. This is because for the NTK and following equation \ref{eqn:fourier_kernel}
\begin{align}
\widehat{K}(\xi, \eta) & = \iint K(x,y) e^{-2\pi i(x \cdot \xi - y \cdot \eta)} dx  dy 
\\
& \stackrel{\text{translation invariant state}}{=} \iint k(x-y) e^{-2\pi i(x \cdot \xi - y \cdot \eta)} dx  dy
\\
&  \stackrel{\text{change of variables}}{=} \iint k(u) e^{-2\pi i((u+y) \cdot \xi - y \cdot \eta)}  du  dy
\\
&  = \iint k(u) e^{-2\pi i(u \cdot \xi)} e^{-2\pi i(y \cdot \xi - y \cdot \eta)}  du  dy
\\
& = \left( \int k(u) e^{-2\pi i(u \cdot \xi)}  du \right) \left( \int e^{-2\pi i y \cdot (\xi - \eta)}  dy \right)
\\
& \propto \widehat{k}(\xi) \delta_{\xi,\eta}.
\end{align}
Thus the sum vanishes except where the arguments of the kernel sum are equal,
\begin{align}
\frac{d \widehat{R}_{\theta(t)}(\xi)}{dt} \propto - \left( \omega(\xi) \widehat{R}_{\theta}(\xi) \widehat{K}_{RR}(\xi,\xi) + \omega(-\xi) \overline{\widehat{R}_{\theta}(-\xi)} \widehat{K}_{RR}(\xi,\xi) \right) .
\end{align}
Since the physical risk is real-valued, $\widehat{R}_\theta(\xi) = \overline{\widehat{R}_\theta(-\xi)}$. Assuming a symmetric weight, we get the collapse
\begin{align}
\frac{d \widehat{R}_{\theta(t)}(\xi)}{dt} \propto - 2 \omega(\xi) \widehat{R}_{\theta}(\xi) \widehat{K}_{RR}(\xi,\xi) ,
\end{align}
which is a linearized version and is impacted by eigenvalue. We refer to \cite{doikov2024spectralpreconditioninggradientmethods} for spectral preconditioning arguments. Let us conclude our arguments with a slight lack of rigor. Eigenvalue decay is discussed in \cite{murray2023characterizingspectrumntkpower} (see \cite{bietti2021deepequalsshallowrelu} \cite{bietti2019inductivebiasneuraltangent} for ReLU arguments) and is typically under smooth $\text{tanh}(\cdot)$ activation
\begin{align}
\lambda(\xi) \sim |\xi|^{-q} Q^{-|\xi|}  
\end{align}
for some $Q$ (see Corollary 4.7; $\text{tanh}(\cdot)$ is real-analytic, so its Taylor coefficients decay exponentially). For simplicity of argument, let us take $\lambda(\xi) \sim F(\xi)$. Thus, for us
\begin{align}
\tilde{\lambda}(\xi) = \omega(\xi) \cdot F(\xi) . 
\end{align}
Thus, high frequencies converge at a faster rate. In particular, $\omega$ corresponds to a polynomial in Fourier space under our gradient-enhanced methods.

\noindent $ \square $

\vspace{2mm}

\textit{Remark.} Our PDEs in our experiments generally have matching boundaries and conditions, although we do not enforce this in a loss exactly. Therefore, a periodic domain is reasonable.

\section{Spectral preconditioning: efficient non-polynomial scaling of the NTK}
\label{app:spectral_preconditioning}

We can precondition the PDE solution to act as a filter with the pseudo-differential to disperse the considered differential orders in the Fourier loss. We can gain contribution to the backpropagation across a variety of spectral orders with nonconstant scalings for these orders. Moreover, we cut the number of convolutions from a quadratic order to linear. To clarify, our techniques in this section are notable because:
\begin{enumerate}[label=\Roman*., nosep]
\item We can contribute to the learning across a variety of spectral orders (this leads to better learning than just learning say only very high frequencies, etc.);
\item We contribute to NTK scaling highly nonlinearly/not in the manner of a polynomial;
\item These methods are efficient by reducing the number of convolutions.
\end{enumerate}

\vspace{2mm}

\noindent Spectral conditioning is a known phenomenon. It is discussed in \cite{doikov2024spectralpreconditioninggradientmethods} as a method to condition optimization in convex and nonconvex settings via the Hessian which accommodates geometry of the optimization landscape in order to improve gradient descent methods, which are affected by the Hessian eigenvalues. The work more relevant to us is \cite{geifman2024controllinginductivebiaswide}, in which the gradient descent trajectory is affected via the eigenvalues of the neural tangent kernel which moreover invoke learning qualities relating to the spectral biases. This is exactly what we will attempt to do in this section. Note that \cite{shi2025inductivegradientadjustmentspectral} is closely related as well.

\vspace{2mm}

\noindent Reasons for the above are because the neural network and non-constant coefficients are differentiated in a binomial-theorem type manner, as opposed to applying a commutativity rule to the PDE so that all constants are moved outside the differentiation, leaving only a high order. To elaborate more, consider $\partial^{\alpha} a \partial^{\beta} u(x,t) = a \partial^{\alpha} \partial^{\beta} u(x,t)$ (the constant moves outside, the derivatives orders coalesce, whereas this is not true for $\partial^{\alpha} a(x,t) \partial^{\beta} u(x,t)$). In particular, notice
\begin{align}
\partial^\alpha (a u) = \sum_{\gamma \leq \alpha} \binom{\alpha}{\gamma} (\partial^\gamma a) (\partial^{\alpha-\gamma} u) ,
\end{align} 
and moreover we assume the commutator
\begin{align}
[ D, P] u = [ (\partial_t + \Gamma)(P) - (P)(\partial_t + \Gamma) ] u \neq  0 
\end{align}
identically since in this case Clairaut's theorem does not apply. We will examine when $Pu$ solves the PDE, not $u$, thus $P$ is a preconditioner. We remark this section is primarily theoretical. We refer to \cite{luo2020exactcomputationlinearfrequency} \cite{zhang2019explicitizingimplicitbiasfrequency} \cite{rahaman2019spectralbiasneuralnetworks} \cite{luo2019theoryfrequencyprinciplegeneral} \cite{xu2018understandingtraininggeneralizationdeep} on literature that establishes spectral frequencies are learned among a collection of orders, as opposed to learning one frequency and proceeding onto the next (i.e. the process of learning spectral frequencies is continuous and deeply intermingling). 

\vspace{2mm}

\noindent First, notice the following:
\begin{align}
\mathcal{F}[D_x^{\beta} P u] = (2 \pi i\xi)^\beta \sum_{\alpha} \Big( \widehat{a}_\alpha(\xi) * [(2 \pi i\xi)^\alpha \widehat{u}(\xi)] \Big) .
\end{align}
There is a single sum in the above, and the number of convolutions required scales in a linear fashion. Thus, only a linear number of convolutions is needed to compute the above. We will show, by rewriting the above in an alternative form, we gain spectral contribution to the eigenvalue decay quadratically in order.

\vspace{2mm}

\noindent Let us consider a gradient-enhanced setup but with pseudo-differential operator
\begin{align}
\partial_t P(x,t,D) u + \Gamma \Big[ P(x,t,D) u(x,t), \{\partial_x^{\beta} \partial_t^{\gamma} P(x,t,D) u(x,t)\}_{\beta,\gamma} \Big] = 0 .
\end{align}
Thus $P$ acts as a preconditioner
\begin{align}
v = Pu 
\end{align}
solves the PDE. Were we to move $P$ outside with nonconstant functions $a$, the operators obey the convolution theorem
\begin{align}
\mathcal{F}\{a(x)D^{\alpha}u\} = \widehat{a}(\xi) * ((2\pi i \xi)^{\alpha} \widehat{u}(\xi)) ,
\end{align}
just as they do when inside. The pseudo-differential operator corresponds to the PDE enhancement and should be applied term-by-term with the same $P(D)$. We highlight $P(D)$ does not correspond to the derivatives in the PDE, and these PDE derivatives are still evaluated with automatic differentiation in the physics loss. Observe via the chain rule
\begin{align}
\partial_t P(x,t,D) u & = \dot{P} u + P \dot{u}
\\
& = \Bigg(  \int_{\mathbb{R}^n} \int_{\mathbb{R}^n} e^{2 \pi i \langle x - y, \xi \rangle} \dot{P}(x,t,\xi) u(y,t) dy d\xi \Bigg) + 
\\
& \ \ \ \ \ \ \ \ \ \ \ \ \ + \Bigg(  \int_{\mathbb{R}^n} \int_{\mathbb{R}^n} e^{2 \pi i \langle x - y, \xi \rangle} P(x,t,\xi) \dot{u}(y,t) dy d\xi \Bigg)  . 
\end{align}
When $P$ is a polynomial in Fourier space with coefficients, we get
\begin{align}
& = \Bigg(  \int_{\mathbb{R}^n} \int_{\mathbb{R}^n} e^{2 \pi i \langle x - y, \xi \rangle} \Big( \sum_{|\alpha| \leq m} \partial_t a(x,t) \xi^{\alpha} \Big) u(y,t) dy d\xi \Bigg)  + 
\\
& \ \ \ \ \ \ \ \ \ \ \ \ \ + \Bigg(  \int_{\mathbb{R}^n} \int_{\mathbb{R}^n} e^{2 \pi i \langle x - y, \xi \rangle} \Big( \sum_{|\alpha| \leq m} a(x,t) \xi^{\alpha} \Big) \dot{u}(y,t) dy d\xi \Bigg)  . 
\end{align}
Endow sufficient regularity conditions and, moreover, with respect to $\Gamma$, we have 
\begin{align}
& D_x^{\beta} P(x,t,D) u(x) 
\\
& = \sum_{|\alpha| \leq m}  \int_{\mathbb{R}^n} \int_{\mathbb{R}^n} D_x^{\beta} \left( e^{2 \pi i \langle x - y, \xi \rangle} a_{\alpha}(x,t) \right) \xi^{\alpha} u(y) dy d\xi 
\\
& = \sum_{|\alpha| \leq m} \sum_{\gamma \leq \beta} \binom{\beta}{\gamma}  \int_{\mathbb{R}^n} \int_{\mathbb{R}^n} \Big( D_x^{\gamma} e^{2 \pi i \langle x - y, \xi \rangle} \Big) \Big( D_x^{\beta - \gamma} a_\alpha(x, t) \Big) \xi^{\alpha} u(y) dy d\xi 
\\
& = \sum_{|\alpha| \leq m} \sum_{\gamma \leq \beta} \binom{\beta}{\gamma}  \int_{\mathbb{R}^n} \int_{\mathbb{R}^n} e^{2 \pi i \langle x - y, \xi \rangle} \Big( D_x^{\beta - \gamma} a_\alpha(x, t) \Big) (2\pi i)^{|\gamma|} \xi^{\alpha + \gamma} u(y) dy d\xi 
\\
& = \sum_{|\alpha| \leq m} \sum_{\gamma \leq \beta} \binom{\beta}{\gamma} (2\pi i)^{|\gamma|}\int_{\mathbb{R}^n}  e^{2 \pi i \langle x, \xi \rangle} \Big( D_x^{\beta - \gamma} a_\alpha(x, t) \Big)  \xi^{\alpha + \gamma} \int_{\mathbb{R}^n}  e^{-2 \pi i \langle y, \xi \rangle} u(y) dy d\xi 
\\
& = \sum_{|\alpha| \leq m} \sum_{\gamma \leq \beta} \binom{\beta}{\gamma} (2\pi i)^{|\gamma|}\Big( D_x^{\beta - \gamma} a_\alpha(x, t) \Big) \Big[  \int_{\mathbb{R}^n}  e^{2 \pi i \langle x, \xi \rangle}   \xi^{\alpha + \gamma} \widehat{u}(\xi) d\xi  \Big]
\\\
& = \sum_{|\alpha| \leq m} \sum_{\gamma \leq \beta} \binom{\beta}{\gamma} (2\pi i)^{|\gamma|}\Big( D_x^{\beta - \gamma} a_\alpha(x, t) \Big) \mathcal{F}^{-1} \Big[ \xi^{\alpha  + \gamma} \widehat{u}(\xi) \Big]
\\
& = \sum_{|\alpha| \leq m} \sum_{\gamma \leq \beta} \binom{\beta}{\gamma}  (2\pi i)^{-|\alpha|} \Big( D_x^{\beta - \gamma} a_\alpha(x, t) \Big) D_x^{\alpha + \gamma} u(x) .
\end{align}
Now let us look at the loss function pointwise
\begin{align}
L(x,t) = | \text{PDE residual} |^2 = | \sum_{|\beta|} \partial^{\beta} u |^2  .
\end{align}
Thus,
\begin{align}
\nabla_{\theta} L = \Big\langle \sum_{|\alpha|} \partial^{\alpha} P u, \nabla_{\theta} \sum_{|\beta|} \partial^{\beta} P  u \Big\rangle .
\end{align}
Now, from the above
\begin{align}
\label{eqn:fourier_nonconstanta}
\nabla_{\theta} D_x^{\beta} P(x,t,D) u_{\theta}(x) =  \sum_{|\alpha| \leq m} \sum_{\gamma \leq \beta} \binom{\beta}{\gamma} (2\pi i)^{|\gamma|}\Big( D_x^{\beta - \gamma} a_\alpha(x, t) \Big) \nabla_{\theta} \mathcal{F}^{-1} \Big[ \xi^{\alpha  + \gamma} \widehat{u}(\xi) \Big] .
\end{align}
Notice $\gamma$ is the contribution from the PDE operator, and $\alpha$ is the contribution from the pseudo-differential. By choosing a low order for $|\alpha|$ we can mitigate the frequency effects in Fourier space.

\vspace{2mm}

\noindent These effects are not realized to the same extent if $P$ is applied to the PDE residual afterwards, and not as a preconditioner. $P$ on the inside forces the derivatives of the weights $a$ to matter, whereas on the outside, $a$ is just a pointwise weight. In both cases $a$ is weight, but in the inside case, so are its derivatives. When $a$ is constant, the differential operators commute.

\vspace{2mm}

\noindent In the inside case, $\nabla_{\theta}$ is now forced to interact with the derivatives of the symbol. In both cases, a collection of orders of $\xi$ is incorporated into the learning, but not the derivatives of the symbol scaling functions. The symbol differentiation is observable in \ref{eqn:fourier_nonconstanta}. Note that different orders of $\xi$ in the gradients correspond to training on different orders of the spectral bias. Thus, by preconditioning, we can obtain gradient contribution from a collection of orders rather that gain newfound contribution from differentiation of $a_{\alpha}(x,t)$. Moreover, we can scale all orders by an arbitrary function, not just by constants, by including nonconstant $a$. These functions are collected into the total functions of the eigenvalue decay (so instead of being a polynomial times exponential for all orders, it is now polynomial times exponential times some other function for all orders).

\vspace{2mm}

\noindent In particular, when considering the preconditioned empirical risk $\Delta$ 
\begin{align}
& \mathcal{L}(\theta) = \sum_{\xi}  | \widehat{\Delta}_{\theta}(\xi)|^2 
\\
& = \sum_{\xi} \Bigg| \widehat{ \dot{P} u + P \dot{u} } + \widehat{\sum_{|\alpha| \leq m} \sum_{\gamma \leq \beta} \binom{\beta}{\gamma} (2\pi i)^{|\gamma|}\Big( D_x^{\beta - \gamma} a_\alpha(x, t) \Big) \mathcal{F}^{-1} \Big[ \xi^{\alpha  + \gamma} \widehat{u}(\xi) \Big] } \Bigg| .
\end{align}
Now, let us apply a dominated convergence-type argument
\begin{align} \nabla_{\theta} \widehat{\Delta}_{\theta}(\xi) &  =  \nabla_{\theta} \text{Re} \Bigg\{ \widehat{ \dot{P} u + P \dot{u} } + \widehat{\sum_{|\alpha| \leq m} \sum_{\gamma \leq \beta} \binom{\beta}{\gamma} (2\pi i)^{|\gamma|}\Big( D_x^{\beta - \gamma} a_\alpha(x, t) \Big) \mathcal{F}^{-1} \Big[ \xi^{\alpha  + \gamma} \widehat{u}(\xi) \Big] } \Bigg\}
\\
& \ \ \ \ \ \ \ + i \nabla_{\theta}  \text{Im} \Bigg\{ \widehat{ \dot{P} u + P \dot{u} } + \widehat{\sum_{|\alpha| \leq m} \sum_{\gamma \leq \beta} \binom{\beta}{\gamma} (2\pi i)^{|\gamma|}\Big( D_x^{\beta - \gamma} a_\alpha(x, t) \Big) \mathcal{F}^{-1} \Big[ \xi^{\alpha  + \gamma} \widehat{u}(\xi) \Big] } \Bigg\}
\end{align}
We can rewrite
\begin{align}
& = \sum_{|\alpha| \leq m} \Bigg( \widehat{\dot{a}}_{\alpha}(\xi) * \big( \xi^{\alpha} \nabla_{\theta} \widehat{u}(\xi) \big) + \widehat{a}_{\alpha}(\xi) * \big( \xi^{\alpha} \nabla_{\theta} \widehat{\dot{u}}(\xi) \big) \Bigg) \\
& \quad + \sum_{|\alpha| \leq m} \sum_{\gamma \leq \beta} \binom{\beta}{\gamma} (2\pi i)^{|\gamma|} \left( \widehat{D_x^{\beta - \gamma} a_\alpha}(\xi) * \big( \xi^{\alpha + \gamma} \nabla_{\theta} \widehat{u}(\xi) \big) \right) . \end{align}
As before as in Appendix \ref{app:spectral_decay},
\begin{align}
\begin{cases}
& \frac{d \widehat{\Delta}_{\theta}(\xi)}{dt}  = - \text{constants} \Bigg\langle \nabla_{\theta} \widehat{\Delta}_{\theta}(\xi), \sum_{\xi'} \omega(\xi') \left[ \overline{\widehat{\Delta}_\theta(\xi')} \nabla_{\theta} \widehat{\Delta}_\theta(\xi') + \widehat{\Delta}_\theta(\xi') \overline{\nabla_{\theta} \widehat{\Delta}_\theta(\xi')} \right]\Bigg\rangle  
\\
& K_{\text{FE}}(\xi,\xi') = \Big\langle \nabla_{\theta} \widehat{\Delta}_{\theta}(\xi), \nabla_{\theta } \widehat{\Delta}_{\theta}(\xi') \Big\rangle  ,
\end{cases}
\end{align}
but the above is done with the preconditioned risk, not the true PDE risk. 

\vspace{2mm}

Another remark is that it is obvious that the leading order will dominate the eigenvalue decay. However, the lower orders still contribute to the gradients of the loss with respect to the parameter. Notice $\nabla_{\theta} \widehat{\Delta}_{\theta}(\xi) \propto \mathcal{O}(p^2)$, $ P \ \text{inside} \propto \mathcal{O}(p)$,$  P \ \text{outside} = \mathcal{O}(p^2)$,
appearances $p$. Thus, we can gain gradient contribution from a quadratic number of orders while only needing a linear number of convolutions. By adding $P$ outside instead of by preconditioning, we require $\mathcal{O}(p^2)$ convolutions. Thus we can reduce the number of convolutions by a power in order to gain gradient contribution from a spectrum of orders. 

\vspace{2mm}

\textbf{Spectral preconditioning example.} We consider an example to illustrate the previous arguments. Let us consider the Allen-Cahn PDE $\partial_t u(x,t) - \alpha \partial_{xx} u(x,t) + 5 u^3 (x,t)  - 5 u(x,t) = 0 $. Let us consider the preconditioner with Sobolev weights
\begin{align}
P(\xi) = (1 + |\xi|^2)^{1/2} + (1 + |\xi|^2)^{3/2} = a_1(\xi) + a_2(\xi) .
\end{align}
Therefore, our physics loss is
\begin{align}
& \EX_{u \sim \delta_{\nu}} \Big| \Big| \partial_t ( (a_1 + a_2) * u ) - \alpha \partial_{xx} ( (a_1 + a_2) * u) 
\\
& \ \ \ \ \ \ \ \ \ \ \ \ + 5 ( (a_1 + a_2) * u ) * ((a_1 + a_2) * u) * ((a_1 + a_2) * u) - 5 (a_1 + a_2) * u \Big| \Big|_{L^2(\Omega \times [0,T])}
\\
& = \EX_{u \sim \delta_{\nu}} \Big| \Big| \partial_t ( (a_1 + a_2) * u ) - \alpha \partial_{xx} ( (a_1 + a_2) * u)  \\
& \ \ \ \ \ \ \ \ \ \ \ \ \ \ \ \ \ \ \ \ \ \ \ \  + 5 \mathcal{F} \Big[ \Big( \mathcal{F}^{-1}( (a_1 + a_2 ) * u ) \Big)^3  \Big] - 5 (a_1 + a_2) * u \Big| \Big|_{L^2(\Omega \times [0,T])}
\\
& = \EX_{u \sim \delta_{\nu}} \Big| \Big| \partial_t ( (a_1 + a_2) * u ) - \alpha (2 \pi i \xi)^2 ( (a_1 + a_2) * u) \\
& \ \ \ \ \ \ \ \ \ \ \ \ \ \ \ \ \ \ \ \ \ \ \ \  + 5 \mathcal{F} \Big[ \Big( \mathcal{F}^{-1}( (a_1 + a_2 ) * u ) \Big)^3  \Big] - 5 (a_1 + a_2) * u \Big| \Big|_{L^2(\Omega \times [0,T])}.
\end{align}

\section{PDE setups}

\subsection{Burger's equation details}
\label{app:burger's}

Here, we consider Burger's equation. Burger's equation is notable because it is characterized by strong rifts between positive and negative "halves of waves" roughly speaking. We will specifically use a non-square domain, thus compatible for Monte Carlo methods alone, although we remark FFTs over irregular domains exist in general.

\vspace{2mm}

Let us consider Burger's equation
\begin{align}
\partial_t u(x,t) + u(x,t) \partial_x u(x,t) - \nu \partial_{xx} u(x,t) = 0 .
\end{align}
Our procedure as in \ref{sec:contribution} assumed linearity, while Burger's equation is a nonlinear PDE. Let us adapt our strategy for the nonlinear term $u \partial_x u$. Recall the identity 
\begin{align}
\frac{1}{2} \partial_x (u^2) = u \partial_x u .
\end{align}
Notice
\begin{align} P(D)(u \partial_x u) & =  \int_{\mathbb{R}^n} e^{2 \pi i \langle x, \xi \rangle} P(\xi) \mathcal{F}(u \partial_x u)(\xi) d\xi 
\\
& =  \int_{\mathbb{R}^n} e^{2 \pi i \langle x, \xi \rangle} P(\xi) \Big( \pi i \xi  \mathcal{F}(u^2)(\xi) \Big) d\xi 
\\
& =  \int_{\mathbb{R}^n} e^{2 \pi i \langle x, \xi \rangle} P(\xi) \Big( \pi i \xi  ( \widehat{u} * \widehat{u})(\xi) \Big) d\xi 
\\
& =  \int_{\mathbb{R}^n} e^{2 \pi i \langle x, \xi \rangle} P(\xi) \Big( \pi i \xi  \int_{\mathbb{R}^n} \widehat{u}(\xi - \eta) \widehat{u}(\eta) d\eta \Big) d\xi \\
& = \mathcal{F}^{-1} \Big( \pi i \xi  P(\xi) (\widehat{u} * \widehat{u}) \Big)  .
\end{align}
Thus the loss is
\begin{align}
\mathcal{L}_{\text{enhanced}} = \int_{0}^{T} \int_{\Omega} \left| W(\xi) \cdot \left[ \widehat{\dot{u}}(\xi, t) + \pi i \xi  \widehat{u^2}(\xi, t) + 4 \pi^2 \nu \xi^2 \widehat{u}(\xi, t) \right] \right|^2 d\xi dt .
\end{align}

\vspace{2mm}

\subsection{Allen-Cahn equation details}
\label{app:allen-cahn}

The Allen-Cahn data is where we observe a drop-off effect due to the Fourier enhanced loss. This is notable because the effect is not gradual; however, it is less notable because this was achieved with FFTs, which require high collocation (and so it is less notable for this reason).

\vspace{2mm}

Our setup is the Allen-Cahn equation
\begin{align}
\begin{cases}
& \partial_t u(x,t) - \alpha \partial_{xx} u(x,t) + 5 u^3 (x,t)  - 5 u(x,t) = 0, \ \ \ (x,t) \in [-1,1] \times [0,1]
\\
& u(x,0) = x^2 \cos( \pi x ) . 
\end{cases}
\end{align}
To deal with the nonlinear term, we take
\begin{align}
P(D)(u^3) =  \int_{\mathbb{R}^n} e^{2 \pi i \langle x, \xi \rangle} P(\xi) \mathcal{F}(u^3)(\xi) d\xi .
\end{align}
The loss in an integral formulation is
\begin{align}
\mathcal{L}_{\text{enhanced}} = \int_{0}^{T} \int_{\Omega} \Big| W(\xi) \cdot \Big[ \widehat{\dot{u}}(\xi, t) + \alpha 4 \pi^2 |\xi|^2 \widehat{u}(\xi, t) +  5 \widehat{u^3}(\xi,t) - 5 \widehat{u}(\xi,t)  \Big] \Big|^2 d\xi dt .
\end{align}
We have used
\begin{align}
\mathcal{F}(\partial_t u - \alpha \partial_{xx} u) = \widehat{\dot{u}} - \alpha (2 \pi i\xi)^2 \widehat{u} = \widehat{\dot{u}} + \alpha 4 \pi^2 \xi^2 \widehat{u} .
\end{align}
However, it can be noted that by the convolution theorem
\begin{align}
\mathcal{F}(u^3)(\xi) = (\widehat{u} * \widehat{u} * \widehat{u})(\xi) .
\end{align}
Thus,
\begin{align}
P(D)(u^3) = \mathcal{F}^{-1} \Big( P(\xi) \cdot  (\widehat{u} * \widehat{u} * \widehat{u}) \Big) .
\end{align}

\vspace{2mm}

\subsection{Korteweg-De Vries (KdV) equation details}
\label{app:kdv}

The KdV equation a serious task from a spectral bias perspective, and the data is essentially characterized by oscillation, or at least more so than our other results. The KdV equation will be our equation of interest in which we examine spectral bias learning effects over the course of early training periods. Recall early training periods are generally more notable than later in a task of spectral learning, since high-frequency features develop later in training rather than earlier.

\vspace{2mm}

We setup our PDE as
\begin{align}
\begin{cases}
& \partial_t u(x,t) + u(x,t) \cdot \partial_x u(x,t) + \delta^2 \partial_{xxx} u(x,t) = 0, \ \ \ (x,t) \in [0,2] \times [0,1]
\\
& u(x,0) =- \alpha \cos ( \pi x ) .
\end{cases}
\end{align}
We will vary $\alpha \in \mathbb{R}^+$ in our experiments, primarily within $[0.75, 1.5]$. To deal with the nonlinear term, we take
\begin{align}
\label{eqn:kdv_nonlinear}
P(D)(u u_x) =  \int_{\mathbb{R}^n} e^{2 \pi i \langle x, \xi \rangle} P(\xi) \mathcal{F}(u u_x)(\xi) d\xi .
\end{align}
Again it is possible to use the identity
\begin{align}
\frac{1}{2} \partial_x (u^2) = u \partial_x u ,
\end{align}
so \ref{eqn:kdv_nonlinear} reduces to
\begin{align}
&  \int_{\mathbb{R}^n} e^{2 \pi i \langle x, \xi \rangle} P(\xi) \mathcal{F} \Big( \frac{1}{2} \partial_x (u^2) \Big) (\xi) d\xi =   \int_{\mathbb{R}^n} e^{2 \pi i \langle x, \xi \rangle} P(\xi) \cdot \pi i \xi  \mathcal{F}(u^2)(\xi) d\xi ,
\end{align}
although this is more of a remark and we used \ref{eqn:kdv_nonlinear} experimentally. The loss is
\begin{align}
\mathcal{L}_{\text{enhanced}} & = \int_{0}^{T} \int_{\Omega} \Bigg| W(\xi) \cdot \Big[ \widehat{\dot{u}}(\xi, t) + \pi i\lambda\xi \widehat{u^2}(\xi, t) - 8 \pi^3 \delta^2 i\xi^3 \widehat{u}(\xi, t) \Big] \Bigg|^2 d\xi dt 
\\
& = \int_{0}^{T} \int_{\Omega} \Bigg| W(\xi) \cdot \Big[ \widehat{\dot{u}}(\xi, t) +  \widehat{u u_x}(\xi, t) - 8 \pi^3 \delta^2 i\xi^3 \widehat{u}(\xi, t) \Big] \Bigg|^2 d\xi dt .
\end{align}
For these experiments, we will typically take
\begin{align}
W(\xi) = 1 + (2\pi i \xi) + (2\pi i \xi)^2 .
\end{align}
The $1$ is crucial to note because this term does not correspond to a derivative.

\vspace{2mm}

\subsection{Navier-Stokes details}
\label{app:navier-stokes}

The Navier-Stokes setup is a clear indication of a higher level of spectral bias to be learned via the complexity of the data. This is also our only higher-dimensional domain in space, therefore these results are notable for this reason as well. Generally we found learning to be more computationally nontrivial here, but this is entirely to be expected for more sophisticated data.

\vspace{2mm}

For our Navier-Stokes experiment, we consider the vorticity-stream formulation
\begin{align}
\begin{cases}
& \partial_t \omega  + u \cdot \nabla \omega = \nu \nabla^2 \omega
\\
& \omega = - \nabla^2  \psi 
\\
& \partial_t \omega + \Big( \partial_y \psi \partial_x \omega - \partial_x \psi \partial_y \omega \Big) - \nu \nabla^2 \omega = 0
\\
& \omega = \nabla \times u 
\\
& (x,t) \times [-2 \pi, 2 \pi ] \times [T_0, T], \nu = 0.01.
\end{cases}
\end{align}
The initial condition is generated via white noise and applying a spectral filter in Fourier space
\begin{gather}
\widehat{w}_{rand}(\xi) = \zeta + i\eta, \quad \zeta, \eta \sim \mathcal{N}(0, 1)
\\
S(|\xi|) = \frac{|\xi|^1}{1 + (|\xi|/4)^4}
\\
\widehat{w}(\xi) = \widehat{w}_{rand}(\xi) \cdot S(|\xi|) \cdot \chi_{\{|\xi| \le 4\}} .
\end{gather}
For our Fourier enhanced loss, one potential solution is the identities
\begin{align}
\begin{cases}
& \widehat{\psi} = \mathcal{F}(\psi)
\\
& \widehat{\omega} = (\xi_x^2 + \xi_y^2) \widehat{\psi}
\\
& \widehat{u} = i \xi_y \widehat{\psi}, \widehat{v} = -i \xi_x \widehat{\psi}
\\
& \widehat{\omega}_x = i \xi_x \widehat{\omega}, \widehat{\omega}_y = i \xi_y \widehat{\omega}
\\
& \text{advection} = \mathcal{F}^{-1}(\widehat{u}) \cdot \mathcal{F}^{-1}(\widehat{\omega}_x) + \mathcal{F}^{-1}(\widehat{v}) \cdot \mathcal{F}^{-1}(\widehat{\omega}_y)
\\
& \widehat{\text{Risk}}(\xi) = \widehat{ \dot{\omega} } + \widehat{ \text{advection}} + \nu (\xi_x^2 + \xi_y^2) \widehat{\omega} .
\end{cases}
\end{align}
Empirically, we found the above did not work as well as the Monte Carlo methods, potentially due to the IFFTs. Instead, we perform
\begin{align}
\begin{cases}
&\Phi_{kj} = -e^{-i \langle \xi, x_n \rangle}
\\
& \widehat{\psi} = \frac{1}{N} \Phi \psi
\\
& \widehat{\omega} = (\xi_x^2 + \xi_y^2) \widehat{\psi}
\\
& u = \text{Re}(\Phi^{\dagger} \widehat{u} ), v = \text{Re}(\Phi^{\dagger} \widehat{v} )
\\
& w_x = \text{Re}( \Phi^{\dagger} ( i \xi_x \widehat{\omega} ) ), w_y = \text{Re}( \Phi^{\dagger} ( i \xi_y \widehat{\omega} ) ) 
\\
& \text{advection} = u \cdot w_x + v \cdot w_y 
\\
& \widehat{\text{Risk}}(\xi) = \widehat{\dot{\omega}} + \frac{1}{N} \Phi \cdot \text{advection} + \nu ( \xi_x^2 + \xi_y^2 ) \widehat{\omega} .
\end{cases}
\end{align}
In our experiments, we will propagate the PDE forward with the numerical solver and extract $t_0 = 0.5$ as our starting time, so our time interval of the PINN is $[0.5,1.5]$.

\section{Additional experimental details}

\begin{algorithm}
\caption{Grid-based enhanced loss}\label{alg:diverse_posterior}
Take meshes $(x,t) \in \Omega_x \times \Omega_t = \{ (x_0 + k\Delta x,j \Delta t), k \in \mathbb{N}^n, j \in \mathbb{N}, (\max_{j \in [j]} j) \Delta t = T \}$\;
Evaluate $u_{\theta}(x,t)$ over mesh\;
Compute: $$  \text{FFT}[f] = \widehat{f}(\xi_1, \dots, \xi_j) = \frac{1}{\prod_k N_k} \sum_{x_1=0}^{N_1 - 1} \dots \sum_{x_j=0}^{N_j - 1} f(x_1, \dots, x_j) e^{-2 \pi i \sum_k x_k \xi_k / s_k}, \ \ \  \text{for} \ f = u, \dot{u}, \hdots \; $$  
$\xi \in \Xi := \text{ collection of Fourier wavenumbers}$\;
$ P(\xi) \leftarrow \sum_j a_j (2 \pi i\xi)^j$\;
$ W(\xi) \leftarrow \frac{ P(\xi) }{  ||P(\xi)||_{L^{\infty}(\Omega \times [0,T])}\texttt{.detach()} + \epsilon } \approx \frac{P(\xi)}{ \max | P(\xi)|\texttt{.detach()} + \epsilon }$\;
Compute residual 
$$ \mathcal{L}_{\text{enhanced}} = \frac{1}{N} \sum_{k=1}^N  \Bigg| \Big( ( 2\pi i \xi)^{|\alpha|=1} + (2\pi i \xi)^{|\alpha|=2} + \hdots \Big) \cdot  \text{PDE residual in Fourier space}\Bigg|^2 \; $$ 
Return $\mathcal{L}_{\text{enhanced}}$\;
\end{algorithm}

\begin{algorithm}
\caption{Mesh-invariant Monte-Carlo enhanced Loss (see \ref{sec:grid_vs_montecarlo}, \ref{sec:conclusions} for limiting discussions) }\label{alg:mc_enhanced_loss}
Take collocation points $(x,t) \in \Omega$ sampled from domain with respect to the uniform measure\;
Evaluate $u_{\theta}(x,t)$ and $\dot{u}_{\theta}(x,t)$ at sampled points\;
$K \leftarrow \text{number of modes}$\;
$k \leftarrow k \in \{1, \dots, K\}$\;
$\xi_k \leftarrow k / L$ \;
$\Phi \in \mathbb{C}^{K \times N}, \Phi_{kj} = e^{-2\pi i \langle \xi_k, x_j \rangle }$\;
$ P(\xi) \leftarrow \sum_j a_j (2\pi i \xi)^j$\;
$ W(\xi) \leftarrow \frac{ P(\xi) }{  ||P(\xi)||_{L^{\infty}(\Omega \times [0,T])\texttt{.detach()}} + \epsilon } \approx \frac{P(\xi)}{ \max | P(\xi)|\texttt{.detach()} + \epsilon }$\;
Compute Fourier projections:
$$\mathcal{F}_{\text{MC}}[f] \leftarrow \frac{|\Omega|}{N} \Phi f  \approx \int_{\mathbb{R}^{n}}  f(x,t) e^{-2 \pi i \langle \xi, x \rangle}  dx \ \ \  \text{for} \ f = u, \dot{u}, \hdots \  \text{(not equispaced)} \; $$ 
Compute residual
$$ \mathcal{L}_{\text{enhanced}} = \frac{1}{K} \sum_{k=1}^K \Bigg| \Big( ( 2\pi i \xi)^{|\alpha|=1} + (2\pi i \xi)^{|\alpha|=2} + \hdots \Big) \cdot  \text{PDE residual in Fourier space}   \Bigg|^2 \; $$ 
Return $\mathcal{L}_{\text{enhanced}}$\;
\end{algorithm}

\begin{algorithm}
\caption{Get radial power spectral density}\label{alg:radial_power_alg}
Compute $\text{error} \leftarrow u_{\theta}(x,t) - u(x,t)$, where $u(x,t) := \text{numerical solution}$\;
$h,w \leftarrow \text{error.shape}$\;
$\Psi \leftarrow \text{FFT-shift} [ \text{FFT} [ \text{error} ] ] $\;
$\psi(\xi) \leftarrow |\Psi(\xi)|^2$\;
$X,Y \leftarrow \text{meshgrid}(h,w)$\;
$r(X,Y) \leftarrow \text{int} \{ \sqrt{ (X - \text{center}_x)^2 + (Y - \text{center}_y)^2 }\} $\;
$R \leftarrow \text{index add}(r, \psi)$\;
$C \leftarrow \text{index add}(r, 1)$\;
$\text{avg } R \leftarrow R / C$\;
$ \text{frequencies} \leftarrow \text{linspace}(0,\text{Nyquist frequency},m)$\;
$\text{Return } \text{frequencies, avg R}$\;
\end{algorithm}

\newpage

\section{Additional figures}

\begin{figure}[htbp]
  \centering
  \begin{subfigure}[b]{\textwidth}
    \centering
    \includegraphics[scale=0.325]{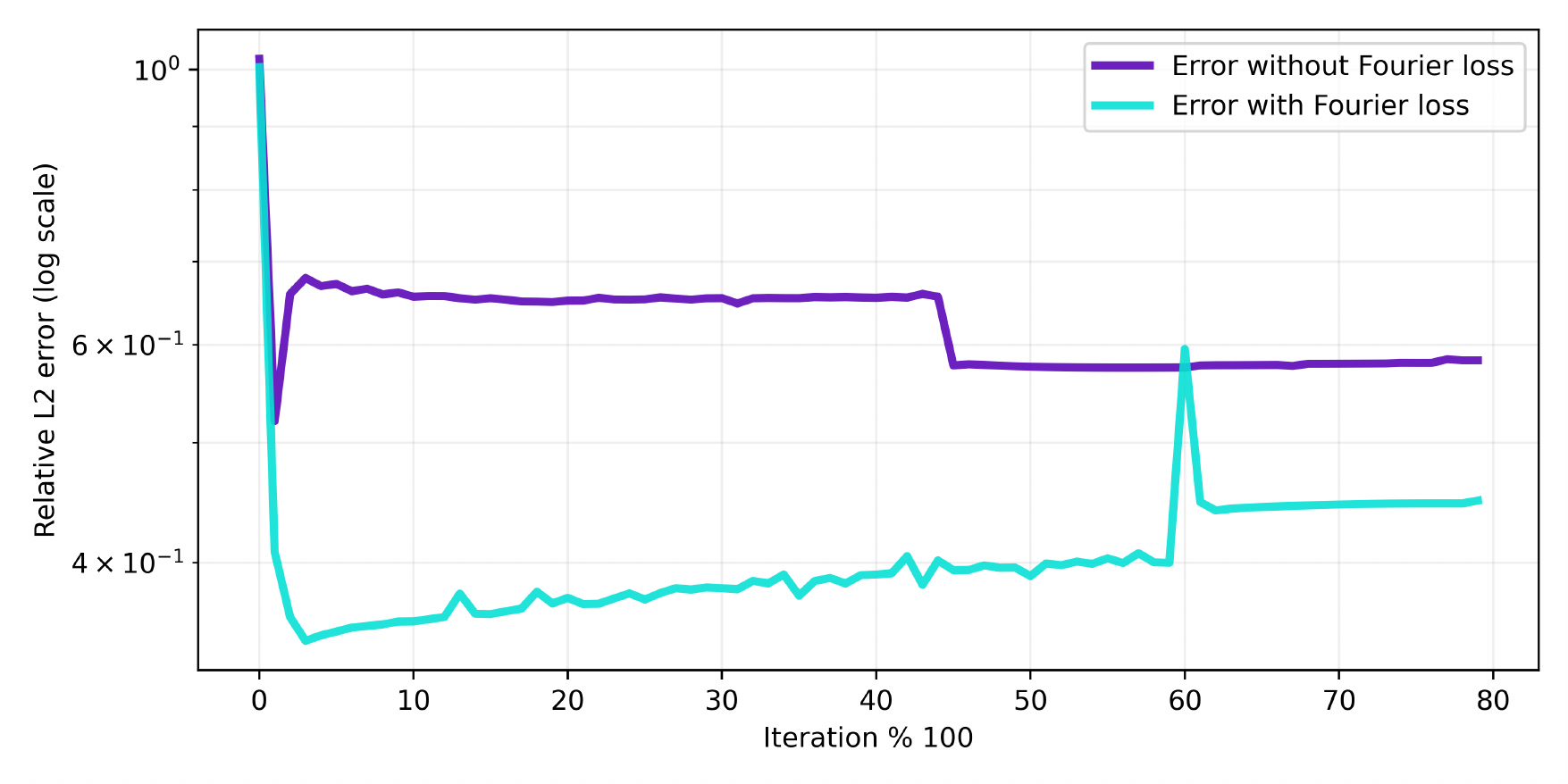}
    \label{fig:error_burgers_nonsquare_fixedxt1}
  \end{subfigure}
  
  \vspace{0.0cm} 
  
  \begin{subfigure}[b]{\textwidth}
    \centering
    \includegraphics[scale=0.325]{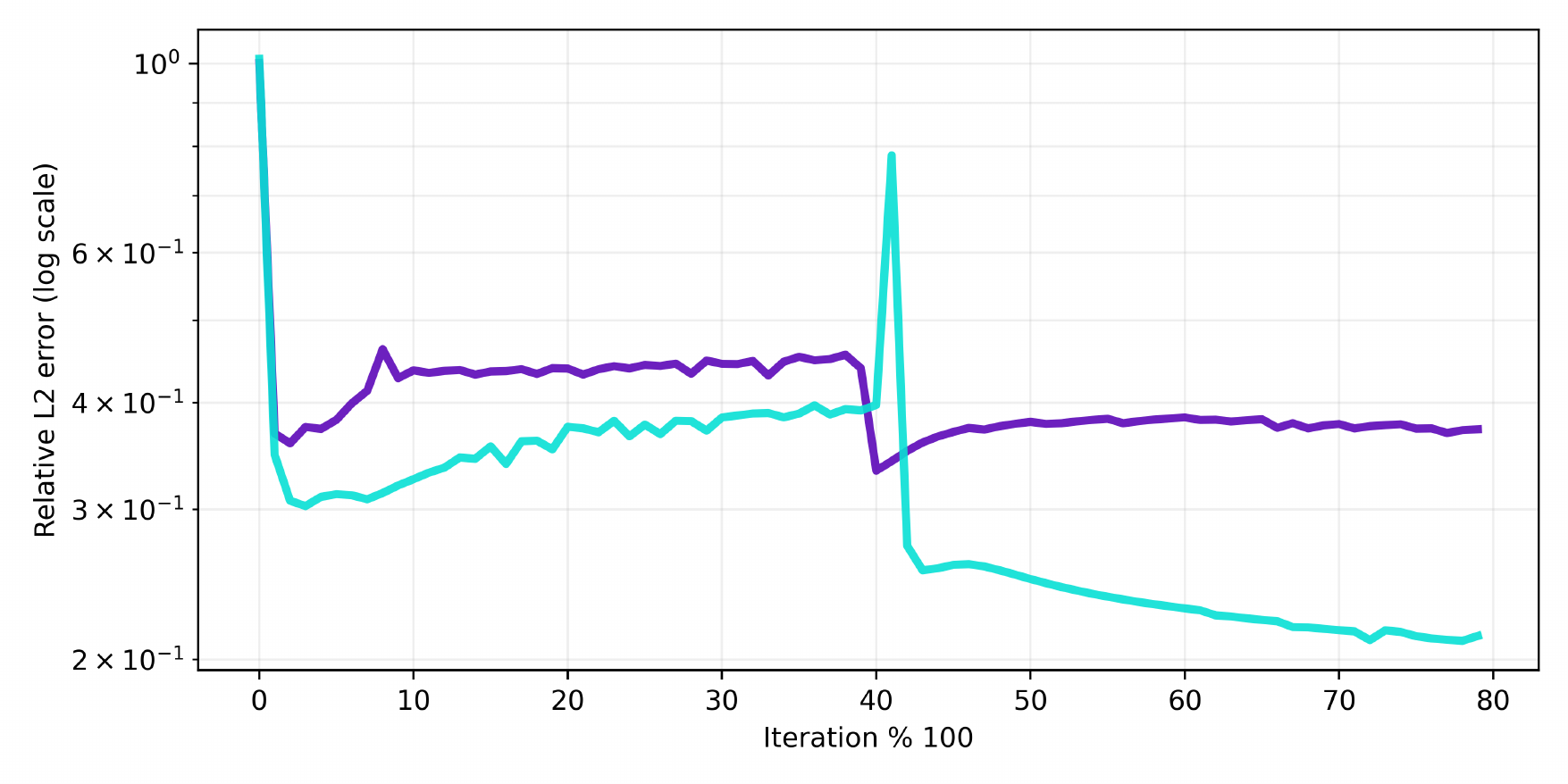}
    \label{fig:error_burgers_nonsquare_fixedxt2}
  \end{subfigure}
  
  \vspace{0.0cm} 

  \begin{subfigure}[b]{\textwidth}
    \centering
    \includegraphics[scale=0.325]{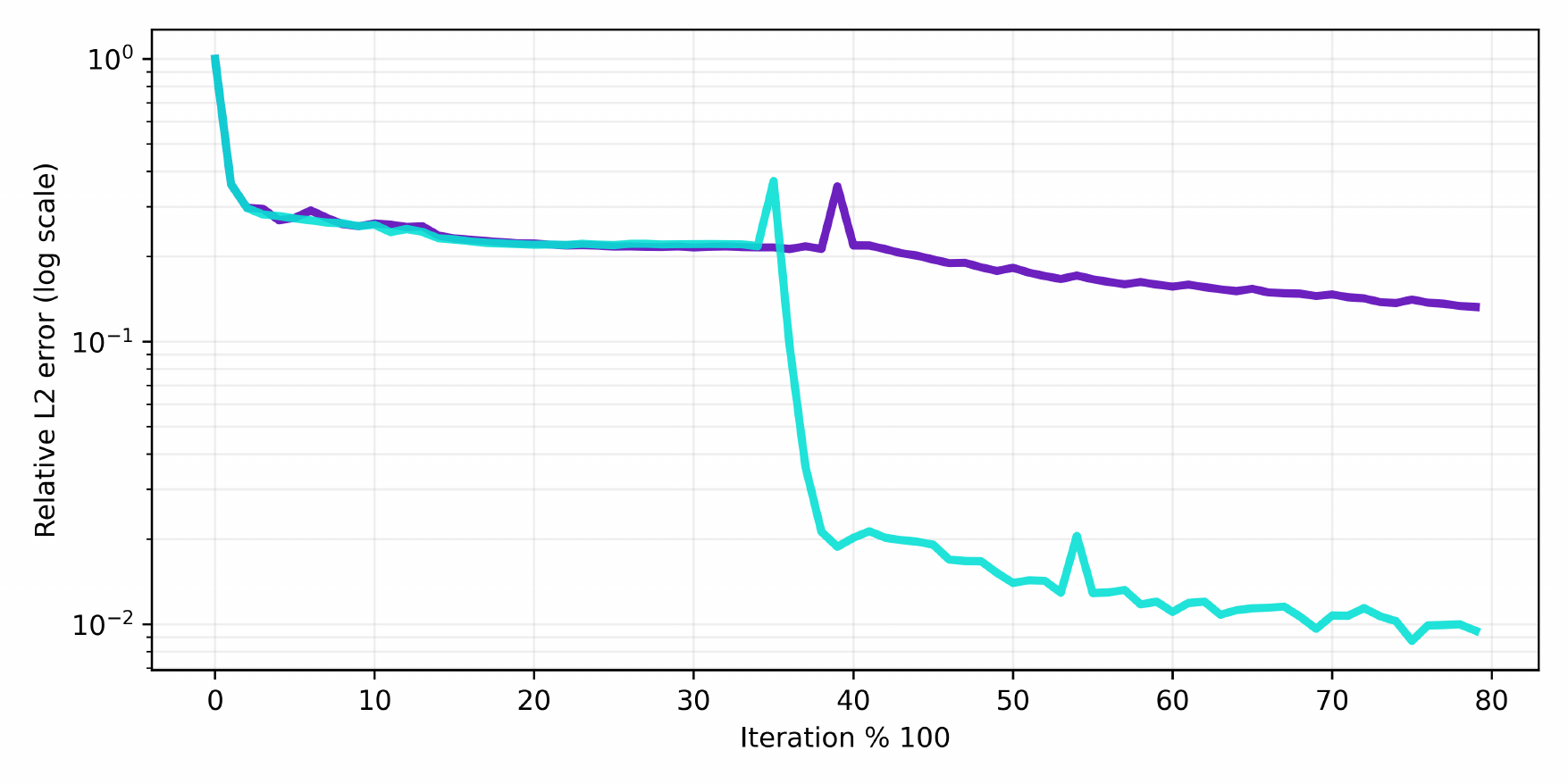}
    \label{fig:error_burgers_nonsquare_fixedxt3}
  \end{subfigure}
  \caption{We plot relative $L^2$ error discretized on an instance of the Burger's equation over $8,000$ training iterations on a triangular domain (so FFTs are not used but rather Monte Carlo techniques are) as in Figure \ref{fig:burgers_nonsquare_images} but here $(x,t)$ are fixed throughout training. This includes fixed $(x,t)$ for both the physics and Fourier losses (the domain is irregular). The error is evaluated along a mesh, thus along points not necessarily seen in training. We implement fixed coefficients with $\lambda_{\text{physics}} = 1, \lambda_{\text{boundary}} = 10$, and take $ \lambda_{\text{enhanced}} = 0.2, 0.5, 0.15 $ (from top to bottom) as the Fourier enhanced loss. We use the Adam optimizer with $\gamma=1\text{e}{-3}$ learning rate, and the Fourier feature embedding with $\sigma=1.0$ in a vanilla MLP. We use a quantile loss with $0.99$ for the enhanced loss. We choose $N=100$ (top); $N=300$ (middle); $N=1,000$ (bottom) collocation points. Our pseudo-differential weight term here takes the form $P(\xi) = 1 + (2\pi i \xi) $ for top-middle and $P(\xi) = 1 + (2\pi i \xi)  + (2\pi i \xi)^2 $ for bottom. We remark we ensured $(x,t)$ had gradients by redefining them each epoch and setting the seed.}
  \label{fig:error_burgers_nonsquare_fixedxt}
\end{figure}

\begin{figure}[htbp]
  \vspace{0mm}
  \centering
  \includegraphics[scale=0.35]{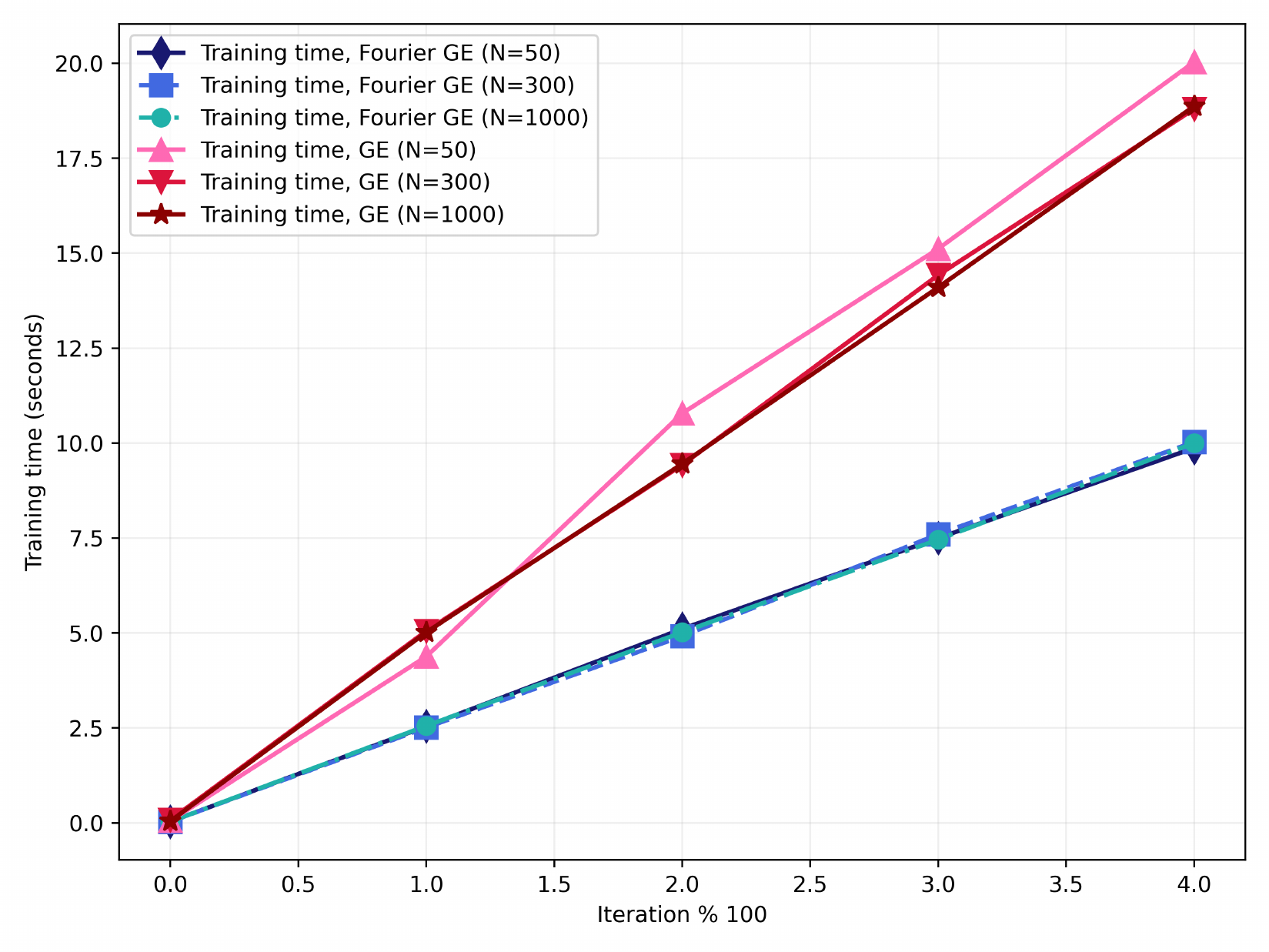}
  \caption{We plot training time versus iteration on the Allen-Cahn equation over $500$ training epochs for varying collocation number $N$ with FFTs. This figure pertains to second order differentiation of the gradient enhancement for both ours and baseline, thus our methods can handle gradient enhancement of high orders rather easily.}
  \label{fig:training_time_fourierge_versus_ge}
\end{figure}

\begin{figure}[htbp]
  \vspace{0mm}
  \centering
  \includegraphics[scale=0.85]{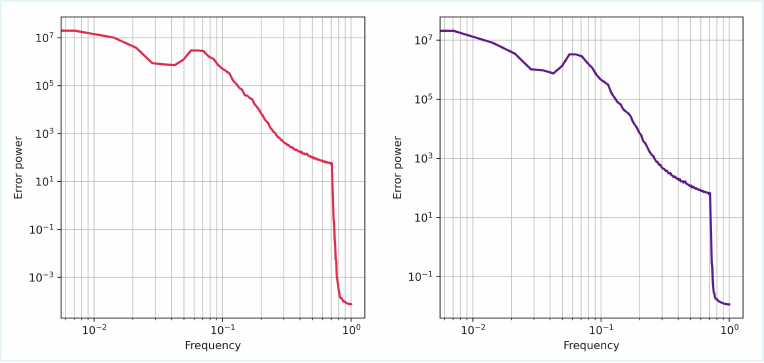}
  \caption{We measure the error power $|| \widehat{u - u_{\theta}} ||_{L^2}^2$ across frequency wavenumber $\xi$ for (left) a Fourier enhanced PINN and (right) a vanilla PINN after 2,000 epochs on KdV data. The figures are similar, but there is a twofold order of mangnitude drop-off on the left for high frequencies, thus the spectral enhanced PINN learns higher frequencies early in training.}
  \label{fig:kdv_error_power}
\end{figure}

\begin{figure}[htbp]
  \vspace{0mm}
  \centering
  \includegraphics[scale=0.35]{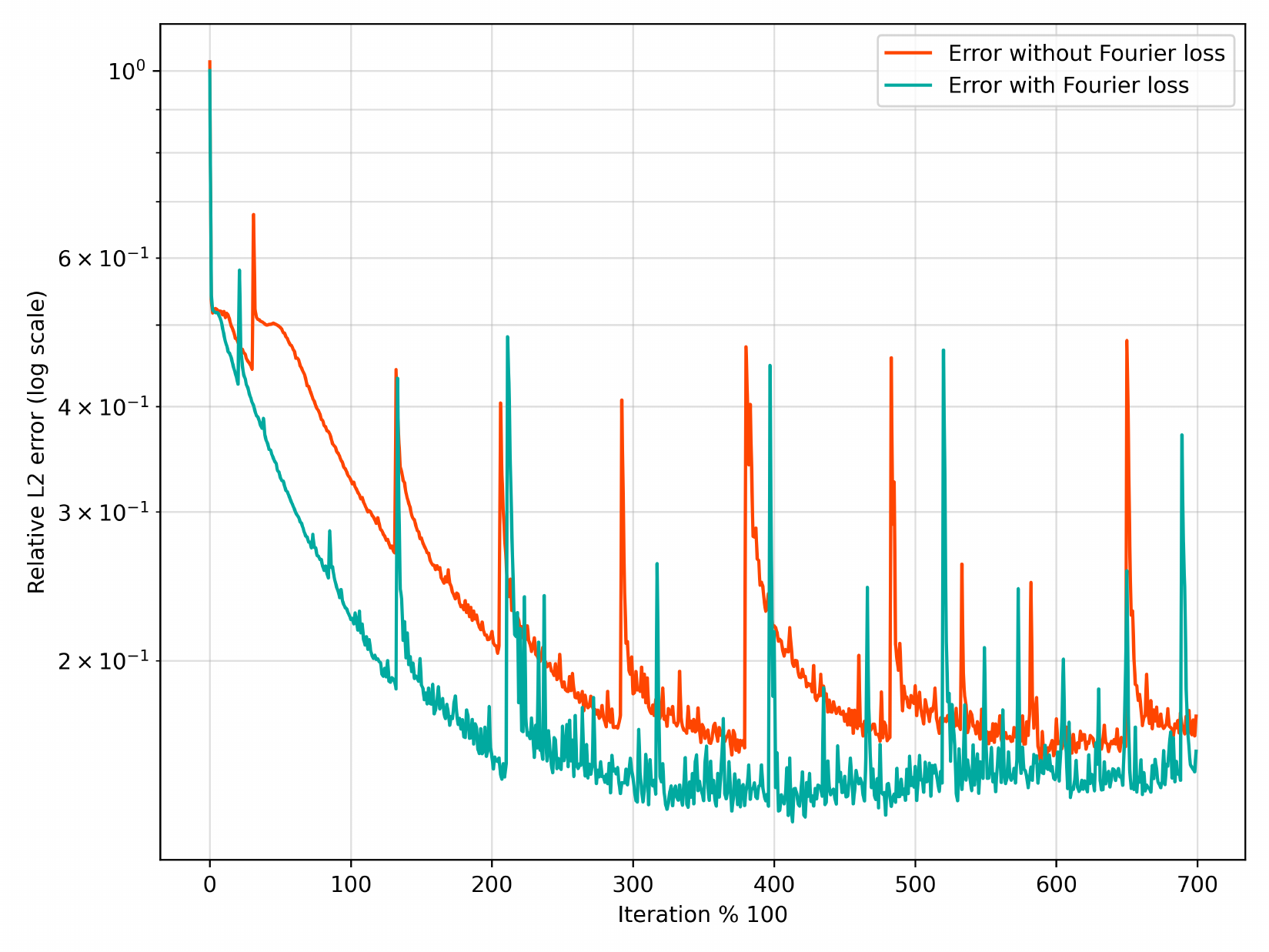}
  \caption{We plot relative $L^2$ error discretized on an instance of the KdV equation over $70,000$ training iterations. We implement fixed coefficients with $\lambda_{\text{physics}} = 1, \lambda_{\text{boundary}} = 10$, and take $ \lambda_{\text{enhanced}} = 0.015 $ as the Fourier enhanced loss. We use the Adam optimizer with $\gamma=1\text{e}{-3}$ learning rate, and the Fourier feature embedding with $\sigma=3.0$ in a vanilla MLP. We use a quantile loss with $0.9$ for the enhanced loss. We choose $N=4,000$ collocation points for the physics loss, and so we emphasize a large batch size. Our pseudo-differential weight term here takes the form $P(\xi) = 1 + (2\pi i \xi) + (2\pi i  \xi)^2$. We also remark we found a low coefficient for $\mathcal{L}_{\text{enhanced}}$ to work better here. }
  \label{fig:error_kdv}
\end{figure}

\begin{figure}[htbp]
  \vspace{0mm}
  \centering
  \includegraphics[scale=0.75]{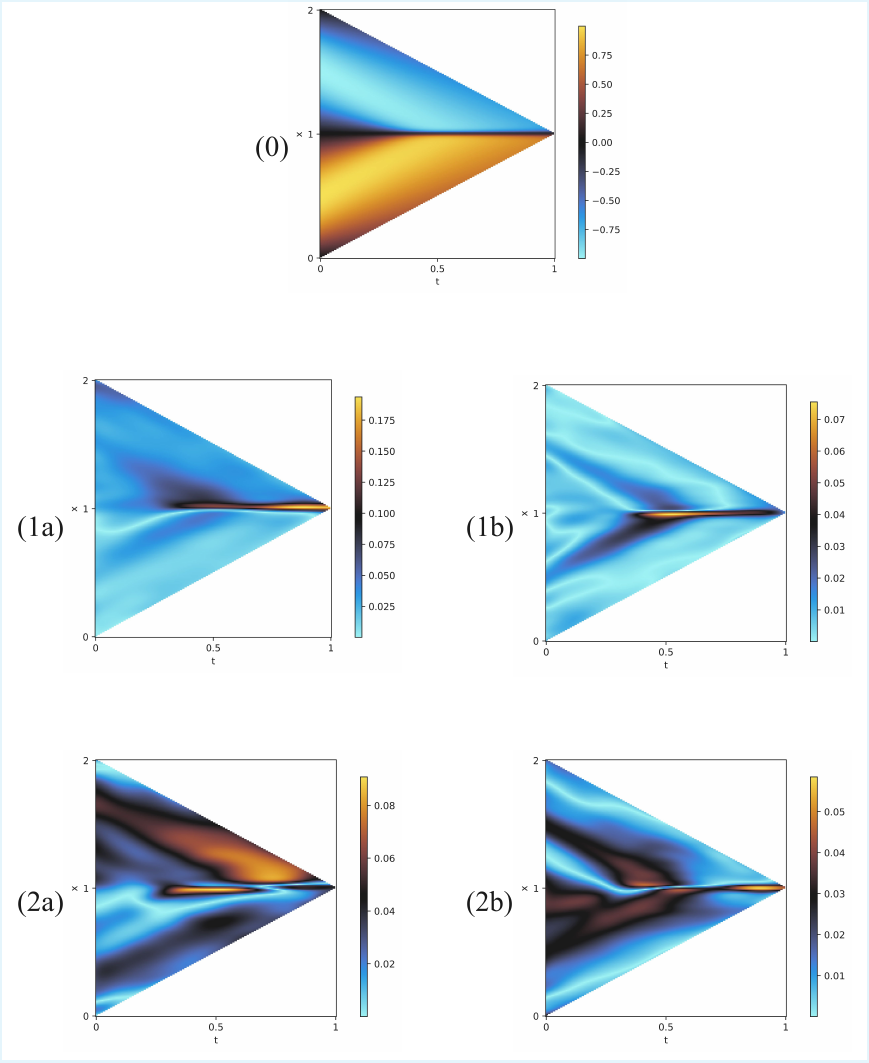}
  \caption{We plot images of pointwise error discretized on an instance of the Burger's equation using advanced techniques over $10,000$ training iterations over an irregular domain with a batch size of $200$. Thus, Monte Carlo approximations were used for the Fourier loss, demonstrating our methods can handle irregular domains. (0) represents the true (numerical) solution. (1a) and (1b) are the Fourier-enhanced PINN at 5,000 and 10,000 iterations, and (2a) and (2c) are a vanilla PINN at 5,000 and 10,000 iterations. As a remark, note that the scales are different. Our method performs better primarily except where the Burger's solution has a rift.}
  \label{fig:burgers_nonsquare_images}
\end{figure}

\begin{figure}[htbp]
  \vspace{0mm}
  \centering
  \includegraphics[scale=0.65]{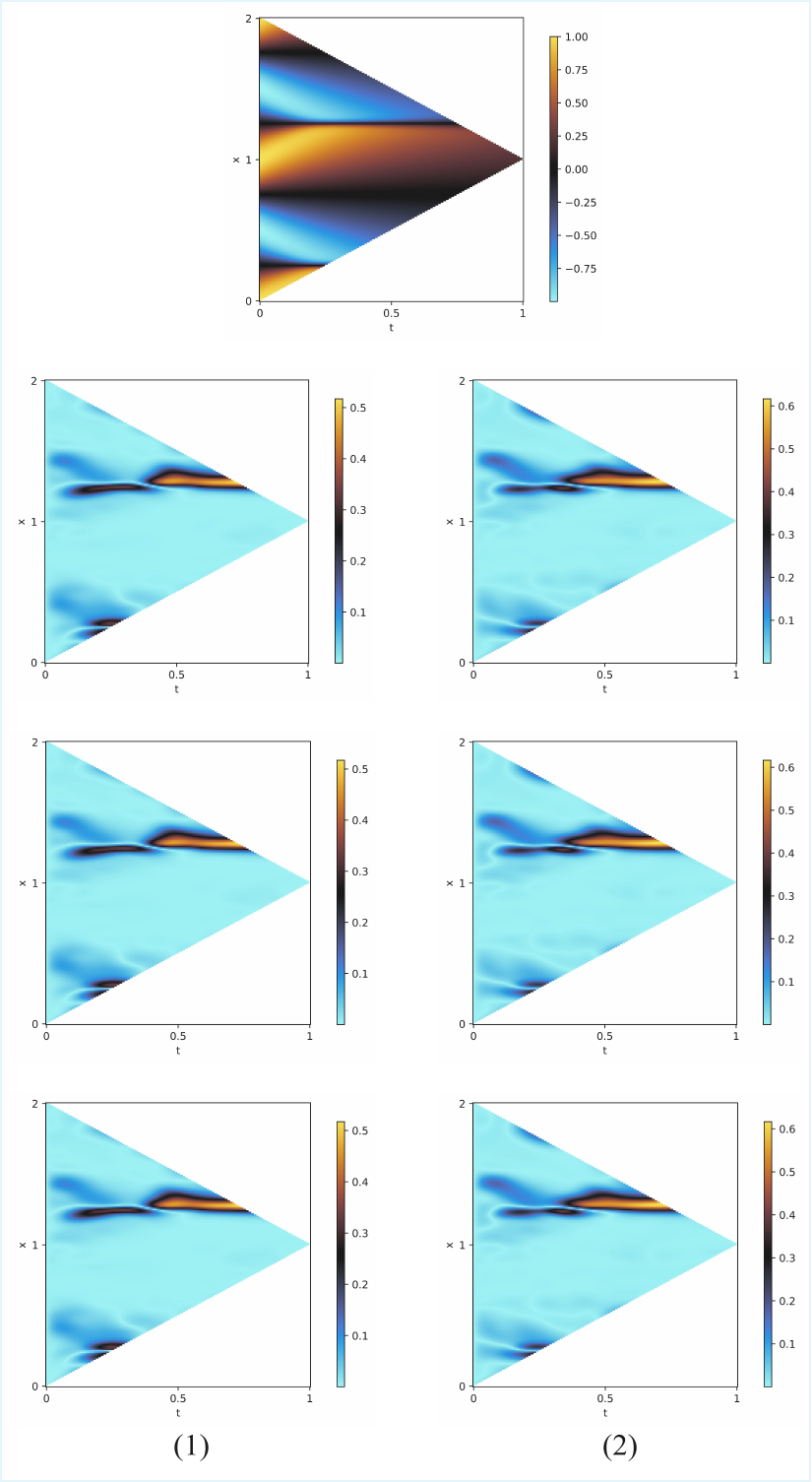}
  \caption{We plot absolute pointwise error of the Burger's equation over $12,000$ training iterations over an irregular domain with a batch size of $500$. (1) in the Fourier enhanced PINN solution with coefficient $\lambda_{\text{enhanced}} = 0.025$ and (2) is the vanilla PINN. As we can see, errors are mostly consistent across retraining, and the Fourier enhanced PINN outperforms. We remark that (1) and (2) look similar but the scales are different.}
  \label{fig:burgers_nonsquare_3samples}
\end{figure}

\begin{figure}[htbp]
  \vspace{0mm}
  \centering
  \includegraphics[scale=0.5]{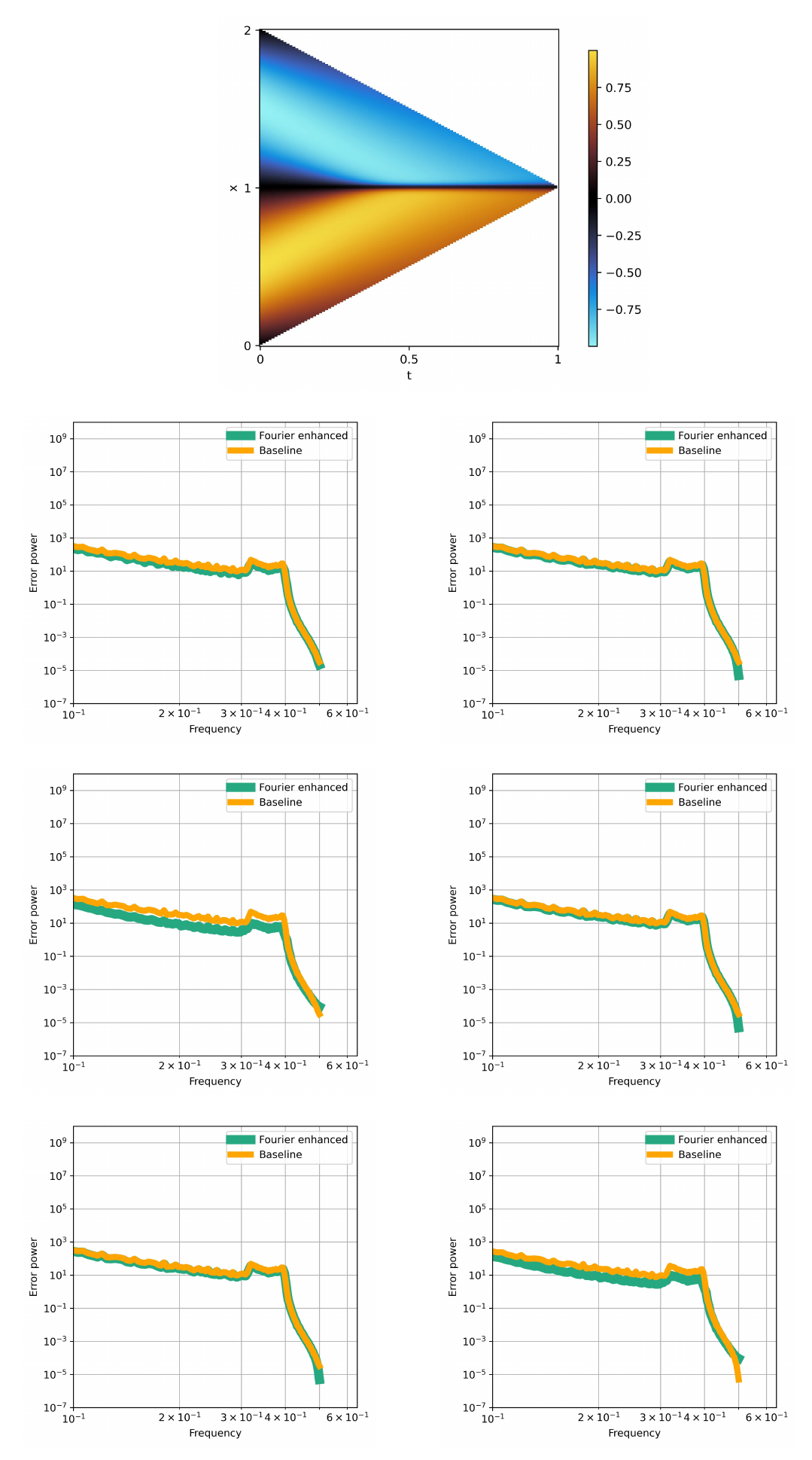}
  \caption{We plot 6 spectral bias learning scenarios corresponding to Algorithm \ref{alg:radial_power_alg} on a solution to the Burger's equation over $3,000$ training iterations on an irregular domain with a batch size of 150 with Monte Carlo techniques. The green is the Fourier enhanced spectral learning, and the orange is the baseline PINN. The top is the true PDE solution. We plot error power versus frequency. Lower is better in general. The rightmost portions of the plot demonstrate learning at the highest frequencies. In 4 out of 6 cases, our proposed methods learn the highest frequencies better. The bulk area, or the uppermost portions of the lines, correspond to high but not highest frequencies. In 6 out of 6, our methods are at least as good for all high frequencies, and for 3 out of 6, our methods outperform in high frequencies. In general, we conclude our methods appeal to spectral bias.}
  \label{fig:burgers_nonsquare_error_power}
\end{figure}

\begin{figure}[htbp]
  \vspace{0mm}
  \centering
  \includegraphics[scale=0.75]{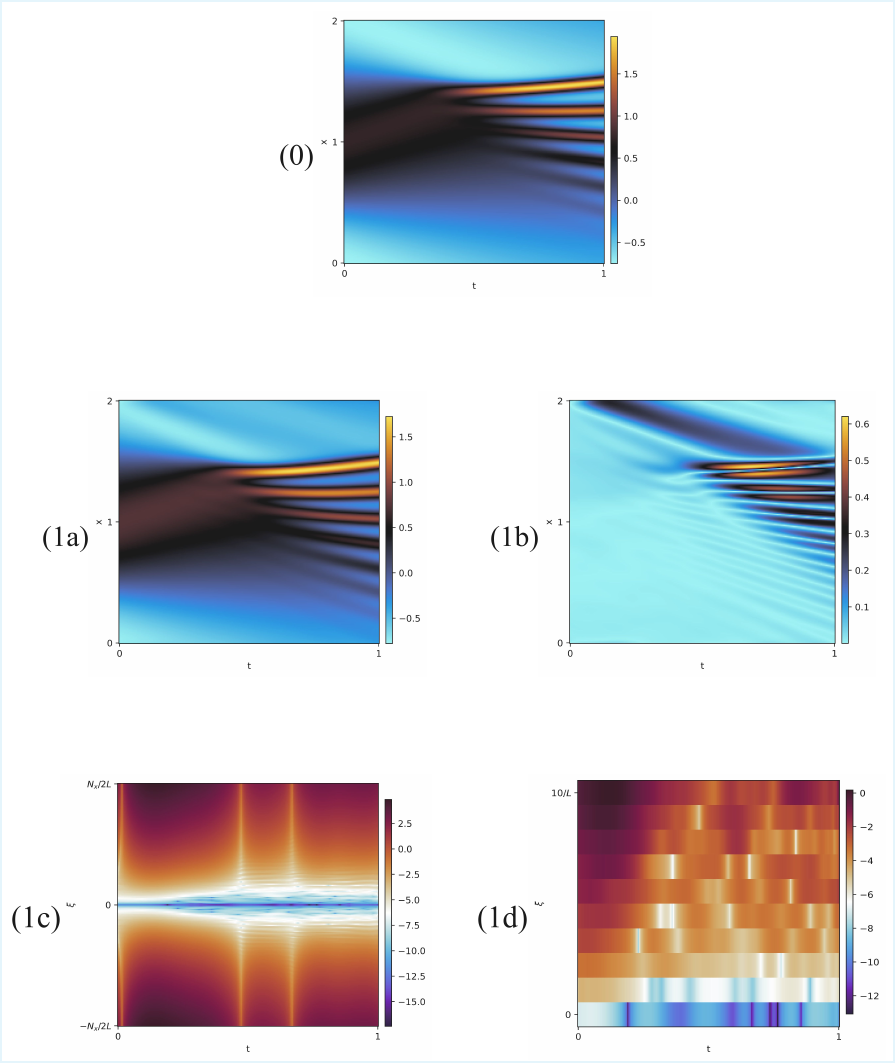}
  \caption{(0) We plot the numerical solution in our Korteweg-De Vries (KdV) experiment; (1a) we plot the PINN solution with a Fourier enhanced loss; (1b) we plot the absolute pointwise error; (1c) we plot the non-truncated Fourier residual with spectral weights in a log plot; (1d) we plot the truncated at Fourier mode 10 residual plot with spectral weights in a log plot. We remark this figure pairs with Figure \ref{fig:error_kdv} but in a new instance of training with fewer iterations ($\sim 35,000$ epochs).}
  \label{fig:kdv_images}
\end{figure}

\begin{figure}[htbp]
  \vspace{0mm}
  \centering
  \includegraphics[scale=0.6]{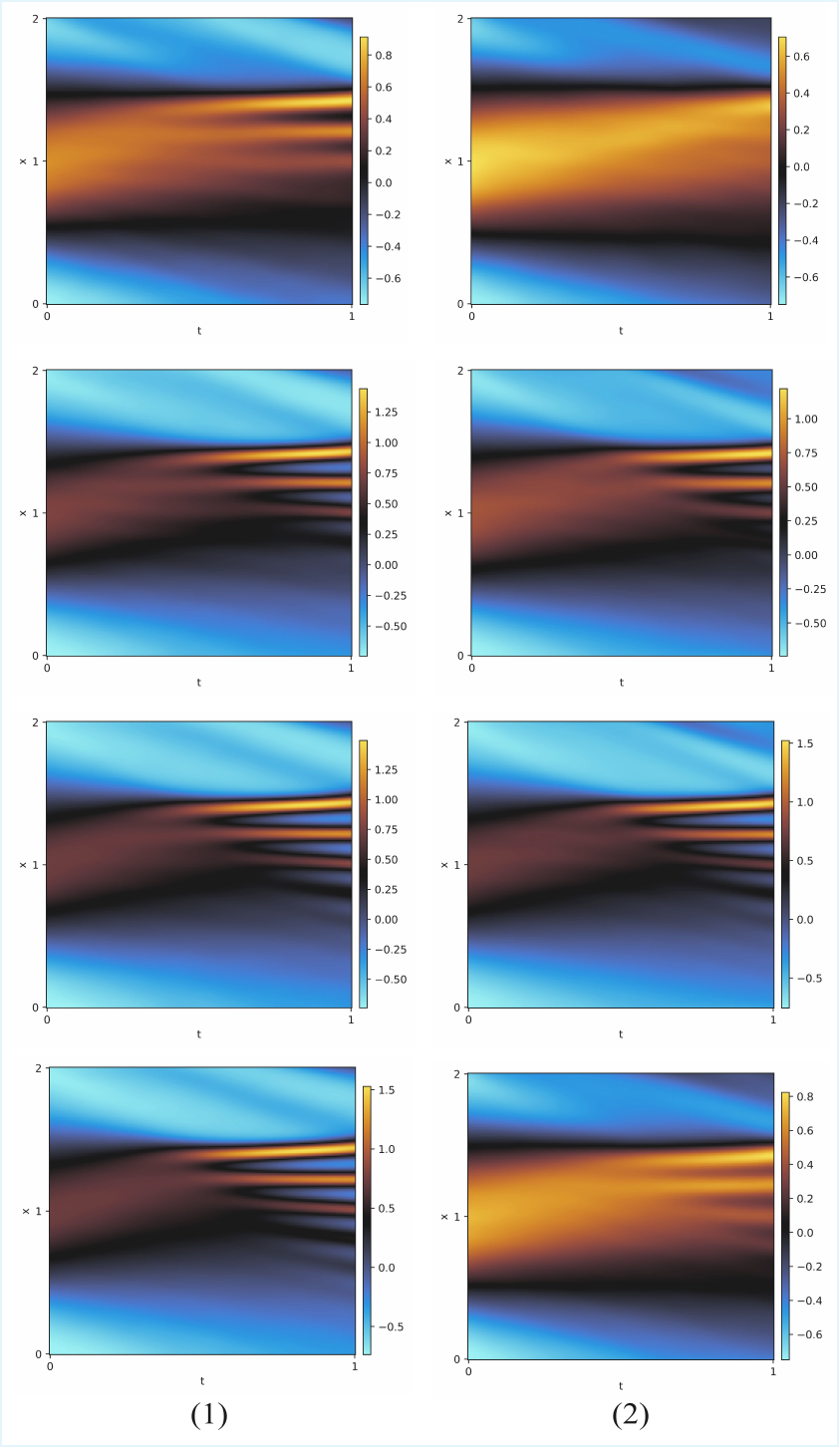}
  \caption{We plot the first 4,000 training epochs in 1,000 epoch chunks for (1) a Fourier enhanced PINN and (2) a baseline PINN. We use a high learning rate $1\text{e}{-3}$. As we can see, the loss is highly stochastic in (2); however, the more interesting part is there is a clear learning of high frequency in the first 1,000 epoch chunk.}
  \label{fig:kdv_images_early_training}
\end{figure}

\begin{figure}[htbp]
  \vspace{0mm}
  \centering
  \includegraphics[scale=0.5]{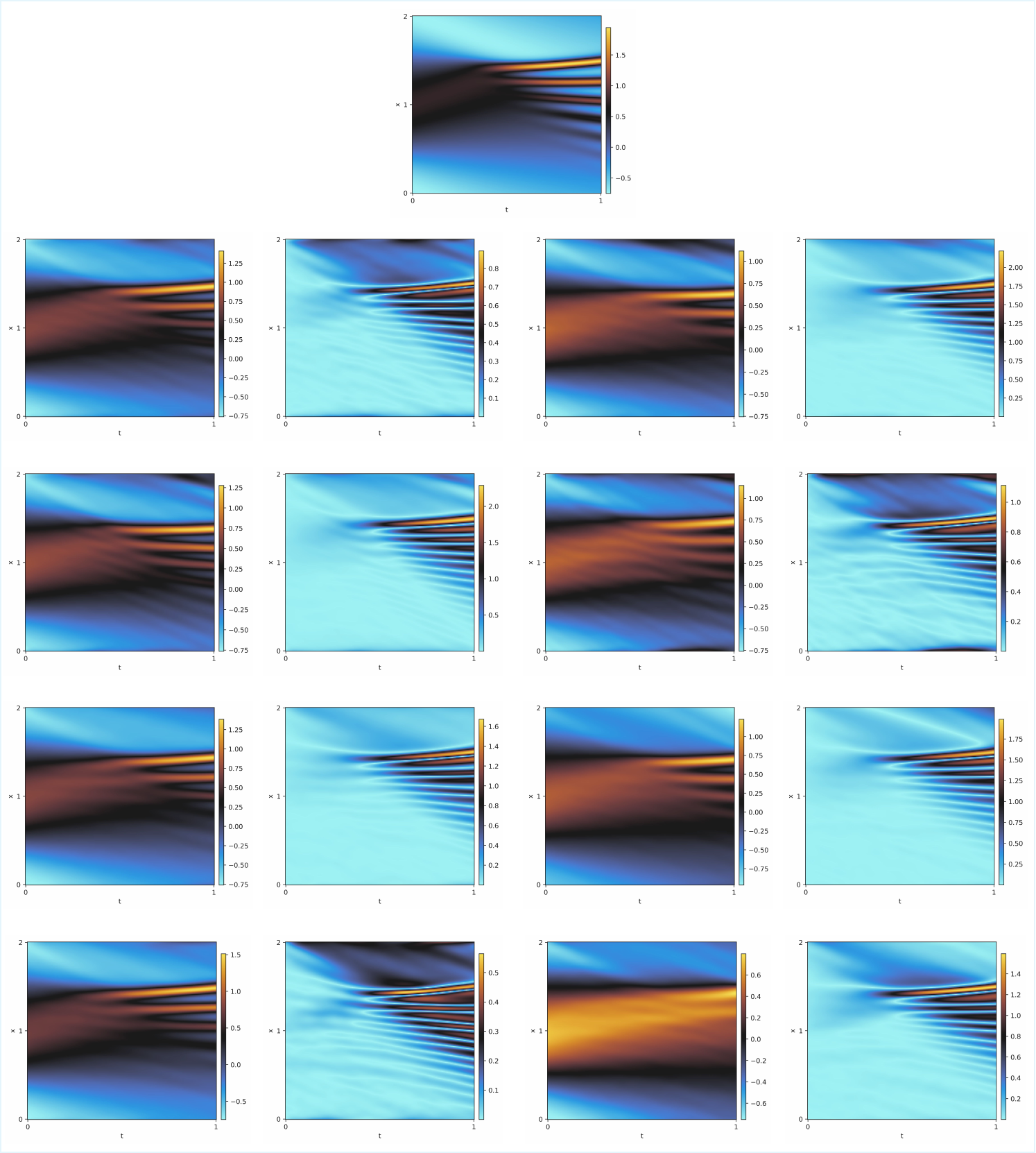}
  \caption{We plot four instances of retraining after a mere 2,000 epochs of the KdV experiment on (left two) a Fourier enhanced PINN; and (right two) a vanilla PINN; and (top) numerical solution. We use Adam optimizer with $1\text{e}{-3}$ learning rate. We set $\lambda_{\text{physics}} = 1$, $\lambda_{\text{boundary}} = 10$, $\lambda_{\text{Fourier}} = 0.01$. We use a vanilla MLP with $\sigma=5.0$ in the Fourier feature embedding. We highlight it is possible this figure demonstrates our methods affect frequency bias, as the left demonstrates higher frequencies learned earlier in training.}
  \label{fig:kdv_images_4samples}
\end{figure}

\begin{figure}[htbp]
  \centering
  \begin{subfigure}[b]{\textwidth}
    \centering
    \includegraphics[scale=0.375]{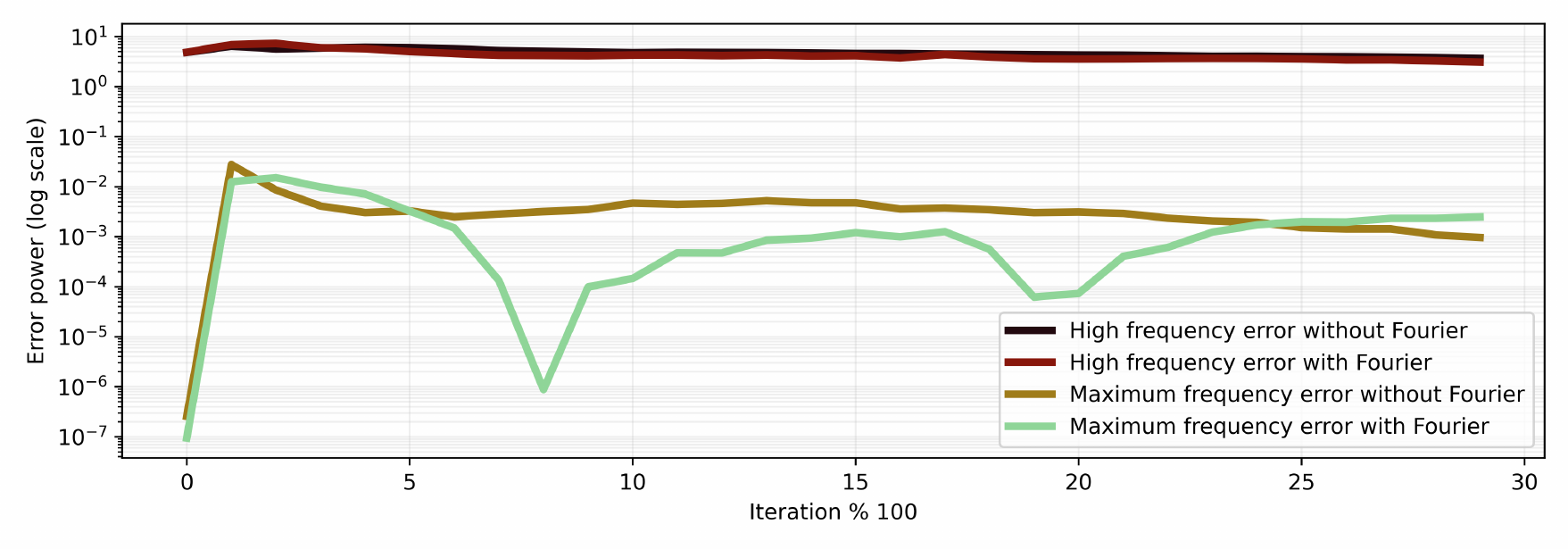}
    \label{fig:kdv_spectral_training_veryhighbatch1}
  \end{subfigure}
  
  \vspace{0.0cm} 
  
  \begin{subfigure}[b]{\textwidth}
    \centering
    \includegraphics[scale=0.375]{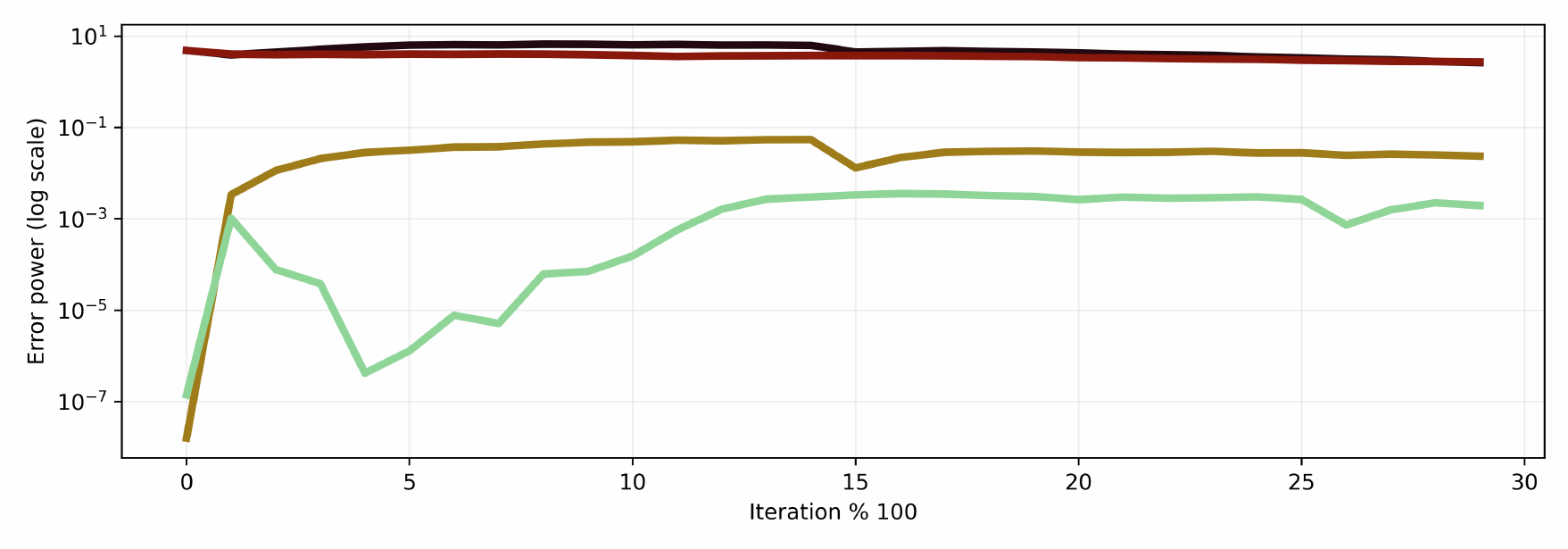}
    \label{fig:kdv_spectral_training_veryhighbatch2}
  \end{subfigure}

  \vspace{0.0cm} 
  
  \begin{subfigure}[b]{\textwidth}
    \centering
    \includegraphics[scale=0.375]{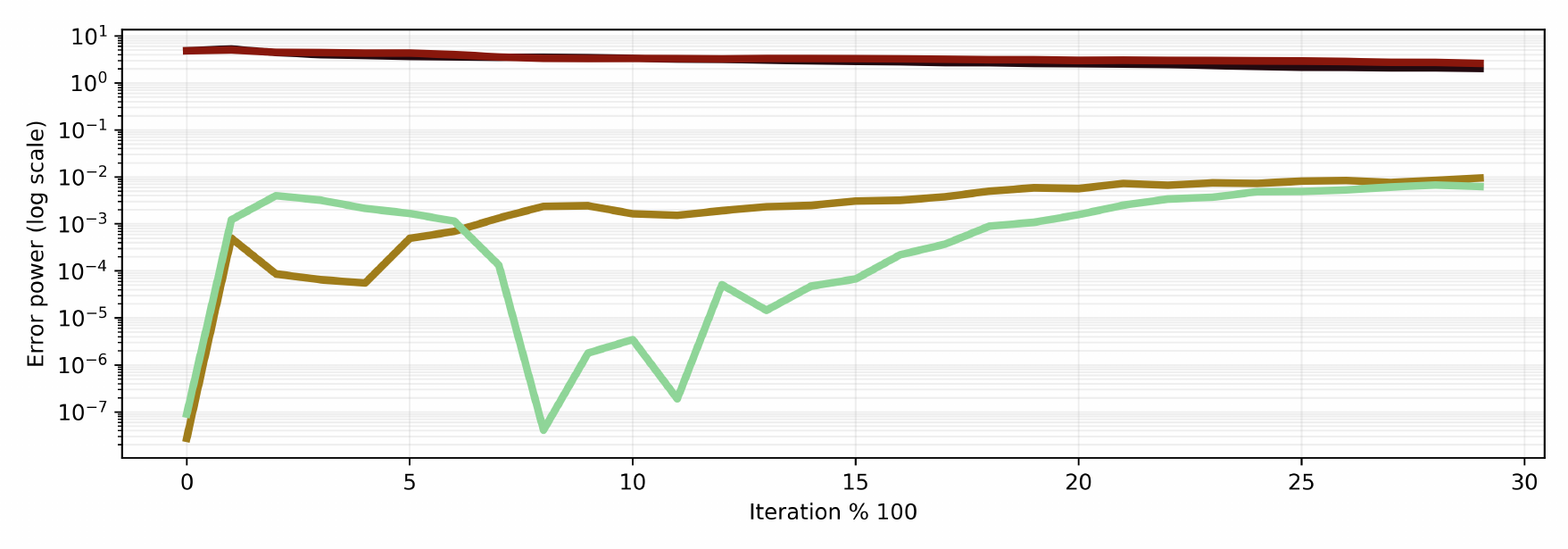}
    \label{fig:kdv_spectral_training_veryhighbatch3}
  \end{subfigure}
  \caption{We plot the spectral bias maximum error power and high error power according to Algorithm \ref{alg:radial_power_alg} for the KdV equation across a Fourier enhanced PINN with FFTs and a baseline PINN across 3,000 training iterations for 3 instances of training. We use learning rate $1\text{e}{-3}$, batch size of 2,500 (which is very high batch size), $\lambda_{\text{Fourier}} = 0.02$, and 10 Fourier modes. We use an enhanced MLP with $\sigma=3.0$ and initial condition $u(x,0) = -0.65 \cos( \pi x)$. Lower is better. We primarily use a very high batch size because the FFT consumes more data points, so we want to conclude that high sampling does not cause the spectral bias mitigation we see with the Fourier enhanced loss.}
  \label{fig:kdv_spectral_training_veryhighbatch}
\end{figure}

\begin{figure}[htbp]
  \centering
  \begin{subfigure}[b]{\textwidth}
    \centering
    \includegraphics[scale=0.375]{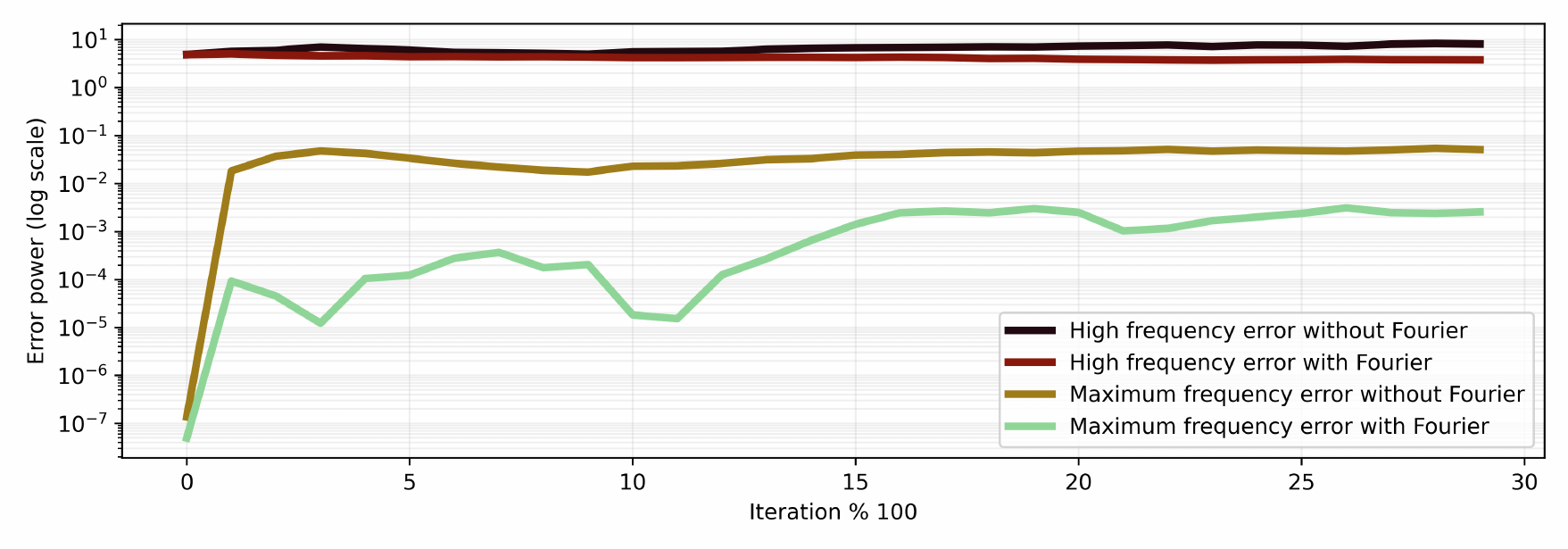}
    \label{fig:kdv_spectral_training_fftsmall1}
  \end{subfigure}
  
  \vspace{0.0cm} 
  
  \begin{subfigure}[b]{\textwidth}
    \centering
    \includegraphics[scale=0.375]{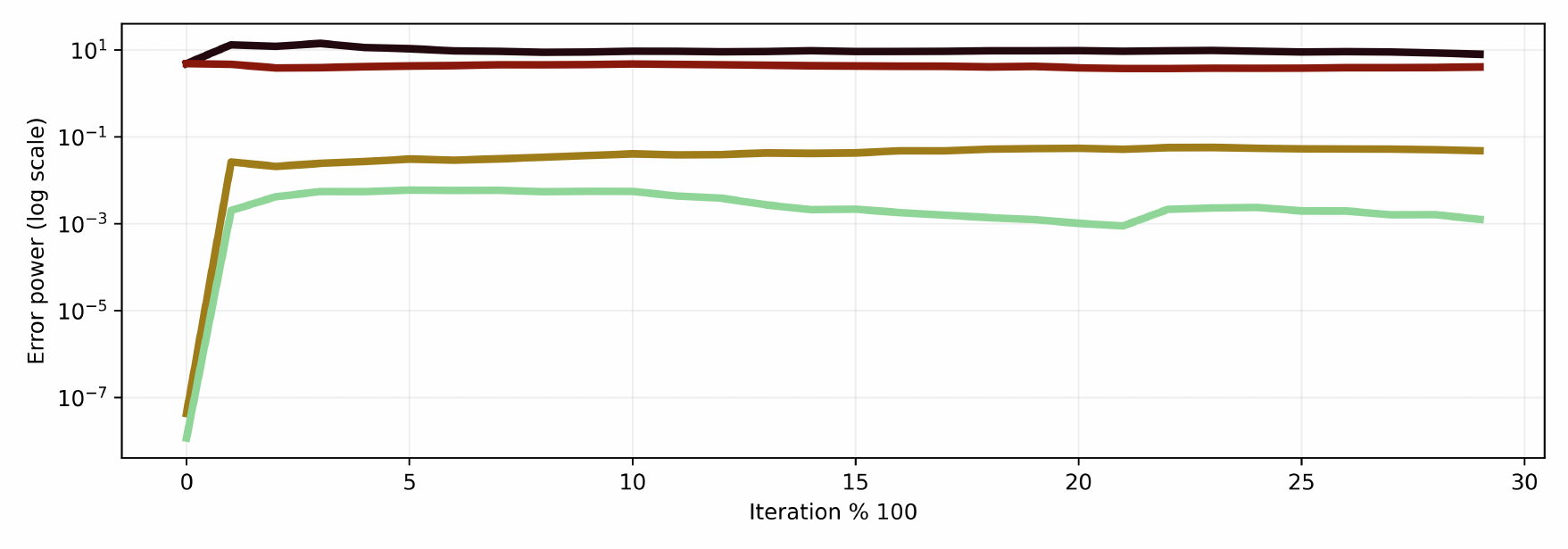}
    \label{fig:kdv_spectral_training_fftsmall2}
  \end{subfigure}

  \vspace{0.0cm} 
  
  \begin{subfigure}[b]{\textwidth}
    \centering
    \includegraphics[scale=0.375]{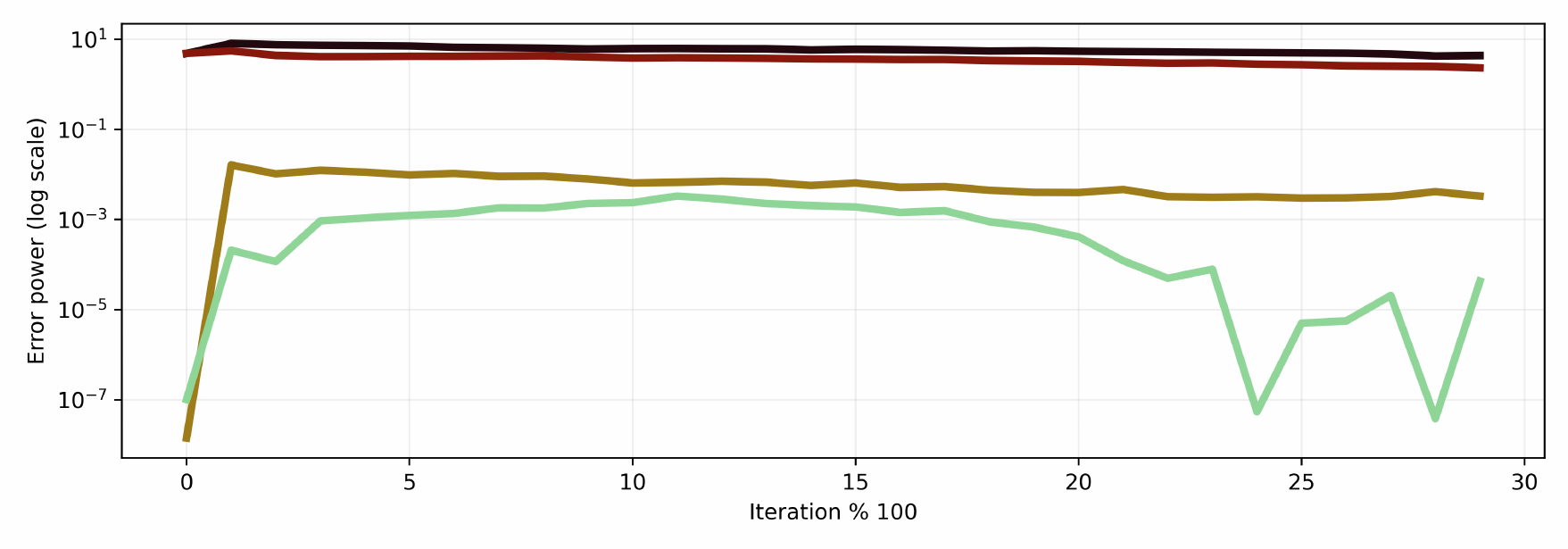}
    \label{fig:kdv_spectral_training_fftsmall3}
  \end{subfigure}
  \caption{We plot the same as Figure \ref{fig:kdv_spectral_training_veryhighbatch} but we use FFT grid sizes of 31-61 rather than 151-301 in order to examine effects of FFT collocation amount on spectral bias. Our batch size is 1,000. We remark that this figure also complements Figure \ref{fig:kdv_fft_smallvslargegrid}. We compare a batch size of 2,500 to a batch size of 1,000 also for a vanilla PINN by comparing to \ref{fig:kdv_spectral_training_veryhighbatch}. Therefore, for a similar batch size comparison, we can compare the Fourier enhanced PINN of this figure to that of the vanilla PINN of \ref{fig:kdv_spectral_training_veryhighbatch}, and clearly the Fourier enhanced PINN outperforms.}
  \label{fig:kdv_spectral_training_fftsmall}
\end{figure}

\begin{figure}[htbp]
  \centering
  \begin{subfigure}[b]{\textwidth}
    \centering
    \includegraphics[scale=0.375]{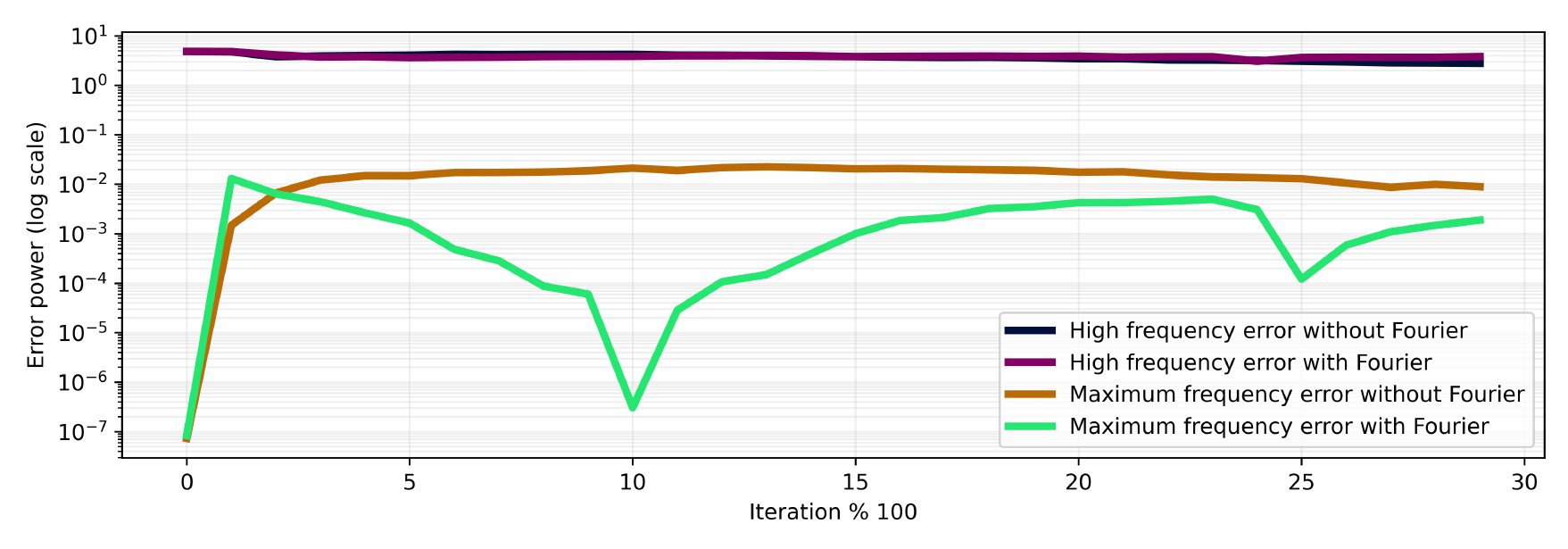}
    \label{fig:kdv_spectral_training_montecarlo1}
  \end{subfigure}
  
  \vspace{0.0cm} 
  
  \begin{subfigure}[b]{\textwidth}
    \centering
    \includegraphics[scale=0.375]{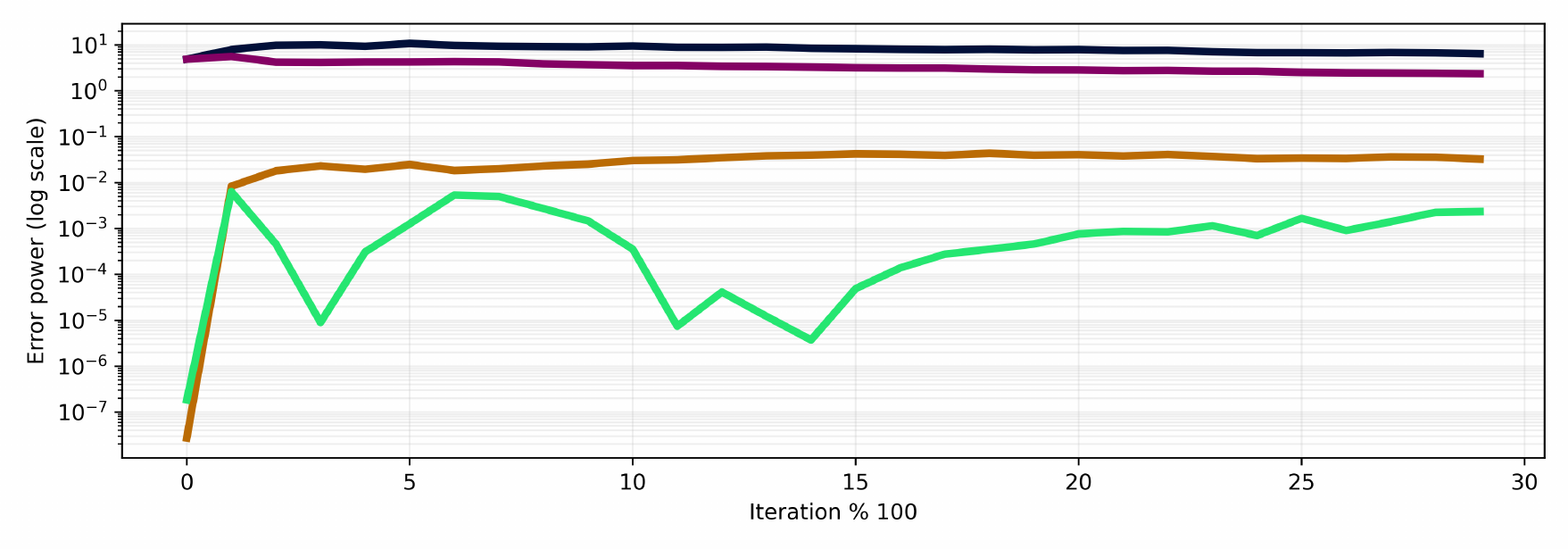}
    \label{fig:kdv_spectral_training_montecarlo2}
  \end{subfigure}

  \vspace{0.0cm} 
  
  \begin{subfigure}[b]{\textwidth}
    \centering
    \includegraphics[scale=0.375]{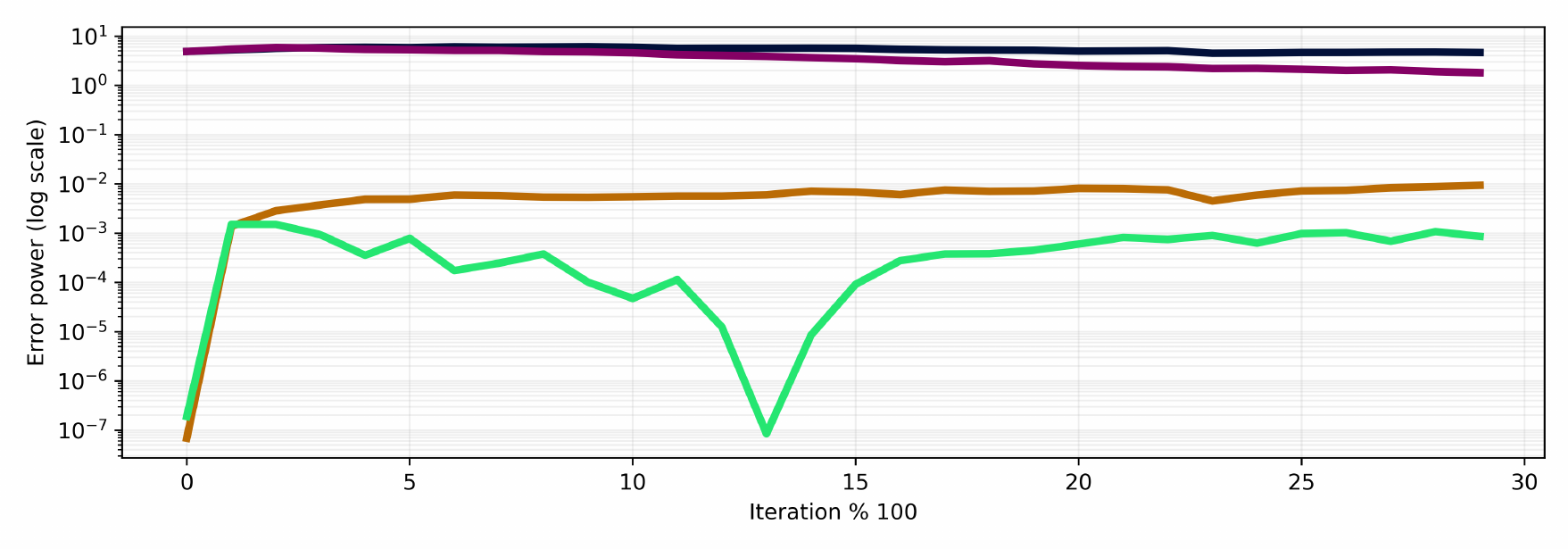}
    \label{fig:kdv_spectral_training_montecarlo3}
  \end{subfigure}
  \caption{We plot the spectral bias maximum error power and high error power according to Algorithm \ref{alg:radial_power_alg} for the KdV equation across a Fourier enhanced PINN with Monte carlo methods and a baseline PINN across 3,000 training iterations for 3 instances of training. We use a batch size of 1,000. The Monte Carlo collocation points are the exact same as the physics points here. We use the same scenario as Figure \ref{fig:kdv_spectral_training_veryhighbatch}.}
  \label{fig:kdv_spectral_training_montecarlo}
\end{figure}

\begin{figure}[htbp]
  \centering
  \begin{subfigure}[b]{\textwidth}
    \centering
    \includegraphics[scale=0.375]{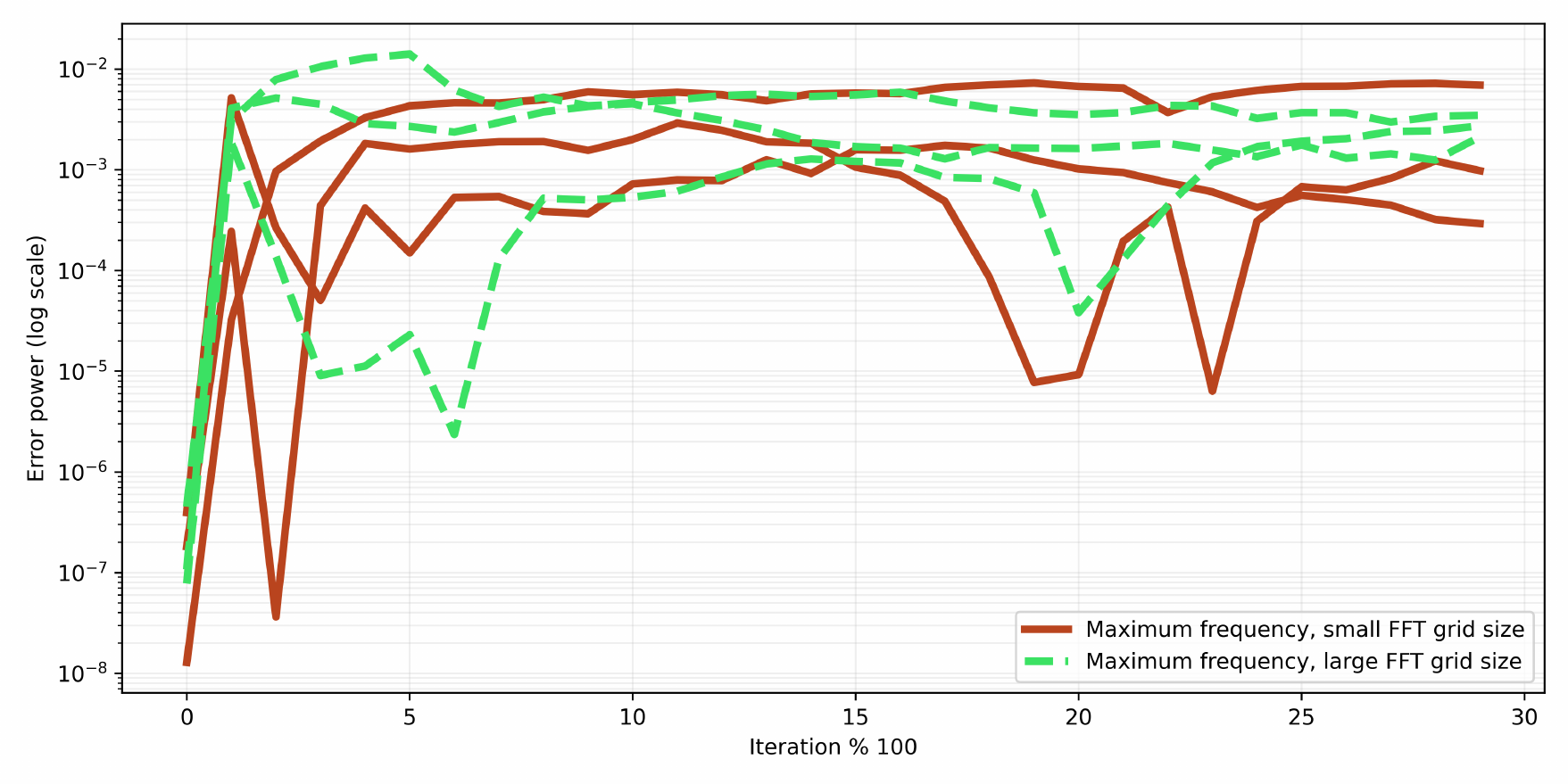}
    \label{fig:kdv_fft_smallvslargegrid}
  \end{subfigure}
  \caption{We plot the spectral bias maximum error power and high error power according to Algorithm \ref{alg:radial_power_alg} for the KdV equation across two FFT scenarios of a Fourier enhanced PINN and across 3 instances of training with: (1) a small FFT spatial grid between 31,61 in dimension each iteration; (2) a large FFT spatial grid between 151, 301 in dimension each iteration. Our hyperparameters are consistent with Figure \ref{fig:kdv_spectral_training_veryhighbatch}. As we can see, the FFT collocation sampling has an indiscernible effect on training and spectral bias.}
  \label{fig:kdv_fft_smallvslargegrid}
\end{figure}

\begin{figure}[htbp]
  \centering
  \begin{subfigure}[b]{\textwidth}
    \centering
    \includegraphics[scale=0.375]{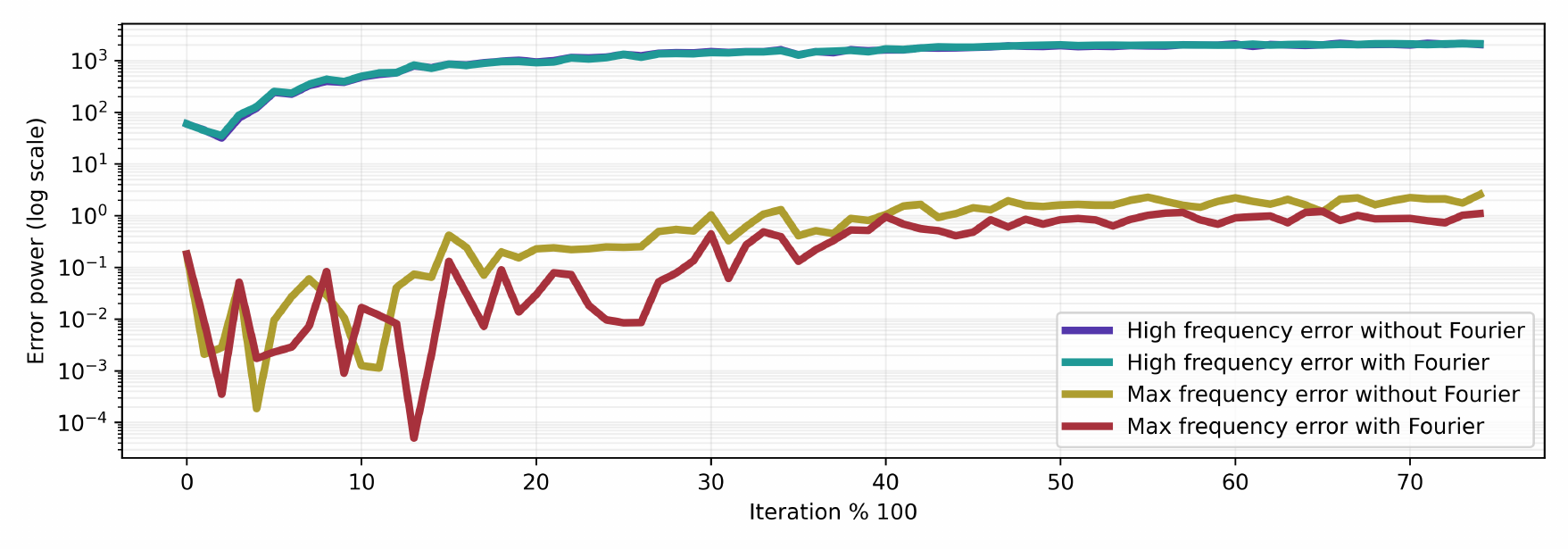}
    \label{fig:navier_stokes_spectral_training1}
  \end{subfigure}

  \begin{subfigure}[b]{\textwidth}
    \centering
    \includegraphics[scale=0.375]{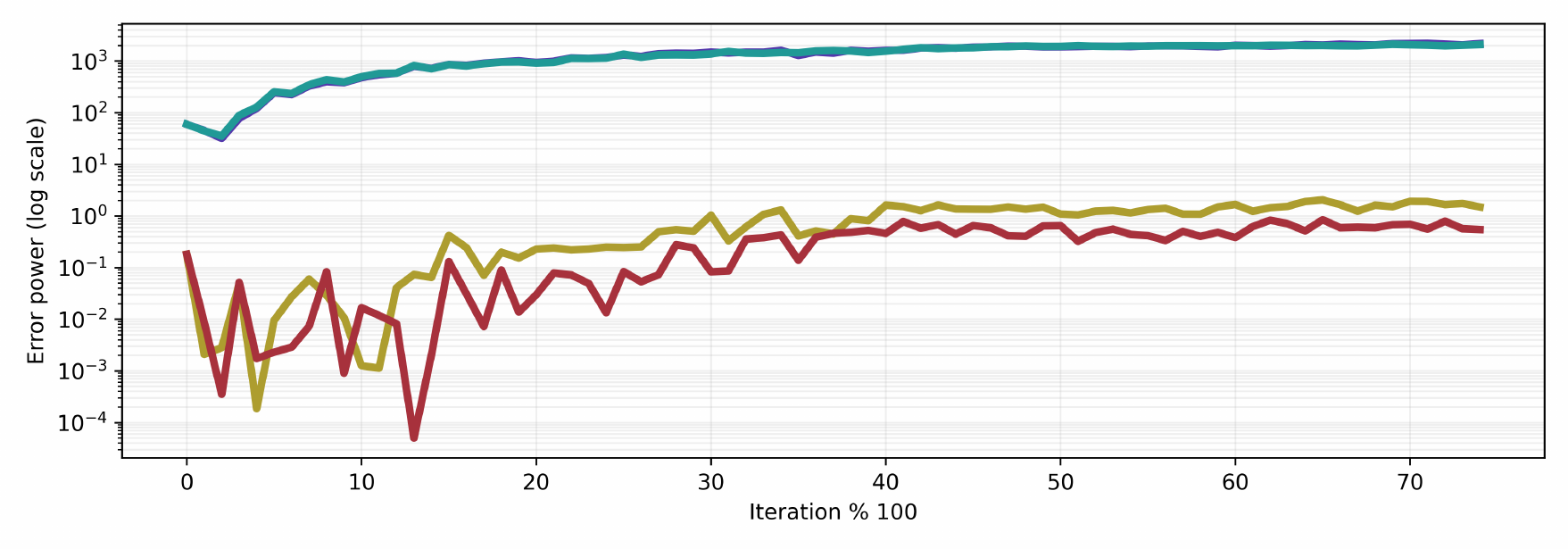}
    \label{fig:navier_stokes_spectral_training2}
  \end{subfigure}
  \caption{We plot the spectral bias maximum error power and high error power according to Algorithm \ref{alg:mc_enhanced_loss} on 2 instances of training Navier-Stokes data corresponding to the scenario of Figure \ref{fig:navier_stokes_pinn_example2} We use Fourier feature embedding of $\sigma = 6.0$ ($2\pi$ omitted), the SOAP optimizer, learning rate of $9.5\text{e}-4$, and $\lambda_{\text{Fourier}}=1$. Generally, we found no change in results with the Adam optimizer. Here, we use Monte Carlo techniques with the same batch as the physics loss with batch size 1,000. It is crucial to note the two instances are near identical, thus this example is highly consistent upon retraining. We restrict time domain to have Lebesgue measure $\mu([t_0,T]) = 0.5$ in this example (compare with Figure \ref{fig:navier_stokes_pinn_example} and section \ref{app:navier-stokes}).}
  \label{fig:navier_stokes_spectral_training}
\end{figure}

\begin{figure}[htbp]
  \vspace{0mm}
  \centering
  \includegraphics[scale=0.625]{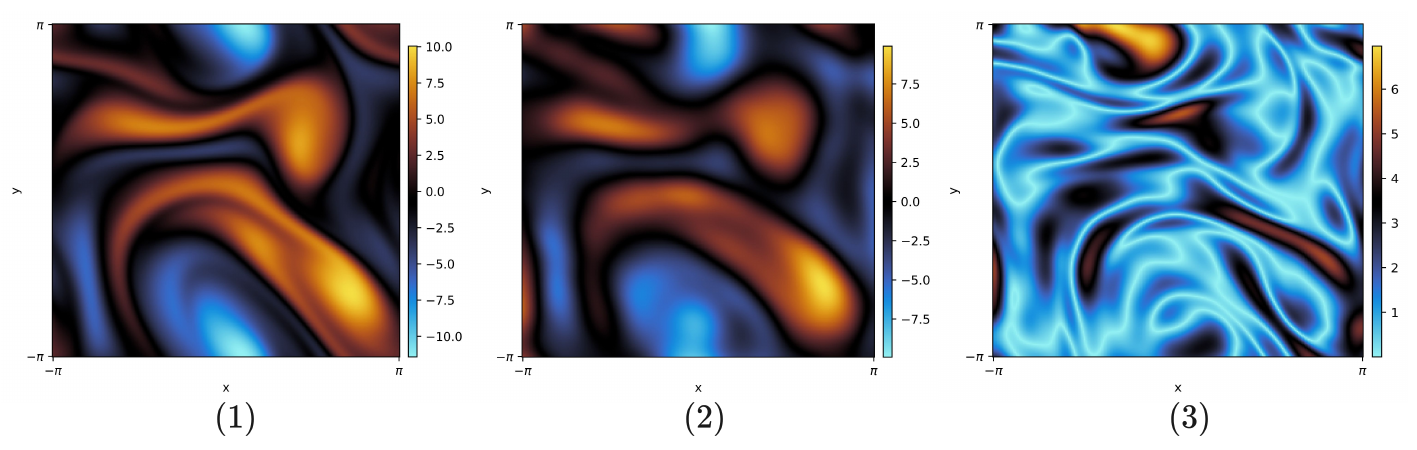}
  \caption{We provide a PINN scenario on Navier-Stokes data corresponding to Figure \ref{fig:navier_stokes_spectral_training}. We plot (1) the true solution examined at $t > t_0$; (2) the PINN solution at $t > t_0$ after $\sim 400$ training iterations (we remark this is still considered early training, so we examine the effects of spectral bias learned more quickly); (3) the absolute pointwise error}
  \label{fig:navier_stokes_pinn_example2}
\end{figure}

\begin{figure}[htbp]
  \vspace{0mm}
  \centering
  \includegraphics[scale=0.525]{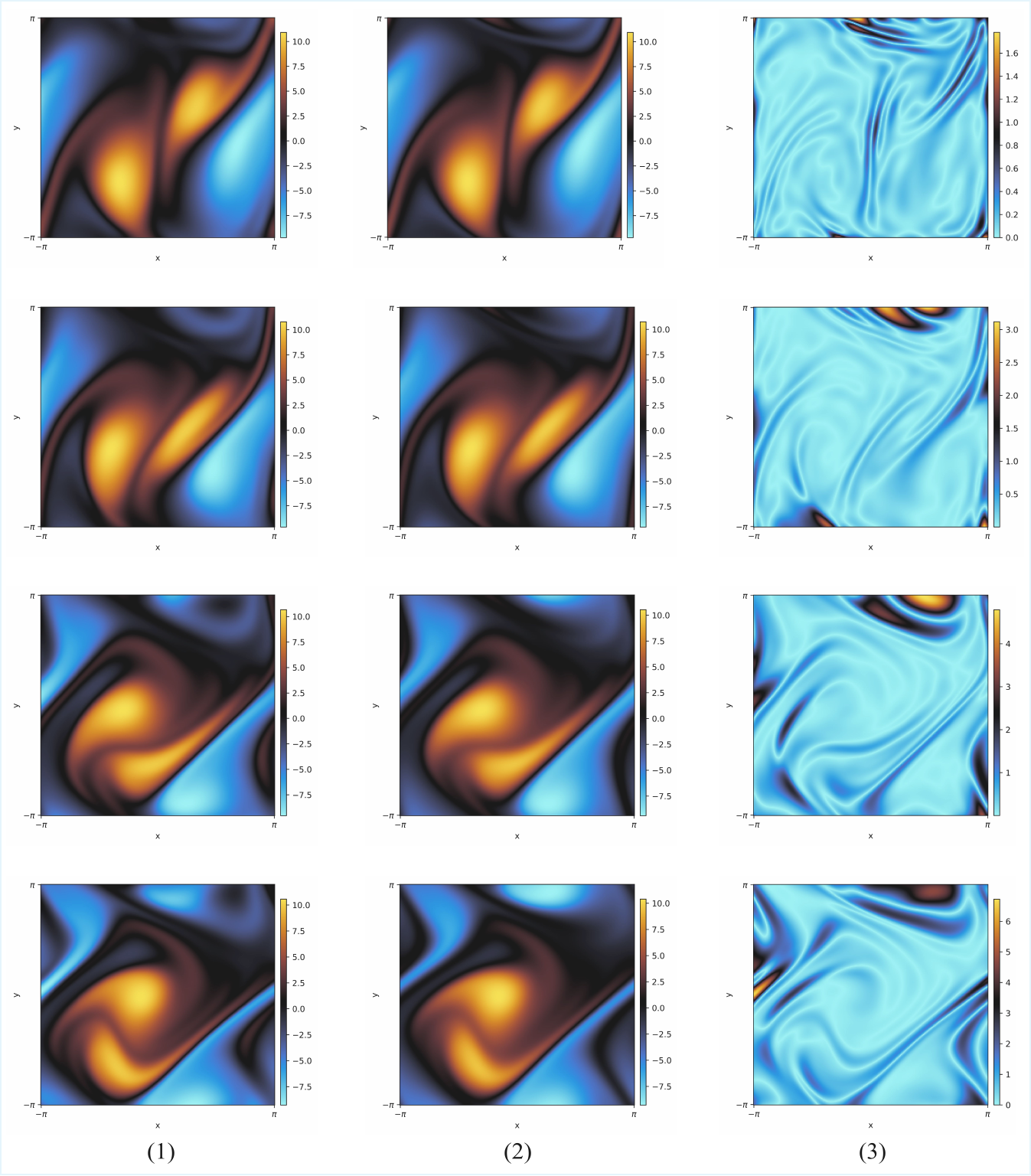}
  \caption{We provide a PINN scenario on Navier-Stokes data under (1) our PINN solution, which is Fourier enhanced (with FFTs), (2) a FFT numerical solver, and (3) the absolute pointwise error. For the PINN, we use a Fourier feature embedding with $\sigma=3.0$ ($ 2\pi$ omitted), the SOAP optimizer with learning rate $1\text{e}{-3}$ and lowered to $5\text{e}{-4}$ in late training, fixed weights of $\lambda_{\text{physics}} = 1$, $\lambda_{\text{boundary}} = 10$, $\lambda_{\text{Fourier}} = 0.025 \times \lambda_{\text{physics}}$ for some of training and some of training with the grad norm procedure of \cite{wang2023expertsguidetrainingphysicsinformed}. We use a batch size of 3,000, and $T_{\text{max}} - T_{\text{min}} = 1$. We use a neural network width of 400 and a depth of 4 with a vanilla architecture with $\sin(\cdot)$ activation, and train for approximately $10,000$ epochs. This data was used for Table \ref{tab:method_errors}. }
  \label{fig:navier_stokes_pinn_example}
\end{figure}

\begin{figure}[htbp]
  \vspace{0mm}
  \centering
  \includegraphics[scale=0.625]{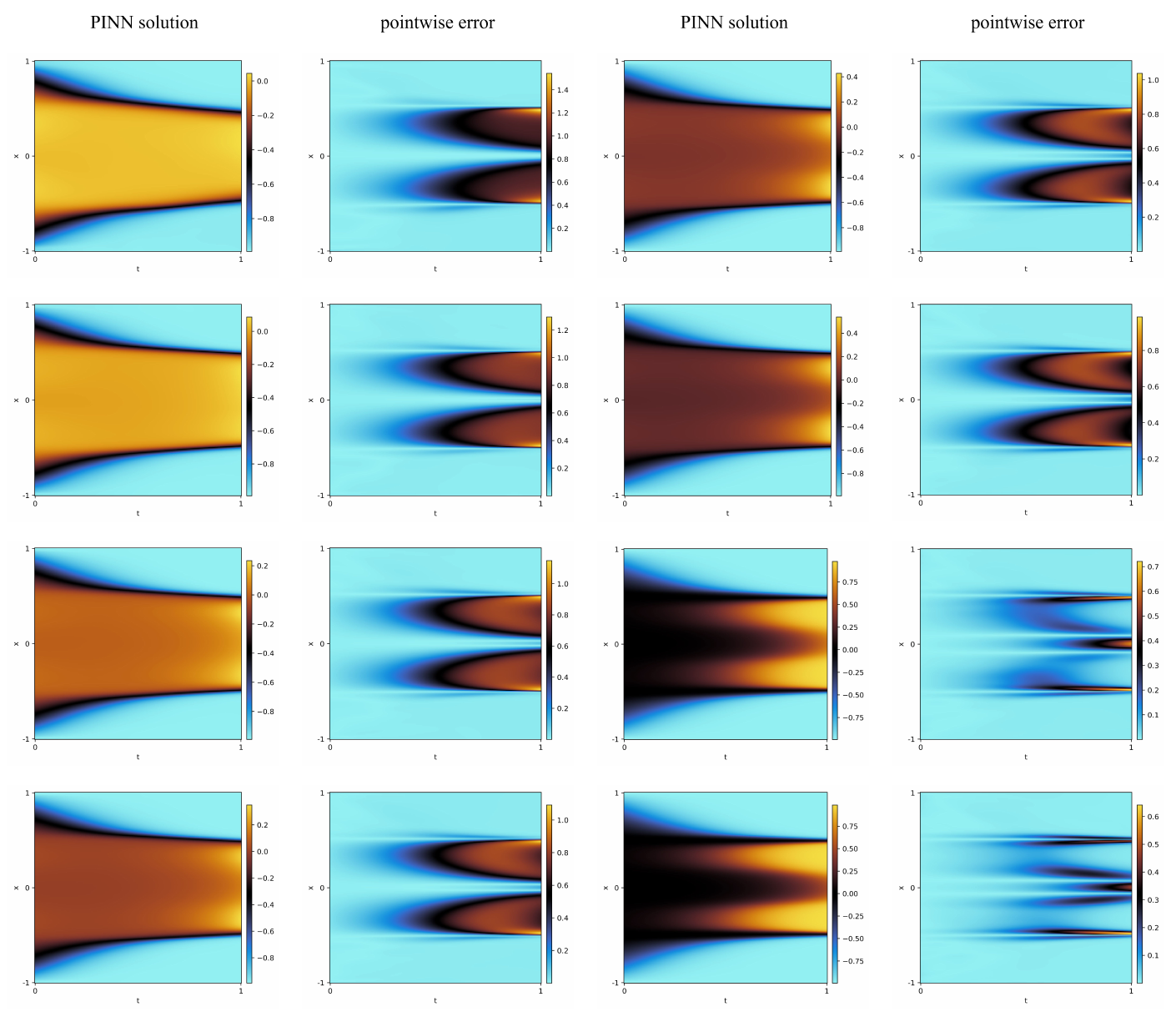}
  \caption{We plot our Allen-Cahn experiment with 35 collocation point sampling and FFTs in eight 30,000 training iterations batch increments. We demonstrate the $L^2$ error effect is a drop-off rather than gradual.}
  \label{fig:allencahn_batched}
\end{figure}

\end{document}